\pdfoutput=1
\documentclass{article}

\usepackage{esfm_preprint}
\usepackage{needspace}
\usepackage{xcolor}
\usepackage{times}
\usepackage{microtype}
\usepackage{graphicx}
\usepackage{subcaption}
\usepackage{booktabs}
\usepackage{bm}
\usepackage[hidelinks]{hyperref}
\usepackage{algorithm}
\usepackage{algorithmic}

\usepackage{amsmath}
\usepackage{amssymb}
\usepackage{mathtools}
\usepackage{amsthm}
\usepackage[capitalize,noabbrev]{cleveref}
\crefname{appendix}{Appendix}{Appendices}
\Crefname{appendix}{Appendix}{Appendices}
\crefname{subappendix}{Appendix}{Appendices}
\Crefname{subappendix}{Appendix}{Appendices}
\crefname{subsubappendix}{Appendix}{Appendices}
\Crefname{subsubappendix}{Appendix}{Appendices}

\numberwithin{equation}{section}

\theoremstyle{plain}
\newtheorem{theorem}{Theorem}[section]
\newtheorem{proposition}{Proposition}[section]

\theoremstyle{definition}

\theoremstyle{remark}

\AddToHook{env/theorem/begin}{\crefalias{section}{theorem}}
\AddToHook{env/proposition/begin}{\crefalias{section}{proposition}}
\AddToHook{env/lemma/begin}{\crefalias{section}{lemma}\crefalias{theorem}{lemma}}
\AddToHook{env/corollary/begin}{\crefalias{section}{corollary}\crefalias{theorem}{corollary}}
\AddToHook{env/definition/begin}{\crefalias{section}{definition}\crefalias{theorem}{definition}}
\AddToHook{env/assumption/begin}{\crefalias{section}{assumption}\crefalias{theorem}{assumption}}
\AddToHook{env/remark/begin}{\crefalias{section}{remark}\crefalias{theorem}{remark}}
\crefname{theorem}{Theorem}{Theorems}
\Crefname{theorem}{Theorem}{Theorems}
\crefname{proposition}{Proposition}{Propositions}
\Crefname{proposition}{Proposition}{Propositions}
\crefname{lemma}{Lemma}{Lemmas}
\Crefname{lemma}{Lemma}{Lemmas}
\crefname{corollary}{Corollary}{Corollaries}
\Crefname{corollary}{Corollary}{Corollaries}
\crefname{definition}{Definition}{Definitions}
\Crefname{definition}{Definition}{Definitions}
\crefname{assumption}{Assumption}{Assumptions}
\Crefname{assumption}{Assumption}{Assumptions}
\crefname{remark}{Remark}{Remarks}
\Crefname{remark}{Remark}{Remarks}

\newcommand{\R}{\mathbb{R}}
\newcommand{\dd}{\,\mathrm{d}}
\newcommand{\e}{\mathrm{e}}
\newcommand{\Acal}{\mathcal{A}}
\newcommand{\Bcal}{\mathcal{B}}

\newcommand{\Dirac}{\delta}
\newcommand{\norm}[1]{\left\lVert #1 \right\rVert}
\newcommand{\abs}[1]{\left\lvert #1 \right\rvert}
\newcommand{\yp}{y_{\star}}
\newcommand{\best}[1]{\ensuremath{\boldsymbol{#1}}}
\newcommand{\second}[1]{\underline{#1}}

\title{ESFM: Reconstructing Unbalanced Dynamics via Reduction and Extended-Space Flow Matching}

\author{Ruoyu Wang$^{2,\dag}$,\;\; Xinyu Wang$^{1,\dag}$,\;\; Tiejun Li$^{1,2,3,4,*}$\\[0.5ex]
\footnotesize $^1$ LMAM and School of Mathematical Sciences, Peking University.\\
\footnotesize $^2$ Center for Machine Learning Research, Peking University.\\
\footnotesize $^3$ National Engineering Laboratory for Big Data Analysis and Applications.\\
\footnotesize $^4$ Center for Data Science, Peking University.\\
\footnotesize $^\dag$ These authors contribute equally to this work.\\
\footnotesize $^*$ Correspondence: \href{mailto:tieli@pku.edu.cn}{tieli@pku.edu.cn}}
\hypersetup{pdftitle={ESFM: Reconstructing Unbalanced Dynamics via Reduction and Extended-Space Flow Matching},
pdfauthor={Ruoyu Wang, Xinyu Wang, Tiejun Li}}

\begin{document}

\maketitle

\begin{abstract}
Biological population dynamics are often unbalanced, involving simultaneous changes in cellular states and population abundance.  Existing unbalanced flow-matching methods typically rely on cone geometry-based loss formulations. While mathematically consistent, these formulations often induce biologically unfavorable state evolutions. We propose {\bf Extended Space Flow Matching (ESFM)}, which lifts unbalanced dynamics into conservative transport in an extended space of state and log mass. This jointly models state transitions and abundance changes as a single conservative process, enabling prescribed conditional paths to be integrated with compatible endpoint couplings and simulation-free flow matching objectives. For linear interpolation of state and log mass, ESFM admits an analytical reduction to a tractable optimal transport formulation. On synthetic and real single-cell snapshot datasets, ESFM accurately predicts both state and abundance dynamics. Extended space formulations thus offer a flexible, biologically preferred framework for unbalanced flow matching.
\end{abstract}

\section{Introduction}
\label{sec:introduction}

Single-cell RNA sequencing (scRNA-seq) provides discrete snapshots of a
continuously evolving biological system. Because sequencing is destructive,
individual cells cannot be followed over time, while experimental cost limits
data collection to a few time points
\citep{zheng2017massively,bunne2024optimal,zhang2025review}.
Single-cell trajectory inference therefore seeks to recover continuous dynamics
from population snapshots \citep{waddingot,trajectorynet}.

Continuous-time models provide several approaches to this problem. Neural
ordinary differential equations (ODEs) parameterize time-dependent velocity
fields but typically require repeated numerical integration during training
\citep{chen2018neural,trajectorynet,mioflow}. Dynamic optimal transport (OT)
provides a complementary geometric view, with the Benamou--Brenier formulation
linking static endpoint couplings to least-action continuous flows
\citep{benamou2000computational}. Flow matching (FM) instead learns continuous
flows by regressing conditional vector fields without simulating the ODE during
training
\citep{cfm_lipman,liu2022flow,albergo2022building,klein2024genot,
rohbeck2025modeling}. In general FM constructions, the endpoint coupling and
conditional path may be specified separately. OT conditional flow matching
(OT-CFM) links them through a common transport geometry by combining OT endpoint
couplings with displacement interpolation
\citep{tong2023conditional,pooladian2023multisample}. This coupling--path
consistency provides a principled geometric basis for flow matching. However,
standard FM and OT-CFM operate on normalized probability measures and do not
directly model changes in total population abundance.

Biological populations can grow or decline through proliferation, death, and
differential survival
\citep{TIGON,DeepRUOT,sun2025variational,peng2026stvcr}. Unbalanced OT relaxes
mass conservation through source terms or relaxed marginal constraints
\citep{liero2016optimal,chizat2018interpolating,chizat2018unbalanced}, and
recent unbalanced flow matching methods incorporate mass variation through
unbalanced couplings, explicit growth dynamics, or unbalanced transport
geometries
\citep{eyring2024unbalancedness,wang2025joint,peng2026wfrfm,ying2026sudo}.
A prominent class of unbalanced flow-matching methods derives endpoint couplings
and conditional dynamics from a prescribed unbalanced transport geometry,
particularly WFR formulation
\citep{liero2016optimal,chizat2018interpolating,chizat2018unbalanced,peng2026wfrfm}. These approaches follow a forward design 
 principle: a prescribed transport geometry or growth penalty in expression state space determines both the endpoint
  coupling and the associated conditional dynamics. While this construction ensures internal consistency, it does not 
  necessarily yield biologically favorable state evolutions, as the resulting dynamics are driven solely by a prescribed, 
  simplistic loss function. 

We address unbalanced dynamics through a reduction approach in an extended space and
propose the \textbf{Extended Space Flow Matching (ESFM)} framework for
unbalanced single-cell dynamics. ESFM introduces a log mass coordinate alongside the cell state, lifting
nonconservative dynamics into conservative transport in an extended space.
In this representation, state transitions and relative growth
are modeled jointly, while physical population mass is recovered by projection.

\begin{figure}[t]
    \centering
    \includegraphics[width=1\textwidth]{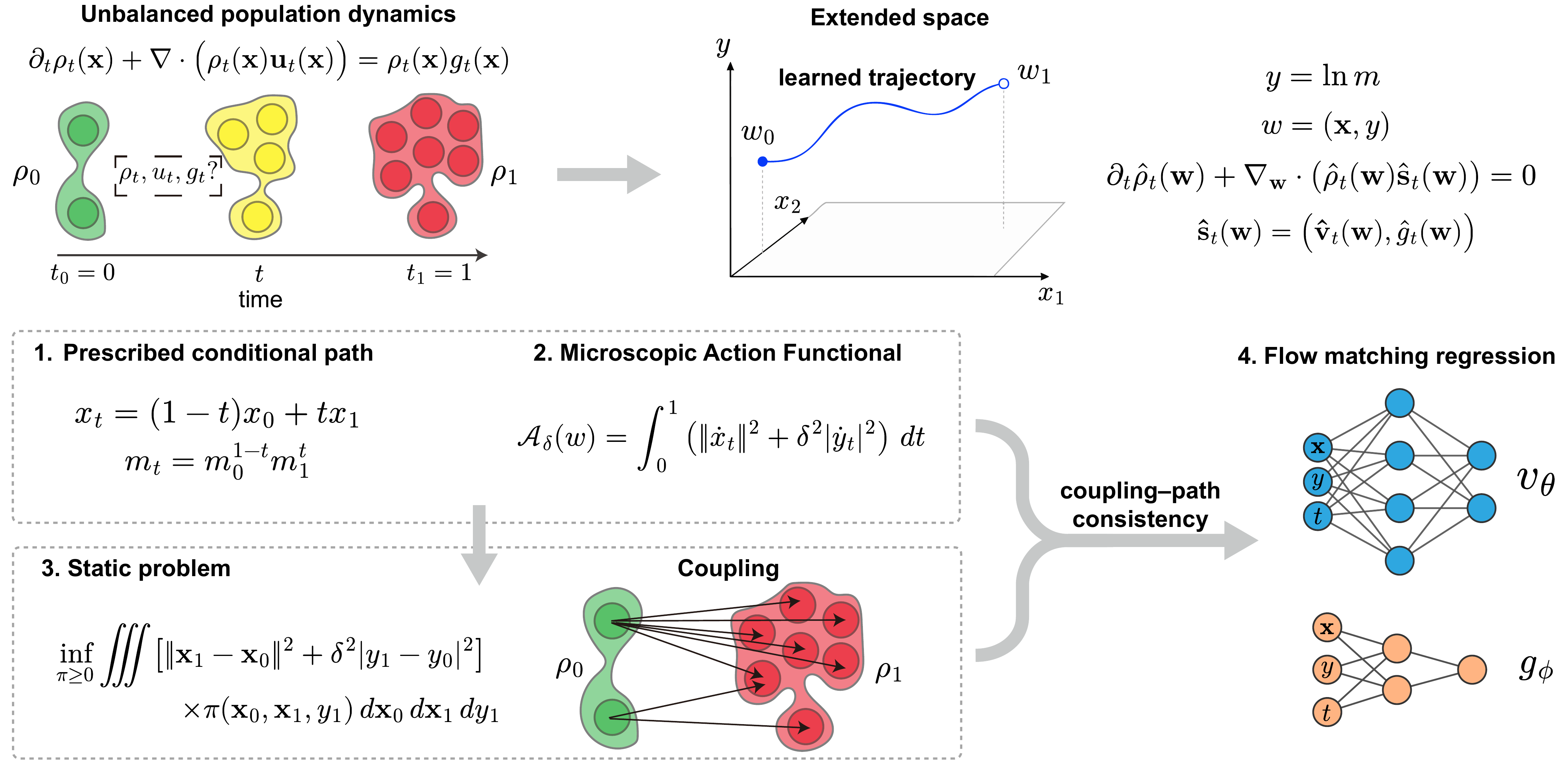}
    \caption{\textcolor{black}{
    \textbf{Overview of Extended Space Flow Matching (ESFM).}
    ESFM lifts unbalanced dynamics to conservative transport in the extended
    space $\bm w=(\bm{x},\ln m)$, starts from a prescribed conditional path,
    and derives a compatible endpoint coupling and flow matching objective.
    }}
    \label{fig:esfm_overview}
\end{figure}

Within this formulation, ESFM adopts a path-first construction for the specific
instance studied here, where linear interpolation of cell state and log mass is
paired with a compatible action, endpoint coupling, and simulation-free flow
matching objective. This design allows application specific dynamical assumptions
to be incorporated through path specification and admits an analytical reduction
to a static optimal transport problem. More general path and action designs can
also be incorporated into the extended space. Figure~\ref{fig:esfm_overview}
summarizes the overall construction.

Our main contributions are as follows:
\begin{itemize}

\item We introduce an extended
space formulation for unbalanced flow matching that lifts unbalanced
dynamics into conservative transport by introducing a log mass coordinate
alongside the state space. 
This formulation enables state transport and relative growth to be modeled
jointly, while conditional paths and endpoint couplings are derived consistently
from a common extended-space transport formulation.

\item We establish theoretical connections between the extended-space dynamic
formulation and its static transport counterpart. For linear state motion and
linear interpolation in log mass, corresponding to constant relative growth,
we analytically eliminate the latent log mass coordinate and derive a
reduced optimal transport problem with an explicit abundance change cost.

\item We develop an entropically regularized Sinkhorn-like solver and a
simulation-free flow matching objective for jointly learning single-cell
state transport and population growth. Experiments demonstrate accurate
prediction of state and abundance dynamics on synthetic and real single-cell
snapshot datasets.

\end{itemize}

\section{Related Work}
\label{sec:related}

\paragraph{Flow matching.}
Flow matching learns continuous dynamics by regressing vector fields associated
with tractable conditional probability paths, avoiding ODE simulation during
training \citep{cfm_lipman,liu2022flow,albergo2022building}. A key design choice
is the conditional path: many methods use simple linear interpolation, while
others adapt or learn interpolants from data geometry, conditioning variables,
or multiple observed snapshots
\citep{liu2022flow,kapusniak2024metric,rohbeck2025modeling,
lee2025mmsfm,kviman2026alicfm}. Flow matching has also been applied directly to
temporal single-cell snapshots using OT-based couplings and coupling refinement
\citep{pu2026snapshots}. Another key choice is the endpoint coupling, which determines
how source and target samples are paired and may be independent, prescribed, or
data dependent \citep{albergo2024datadependent}. In general flow matching
constructions, the coupling and conditional path can be specified independently
and need not arise from a common geometric or variational principle.

\paragraph{Optimal transport and coupling--path consistency.}
Optimal transport links endpoint coupling and conditional paths through a common
geometry. In quadratic OT, the static Kantorovich and dynamic Benamou--Brenier
formulations induce the same squared Euclidean endpoint cost and constant-speed
linear interpolation
\citep{kantorovich1942translocation,benamou2000computational}. OT-CFM exploits
this correspondence by sampling endpoint pairs from the OT coupling and
regressing velocities along the associated displacement paths
\citep{tong2023conditional}. Related methods extend this construction through
minibatch couplings, entropic regularization, Schr\"odinger bridges, or
geometry-aware paths
\citep{pooladian2023multisample,klein2024genot,sflowmatch,
kapusniak2024metric}. Lagrangian Flow Matching further derives least-action
paths and static transport costs from a prescribed Lagrangian
\citep{du2026lagrangian}. These methods start from a prescribed geometry or action. ESFM instead starts from a prescribed conditional path together with a compatible action and derives the corresponding endpoint coupling. 

\paragraph{Unbalanced transport and single-cell dynamics.}
Changes in cell abundance due to proliferation and death make unbalanced
transport particularly relevant for single-cell snapshot dynamics. Unbalanced
OT relaxes mass conservation, while WFR jointly models state transport and
population growth
\citep{benamou2003numerical,chizat2018unbalanced,liero2016optimal,
chizat2018interpolating,kondratyev2016}. These ideas have motivated continuous
single-cell models including TIGON, DeepRUOT, Var-RUOT, stVCR, and MIOFlow 2.0
\citep{TIGON,DeepRUOT,sun2025variational,peng2026stvcr,sun2026mioflow2},
as well as simulation-free approaches based on relaxed couplings, explicit
growth fields, branching dynamics, or general growth penalties
\citep{eyring2024unbalancedness,wang2025joint,maddu2026probability,
ying2026beyond,ying2026sudo}. WFR-FM and its extensions derive conditional paths
and couplings from WFR geometry, while SUDO generalizes this geometry-first view
to convex growth penalties
\citep{peng2026wfrfm,peng2026mustfm,wang2026wfrmfm,ying2026sudo}. ESFM instead
adopts a path-first design, deriving a compatible action, endpoint coupling, and
flow matching objective from a prescribed conditional path.

\section{Mathematical Background}
\label{sec:background}

\paragraph{Conditional flow matching.}
Flow matching learns a time-dependent vector field that transports a source
distribution to a target distribution. Conditional flow matching (CFM)
avoids requiring the marginal probability path and its velocity field in
closed form by introducing tractable conditional probability paths
\citep{cfm_lipman,tong2023conditional}. For a condition $z$, typically an
endpoint pair $z=(\bm{x}_0,\bm{x}_1)$, CFM minimizes
\begin{equation}
    \mathcal L_{\mathrm{CFM}}(\theta)
    =
    \mathbb E_{t,z,\bm{x}\sim\rho_t(\cdot\mid z)}
    \left[
    \norm{
    \bm u_\theta(\bm{x},t)
    -
    \bm u_t(\bm{x}\mid z)
    }^2
    \right].
    \label{eq:background-cfm}
\end{equation}
Thus, an endpoint-conditioned CFM construction involves two key components:
the distribution of endpoint pairs and the conditional path associated with
each pair. Additional background on marginal flow matching is provided in
Appendix~\ref{app:background-fm-ot}.

\paragraph{OT and coupling--path consistency.}
Optimal transport provides a principled way to link these two components
through a common transport geometry. For quadratic OT, the static
Kantorovich problem determines an optimal endpoint coupling, while the
corresponding particle-level geodesic is the constant-speed interpolation
\begin{equation*}
    \bm{x}_t
    =
    (1-t)\bm{x}_0+t\bm{x}_1,
    \qquad
    \bm u_t(\bm{x}_t\mid\bm{x}_0,\bm{x}_1)
    =
    \bm{x}_1-\bm{x}_0.
\end{equation*}
The endpoint coupling and conditional path therefore arise from the same
transport geometry. OT-CFM exploits this correspondence by sampling endpoint
pairs from the OT coupling and regressing the velocity field along the
associated displacement interpolation
\citep{benamou2000computational,tong2023conditional}. We refer to this shared
geometric origin as \emph{coupling--path consistency}.

\paragraph{Unbalanced dynamics and WFR geometry.}
To describe changes in population abundance, the state velocity is augmented
by a growth field. Along a particle trajectory carrying physical
mass $m_t>0$, the growth rate is
\begin{equation}
    g_t
    =
    \frac{\dd \ln m_t}{\dd t}
    =
    \frac{\dot m_t}{m_t}.
    \label{eq:background-growth-rate}
\end{equation}
At the population level, a finite mass density $\rho_t$ evolves according to
the unbalanced continuity equation
\begin{equation}
    \partial_t \rho_t(\bm{x})
    +
    \nabla\cdot
    \left(
    \rho_t(\bm{x})\bm u_t(\bm{x})
    \right)
    =
    \rho_t(\bm{x})g_t(\bm{x}).
    \label{eq:physical-continuity-main}
\end{equation}

A representative geometry for such dynamics is
Wasserstein--Fisher--Rao (WFR), whose dynamic formulation jointly penalizes
state transport and growth
\citep{liero2016optimal,chizat2018unbalanced}:
\begin{equation*}
\begin{aligned}
    \inf_{\rho,\bm u,g}\quad
    &
    \int_0^1
    \int_{\R^d}
    \left(
    \norm{\bm u_t(\bm{x})}^2
    +
    \delta^2\abs{g_t(\bm{x})}^2
    \right)
    \rho_t(\bm{x})
    \dd\bm{x}\dd t,
    \\
    \text{s.t.}\quad
    &
    \partial_t\rho_t
    +
    \nabla\cdot(\rho_t\bm u_t)
    =
    \rho_t g_t,
    \qquad
    \rho_{t=0}=\rho_0,
    \quad
    \rho_{t=1}=\rho_1.
\end{aligned}
\end{equation*}
As in balanced OT, the WFR geometry jointly determines endpoint pairing and
particle-level interpolation, providing the geometric basis of WFR-FM
\citep{peng2026wfrfm}. 

\section{ESFM Framework}
\label{sec:esfm-framework}
\textcolor{black}{Our starting point is to incorporate assumptions about state and mass evolution
through conditional particle paths, and to construct a transport problem consistent with those paths.
We first introduce an extended space in which both state motion and mass change are represented as transport.
As illustrated in \cref{fig:esfm_overview}, we then prescribe linear state and log-mass paths,
construct a compatible microscopic action, derive the dynamic and static OT problems,
and reduce the latter to a computable coupling in the original state space.
Finally, this coupling and the prescribed paths provide the flow-matching regression targets.}

\subsection{Extended Space}
\label{subsec:background-extended-space}

\textcolor{black}{To treat state motion and growth within a single transport description,
we represent mass change as motion along an additional log-mass coordinate.}
ESFM defines the dynamics in the extended space $\mathcal W=\R^d\times\R$,
where each particle is described by its cell state $\bm{x}$ and log mass
$y=\ln m$. The joint state $\bm w=(\bm{x},y)$ evolves according to
\begin{equation*}
    \frac{\dd}{\dd t}
    \begin{pmatrix}
        \bm{x}_t\\
        y_t
    \end{pmatrix}
    =
    \begin{pmatrix}
        \hat{\bm u}_t(\bm{x}_t,y_t)\\
        \hat g_t(\bm{x}_t,y_t)
    \end{pmatrix}.
\end{equation*}
Here $\hat{\bm u}_t$ and $\hat g_t$ denote the state velocity and relative
growth field in the extended space.
\textcolor{black}{Since $\dot y_t=\dot m_t/m_t$, velocity along this coordinate is precisely the relative growth rate.}
Let $\hat\rho_t$ denote the particle number density. Its weighted projection
recovers the physical mass density:
\begin{equation*}
    \rho_t(\bm{x})=\int_{\R}\e^y\hat\rho_t(\bm{x},y)\dd y.
\end{equation*}
We assume $\int_{\R^d}\rho_0(\bm x)\dd\bm x=1$.
We assume that each source particle has unit mass at $t=0$.
\textcolor{black}{The observed terminal population specifies physical mass, but does not prescribe
how it is distributed over the latent log-mass coordinate. Accordingly, the boundary conditions are}
\begin{equation}
    \hat\rho_0(\bm{x},y)=\rho_0(\bm{x})\Dirac_0(y),
    \qquad
    \int_{\R}\e^y\hat\rho_1(\bm{x},y)\dd y=\rho_1(\bm{x}).
\label{eq:extended-boundary}
\end{equation}
Thus $\hat\rho_0$ and $\hat\rho_1$
are probability measures, whereas $\rho_1$ need not have unit mass.

\textcolor{black}{Particle weights change through motion along $y$, while particle number is conserved
in the full extended space. This gives the conservative continuity equation:}
\begin{equation}
    \partial_t\hat\rho_t
    +\nabla_{\bm x}\cdot(\hat\rho_t\hat{\bm u}_t)
    +\partial_y(\hat\rho_t\hat g_t)=0.
\label{eq:background-extended-continuity}
\end{equation}
\textcolor{black}{The following theorem shows that mass-weighted projection of the extended-space continuity equation recovers the original unbalanced dynamics in Eq.~\eqref{eq:physical-continuity-main}.}
\begin{theorem}
\label{thm:extended-projection}
Assume that the extended density and growth field satisfy
\[
    \lim_{y\to\pm\infty}
    \e^y\hat\rho_t(\bm x,y)\hat g_t(\bm x,y)=0.
\]
After mass projection onto the physical state space,
\cref{eq:background-extended-continuity} implies
\cref{eq:physical-continuity-main} with the corresponding projected velocity
and growth fields.
\end{theorem}
\textcolor{black}{Intuitively, transport along log mass becomes a growth or death source term
after mass-weighted projection. Thus, conservation in the extended space is compatible
with changing physical population mass.}
The projected fields and proof are given in \cref{app:projection}.

\subsection{Prescribed Path and Compatible Action}
\label{sec:prescribed-path-action}

\textcolor{black}{We now specify the conditional dynamics that will guide the transport construction.
Over a short interval with approximately stable growth conditions, constant relative growth
provides a simple prior for mass evolution; straight state motion serves as a local approximation.}
For extended endpoints
$\bm w_0=(\bm x_0,y_0)$ and
$\bm w_1=(\bm x_1,y_1)$,
we first prescribe linear interpolation in both state and log mass:
\begin{equation}
    \bm{x}_t=(1-t)\bm{x}_0+t\bm{x}_1,
    \qquad
    y_t=(1-t)y_0+ty_1,
    \qquad
    m_t=m_0^{1-t}m_1^t.
\label{eq:main-path}
\end{equation}
This path corresponds to constant state velocity and constant relative growth
between the prescribed endpoints.

\textcolor{black}{To make endpoint pairing reflect these same assumptions, we seek an action
whose minimizing paths are the prescribed interpolants. A compatible choice is}
\begin{equation}
    \Acal_\delta(\bm w)
    =\int_0^1
    \left(\norm{\dot{\bm{x}}_t}^2+\delta^2\abs{\dot y_t}^2\right)\dd t,
\label{eq:main-action}
\end{equation}
where $\delta>0$ controls the cost of changing log mass relative to state transport.
\textcolor{black}{For fixed endpoints, constant extended velocity minimizes this quadratic action.
The minimum is therefore the cost of connecting the endpoints along the prescribed path:}
\begin{equation}
    c(\bm w_0,\bm w_1)
    =\norm{\bm{x}_1-\bm{x}_0}^2+\delta^2\abs{y_1-y_0}^2.
\label{eq:main-pair-cost}
\end{equation}
The optimality argument is given in \cref{app:dynamic-static-equivalence}.

More general formulation and examples with different prescribed paths are discussed in 
\cref{app:general-extensions}; however, such cases generally do not admit analytically reduced formulations.

\subsection{Extended Space Dynamic and Static OT}
\label{subsec:main-projection}

\textcolor{black}{The microscopic action specifies the cost of one path; we next seek a population flow
that matches the observed endpoints at minimum total cost.
Integrating the action over the extended particle distribution gives the dynamic OT problem,
subject to \cref{eq:background-extended-continuity,eq:extended-boundary}:}
\begin{equation}
    \Bcal_\delta(\rho_0,\rho_1)
    =
    \inf_{\hat\rho,\hat{\bm u},\hat g}
    \int_0^1\int_{\mathcal W}
    \left(\norm{\hat{\bm u}_t(\bm w)}^2+\delta^2\abs{\hat g_t(\bm w)}^2\right)
    \hat\rho_t(\bm w)\dd\bm w\dd t.
\label{eq:main-dynamic-esfm}
\end{equation}
\textcolor{black}{To compute an endpoint coupling without optimizing an entire time-dependent flow,
we use the minimum path cost in \cref{eq:main-pair-cost} to formulate the static problem:}
\begin{equation}
    \mathcal K_\delta(\rho_0,\rho_1)
    =
    \inf_{\substack{\hat\rho_1\in\mathcal P(\mathcal W)\\
    \hat\pi\in\Pi(\hat\rho_0,\hat\rho_1)}}
    \int_{\mathcal W^2}
    c(\bm w_0,\bm w_1)\dd\hat\pi(\bm w_0,\bm w_1).
\label{eq:main-extended-static}
\end{equation}
Here $\mathcal P(\mathcal W)$ denotes the set of probability measures on
$\mathcal W$, and $\Pi(\hat\rho_0,\hat\rho_1)$ denotes probability couplings with
these marginals, subject to \cref{eq:extended-boundary}.
Unlike standard OT, ESFM optimizes $\hat\rho_1$ jointly with the coupling,
while fixing only its weighted projection.
\textcolor{black}{The next theorem shows that this endpoint optimization retains the optimal value
of the dynamic problem, linking the coupling to the chosen action and paths.}
\begin{theorem}
\label{thm:extended-dynamic-static}
The extended dynamic and static problems have equal optimal values:
$\Bcal_\delta(\rho_0,\rho_1)=\mathcal K_\delta(\rho_0,\rho_1)$.
\end{theorem}
The proof is provided in \cref{app:dynamic-static-equivalence}.

\subsection{Reduced ESFM Static OT}
\label{subsec:key-ot-result}
\label{subsec:main-computation}

\textcolor{black}{We now investigate how this static problem can be reformulated as an optimal transport problem in the original state space.}

The source condition fixes $y_0=0$, so the extended plan can be written as
$\pi(\bm{x}_0,\bm{x}_1,y_1)$. To bound individual particle growth, we impose
$y_1\le M$ with $M>1$. The choice of $M$ is described in \cref{app:experimental-details}.
Since $y_1=\ln m_1$, negative values of $y_1$ remain allowed and correspond
to mass decay.
Define the two semi-couplings
\[
    \gamma_0(\bm{x}_0,\bm{x}_1)=\int\pi(\bm{x}_0,\bm{x}_1,y_1)\dd y_1,
    \qquad
    \gamma_1(\bm{x}_0,\bm{x}_1)=\int\e^{y_1}\pi(\bm{x}_0,\bm{x}_1,y_1)\dd y_1.
\]
For each endpoint pair, $\gamma_0$ and $\gamma_1$ represent the initial and final mass,
respectively. Their ratio
$R(\bm{x}_0,\bm{x}_1)=\dd\gamma_1/\dd\gamma_0(\bm{x}_0,\bm{x}_1)$ is the local growth ratio, with $0<R(\bm{x}_0,\bm{x}_1)\le\e^M$.

\textcolor{black}{The following theorem shows how the extended-space static OT problem in
\cref{eq:main-extended-static} can be reformulated on the original state space.}
\begin{theorem}
\label{thm:local-log-mass}
For $M>1$, the extended static problem in \cref{eq:main-extended-static}
with the additional constraint $y_1\le M$ is equivalent to
\begin{equation}
\begin{aligned}
    \inf_{\gamma_0,\gamma_1\ge0}\quad
    &
    \iint
    \left[
    \norm{\bm{x}_1-\bm{x}_0}^2
    +
    \delta^2\Phi_M\!\left(R(\bm{x}_0,\bm{x}_1)\right)
    \right]\dd\gamma_0,
    \\
    \text{s.t.}\quad
    &
    \int\gamma_0(\bm{x}_0,\bm{x}_1)\dd\bm{x}_1=\rho_0(\bm{x}_0),
    \qquad
    \int\gamma_1(\bm{x}_0,\bm{x}_1)\dd\bm{x}_0=\rho_1(\bm{x}_1),
    \\
    &
     0<R(\bm{x}_0,\bm{x}_1)\le\e^M.
\end{aligned}
\label{eq:main-reduced-static}
\end{equation}
Here $\Phi_M(\cdot)$ is given by
\begin{equation*}
    \Phi_M(r)
    =
    \begin{cases}
    (\ln r)^2,
    &0<r\le \e^{\yp},
    \\[0.4em]
    \displaystyle
    \frac{2\yp}{\e^{\yp}}r+\yp^2-2\yp,
    &\e^{\yp}<r\le \e^M.
    \end{cases}
\end{equation*}
where $\yp\in(0,1)$ is the threshold determined by $M$ through the equaion \( M^2- \yp^2+2\yp-2\yp\e^{M-\yp}=0\).
\end{theorem}

\textcolor{black}{In this reduced formulation, state transport retains its squared Euclidean cost,
while mass change is penalized through $\delta^2\Phi_M(R)$.
The reduction preserves the extended-space optimum while making the coupling computable
using state pairs and growth ratios.}
The reduction and local optimizer are proved in \cref{app:local-log-mass-proof}.

\textcolor{black}{To compute the reduced coupling efficiently, we introduce entropic
regularization and derive an iterative solver for the resulting problem.}
\begin{proposition}
\label[proposition]{prop:entropic-solver}
The entropic regularization of \cref{eq:main-reduced-static} admits a
Sinkhorn-like algorithm with explicit alternating row and column updates.
\end{proposition}
The update formulas and proof are given in \cref{app:solver}.
For training, sample state pairs from normalized $\gamma_0$, set $y_0=0$,
and draw $y_1$ from the local optimizer in \cref{eq:appendix-log-mass-sampling}.

\Needspace{\baselineskip}
\textcolor{black}{We also check that the optimal transport cost is independent of how
source mass at a fixed state is divided among particles.}
\begin{proposition}
\label[proposition]{prop:particle-splitting}
Splitting a source particle into particles at the same state, while preserving
its total weight, does not change the optimal value of \cref{eq:appendix-discrete}.
\end{proposition}
The proof is given in \cref{app:particle-splitting}.

\subsection{Conditional Paths and ESFM Training}
\label{subsec:main-training}

\textcolor{black}{The coupling now supplies endpoint pairs and masses. The remaining task is to learn
velocity and growth fields from the conditional paths connecting these endpoints.}
We apply the balanced CFM construction of \citet{cfm_lipman} in extended
space, using separate networks $\bm v_\theta$ and $g_\phi$ to form
$\bm s_{\theta,\phi}=(\bm v_\theta,g_\phi)$.

\noindent\textbf{Marginal objective.}
For an extended density path $\hat\rho_t$ with velocity components
$(\textcolor{black}{\hat{\bm u}_t},\textcolor{black}{\hat g_t})$, the ideal regression objective is
\begin{equation}
    \mathcal L_{\mathrm{ESFM}}^{\mathrm{marg}}(\theta,\phi)
    =
    \int_0^1\int_{\mathcal W}
    \Big[
    \norm{\bm v_\theta(\bm w,t)-\textcolor{black}{\hat{\bm u}_t}(\bm w)}^2
    +\lambda\abs{g_\phi(\bm w,t)-\textcolor{black}{\hat g_t}(\bm w)}^2
    \Big]
    \hat\rho_t(\bm w)\dd\bm w\dd t.    
\label{eq:main-marginal-loss}
\end{equation}
\textcolor{black}{The population-level fields are unknown, but the velocity and growth rate along
each prescribed conditional path are explicit. We therefore replace this ideal objective
with regression against those conditional targets.}

\noindent\textbf{Conditional path.}
Let $\bm z=(\bm w_0,\bm w_1)\in\mathcal Z=\mathcal W^2$ have the
endpoint distribution $q$ obtained in \cref{subsec:key-ot-result}.
We use the ESFM geodesic $\bm\mu_t(\bm z)=(1-t)\bm w_0+t\bm w_1$
as the mean of a Gaussian path with fixed covariance:
\begin{equation*}
    \hat\rho_t(\bm w\mid\bm z)
    =
    \mathcal N\!\left(
    \bm w\mid\bm\mu_t(\bm z),\sigma^2 I_{d+1}
    \right).
\end{equation*}
As $\sigma\to0$, the Gaussian path recovers the deterministic ESFM path
$\bm w_t=\bm\mu_t(\bm z)$.
Its conditional velocity is constant:
\[
    \textcolor{black}{\hat{\bm s}_t}(\bm w\mid\bm z)
    =\bigl(\textcolor{black}{\hat{\bm u}_t}(\bm w\mid\bm z),\textcolor{black}{\hat g_t}(\bm w\mid\bm z)\bigr)
    =\bm w_1-\bm w_0.
\]
Mixing the conditional densities over $q$ gives $\hat\rho_t$; averaging
their velocities conditional on the current state gives $(\textcolor{black}{\hat{\bm u}_t},\textcolor{black}{\hat g_t})$.
The explicit formulas are provided in \cref{app:conditional-marginal-equivalence}.

\noindent\textbf{Conditional objective.}
The computable training loss is
\begin{equation}
\begin{aligned}
    \mathcal L_{\mathrm{ESFM}}^{\mathrm{cond}}(\theta,\phi)
    &=
    \int_0^1\int_{\mathcal Z}\int_{\mathcal W}
    \Big[
    \norm{\bm v_\theta(\bm w,t)-\textcolor{black}{\hat{\bm u}_t}(\bm w\mid\bm z)}^2
    \\
    &\qquad +\lambda
    \abs{g_\phi(\bm w,t)-\textcolor{black}{\hat g_t}(\bm w\mid\bm z)}^2
    \Big]
    \hat\rho_t(\bm w\mid\bm z)q(\bm z)
    \dd\bm w\dd\bm z\dd t.
\end{aligned}
\label{eq:main-loss}
\end{equation}
Here $\lambda>0$ balances the state and log mass regression errors.
\textcolor{black}{The following result shows that conditional regression has the same parameter
gradient as the ideal marginal objective, so these tractable targets suffice for training.}
\begin{theorem}
\label{thm:conditional-marginal-esfm}
Assume that $\hat\rho_t(\bm w)>0$ for all $\bm w\in\mathcal W$ and $t\in[0,1]$, 
and that $q(\bm z)$ is independent of $\bm w$ and $t$. Then
\[
    \nabla_{\theta,\phi}
    \mathcal L_{\mathrm{ESFM}}^{\mathrm{cond}}(\theta,\phi)
    =
    \nabla_{\theta,\phi}
    \mathcal L_{\mathrm{ESFM}}^{\mathrm{marg}}(\theta,\phi).
\]
\end{theorem}
The proof is provided in \cref{app:conditional-marginal-equivalence}.

\textcolor{black}{Training thus requires only sampling the endpoint law and evaluating the prescribed
paths and their derivatives, without ODE integration. The coupling and regression targets
are linked through the same action and conditional paths.}
\begin{algorithm}[H]
\caption{ESFM conditional flow matching training}
\label{alg:main-training}
\begin{algorithmic}[1]
\STATE \textbf{Input:} endpoint law $q(\bm z)$ with $\bm z=(\bm w_0,\bm w_1)$; $\lambda,\sigma$; batch size $B$, training steps $K$, and learning rates $\eta_v,\eta_g$.
\FOR{$k=1,\ldots,K$}
    \STATE Sample $\{\bm z_b=(\bm w_{0,b},\bm w_{1,b})\}_{b=1}^B$ independently from $q(\bm z)$.
    \FOR{$b=1,\ldots,B$}
        \STATE Sample $t_b\sim\mathcal U[0,1]$ and $\bm\epsilon_b\sim\mathcal N(0,I_{d+1})$.
        \STATE $\bm w_{t,b}\gets(1-t_b)\bm w_{0,b}+t_b\bm w_{1,b}+\sigma\bm\epsilon_b$.
        \STATE $\bm s_b^\star\gets\bm w_{1,b}-\bm w_{0,b}=(\bm u_b^\star,g_b^\star)$.
    \ENDFOR
    \STATE $\widehat{\mathcal L}_{\bm u}\gets B^{-1}\sum_{b=1}^B\norm{\bm v_\theta(\bm w_{t,b},t_b)-\bm u_b^\star}^2$.
    \STATE $\widehat{\mathcal L}_{g}\gets B^{-1}\sum_{b=1}^B\abs{g_\phi(\bm w_{t,b},t_b)-g_b^\star}^2$.
    \STATE $\theta\gets\theta-\eta_v\nabla_\theta\widehat{\mathcal L}_{\bm u}$.
    \STATE $\phi\gets\phi-\eta_g\lambda\nabla_\phi\widehat{\mathcal L}_{g}$.
\ENDFOR
\STATE \textbf{Output:} the trained vector field in extended space $\bm s_{\theta,\phi}=(\bm v_\theta,g_\phi)$.
\end{algorithmic}
\end{algorithm}

Detailed comparisons with VGFM, WFR-FM, and USB are provided in
\cref{app:method-relations}. The inference
procedure is given in \cref{app:inference-details}.

\section{Numerical Results}
\label{sec:results}

We evaluate ESFM through four questions:
\textbf{Q1}: Can a prescribed path consistent with known growth prior improve
recovery of state-dependent growth dynamics?
\textbf{Q2}: Does the numerical ESFM construction recover the expected
path--coupling consistency?
\textbf{Q3}: Does the prescribed path design preserve accurate state and mass
prediction?
\textbf{Q4}: Does ESFM generalize to unseen snapshots in real single-cell data?
Implementation settings and metric definitions are provided in
\cref{app:experimental-details,app:metrics}.

\paragraph{Recovery of state-dependent growth dynamics (Q1).}
We first test whether a prescribed growth path consistent with known qualitative
growth behavior improves recovery of the underlying growth field. We use a
controlled variant of the Simulation Gene benchmark with a known
state-dependent growth field, containing a proliferative branch with growth near
$g_{\max}$ and an approximately quiescent branch with near-zero growth. ESFM and
WFR-FM are trained on the same population snapshots and evaluated against the
ground-truth growth field.

\begin{figure}[H]
    \centering
    \includegraphics[width=0.9\textwidth]
    {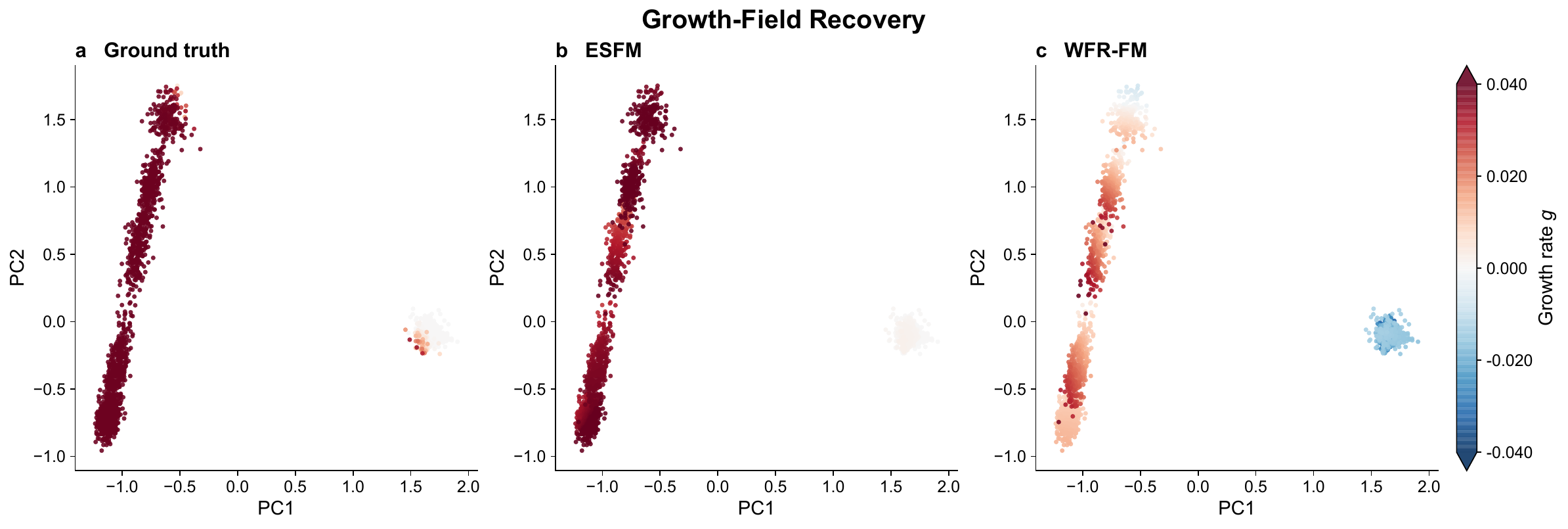}
    \caption{
    Growth field recovery on the saturating growth Simulation Gene benchmark.
    (a) Ground-truth growth rate.
    (b) ESFM prediction.
    (c) WFR-FM prediction.
    }
    \label{fig:saturating-growth-g-recovery}
\end{figure}

As shown in \cref{fig:saturating-growth-g-recovery}, ESFM recovers both the
near-zero growth of the quiescent population and the saturated growth of the
proliferative branch. Using the coefficient of determination $R^2$, ESFM
achieves $R^2=0.922$, whereas WFR-FM underestimates the growth magnitude and
yields $R^2=-0.931$. Thus, in this controlled setting, ESFM more faithfully
recovers the underlying state-dependent growth dynamics.
\textcolor{black}{Additional comparisons with VGFM, DeepRUOT and Var-RUOT, including growth-field
and trajectory recovery on held-out snapshots, are provided in
\cref{app:synthetic-benchmarks}.}

\paragraph{Numerical verification of path--coupling consistency (Q2).}
We compare the static coupling cost $A_{\mathrm{stat}}$ with the dynamic action
$A_{\mathrm{dyn}}$ of ESFM's learned dynamics. The static value is obtained by
solving the entropy-regularized static OT problem and evaluating the coupling's
cost. The dynamic value is computed by
numerical integration along ODE trajectories generated by the learned
velocity and growth fields.
Across the evaluated datasets, \cref{tab:esfm-path-coupling-consistency} shows
close agreement between the two values, indicating that the learned dynamics
approximately recover the corresponding static coupling costs. These results provide numerical support for ESFM's
path--coupling consistency. Computational
details are provided in
\cref{app:path-coupling-action}.

\begin{table}[H]
\centering
\caption{
Static coupling cost and learned dynamic action.
}
\label{tab:esfm-path-coupling-consistency}

\footnotesize
\setlength{\tabcolsep}{5pt}
\renewcommand{\arraystretch}{0.88}

\begin{tabular}{lrrr}
\toprule
Dataset & $A_{\mathrm{stat}}$ & $A_{\mathrm{dyn}}$ & Gap (\%) \\
\midrule
Gene     & 2.405    & 2.340    & 2.677 \\
Dyngen   & 29.628   & 30.187   & 1.885 \\
Gaussian & 13.343   & 13.159   & 1.374 \\
EMT      & 1.304    & 1.280    & 1.819 \\
EB       & 886.117  & 884.563  & 0.175 \\
CITE     & 3251.004 & 3228.955 & 0.678 \\
Mouse    & 310.677  & 304.636  & 1.944 \\
\bottomrule
\end{tabular}
\end{table}

\paragraph{Predictive accuracy under prescribed path design (Q3).}
We next test whether imposing the prescribed path compromises predictive
accuracy. ESFM is evaluated on Simulation Gene, Dyngen, and a 1000D Gaussian
mixture using the 1-Wasserstein distance $\mathcal W_1$ for distributional
accuracy and relative mass error (RME) for mass prediction. Additional results are provided in
\cref{app:synthetic-benchmarks}.

As shown in \cref{tab:esfm-synthetic}, ESFM achieves the best $\mathcal W_1$ and
RME on Gene and Dyngen, and the lowest RME on Gaussian while remaining
competitive in distributional accuracy. These results show that the prescribed
path design preserves accurate endpoint prediction while providing reliable mass
estimation.

\begin{table}[H]
\centering
\captionsetup{skip=3pt}
\caption{
Synthetic benchmark performance measured by mean $\mathcal W_1$ and RME.
RME is reported only for unbalanced methods.
}
\label{tab:esfm-synthetic}

\setlength{\tabcolsep}{3.5pt}
\renewcommand{\arraystretch}{0.90}

\resizebox{0.92\textwidth}{!}{
\begin{tabular}{lcccccc}
\toprule
\textbf{Method} &
\multicolumn{2}{c}{Gene (2D)} &
\multicolumn{2}{c}{Dyngen (5D)} &
\multicolumn{2}{c}{Gaussian (1000D)} \\
\cmidrule(lr){2-3} \cmidrule(lr){4-5} \cmidrule(lr){6-7}
& $\mathcal W_1$ ($\downarrow$) & RME ($\downarrow$)
& $\mathcal W_1$ ($\downarrow$) & RME ($\downarrow$)
& $\mathcal W_1$ ($\downarrow$) & RME ($\downarrow$) \\
\midrule
MMFM \citep{rohbeck2025modeling} & 0.298 & --- & 1.371 & --- & 2.833 & --- \\
SF2M \citep{sflowmatch} & 0.224{\tiny$\pm 0.007$} & --- & 1.277{\tiny$\pm 0.017$} & --- & 3.543{\tiny$\pm 0.002$} & --- \\
MIOFlow \citep{mioflow} & 0.148 & --- & 0.965 & --- & 2.858 & --- \\
DeepRUOT \citep{DeepRUOT} & 0.043{\tiny$\pm 0.002$} & 0.017{\tiny$\pm 0.001$} & 0.623{\tiny$\pm 0.032$} & 0.065{\tiny$\pm 0.011$} & 3.785{\tiny$\pm 0.009$} & 0.303{\tiny$\pm 0.070$} \\
Var-RUOT \citep{sun2025variational} & 0.079{\tiny$\pm 0.003$} & 0.008{\tiny$\pm 0.002$} & 0.522{\tiny$\pm 0.008$} & 0.177{\tiny$\pm 0.007$} & 2.813{\tiny$\pm 0.004$} & 0.041{\tiny$\pm 0.006$} \\
UOT-FM \citep{eyring2024unbalancedness} & 0.093 & 0.010 & 1.204 & 0.097 & 2.771 & 0.033 \\
VGFM \citep{wang2025joint} & 0.046 & 0.006 & 0.598 & 0.037 & 3.010 & 0.037 \\
WFR-FM \citep{peng2026wfrfm} & \second{0.019} & \second{0.001} & 0.135 & \second{0.005} & 2.233 & 0.044 \\
USB \citep{ying2026beyond} & \second{0.019}{\tiny$\pm 0.000$} & 0.002{\tiny$\pm 0.000$} & \second{0.131}{\tiny$\pm 0.001$} & 0.007{\tiny$\pm 0.000$} & \best{2.136}{\tiny$\pm 0.002$} & \second{0.004}{\tiny$\pm 0.004$} \\
 \textcolor{black}{OT-MFM \citep{kapusniak2024metric}} & \textcolor{black}{0.311} & \textcolor{black}{---} & \textcolor{black}{1.767} & \textcolor{black}{---} & \textcolor{black}{3.794} & \textcolor{black}{---} \\
 \textcolor{black}{WLF-UOT \citep{tong_action}} & \textcolor{black}{0.618} & \textcolor{black}{---} & \textcolor{black}{2.056} & \textcolor{black}{---} & \textcolor{black}{4.101} & \textcolor{black}{---} \\
 \textcolor{black}{OTP-FM \citep{kansal2026otpfm}} & \textcolor{black}{0.433} & \textcolor{black}{---} & \textcolor{black}{1.142} & \textcolor{black}{---} & \textcolor{black}{\second{2.139}} & \textcolor{black}{---} \\
\textbf{ESFM (ours)} & \best{0.017} & \best{0.0005} & \best{0.111} & \best{0.003} & 2.428 & \best{0.0001} \\
\bottomrule
\end{tabular}
}
\end{table}

\begin{figure}[t]
    \centering
    \includegraphics[width=0.82\textwidth]{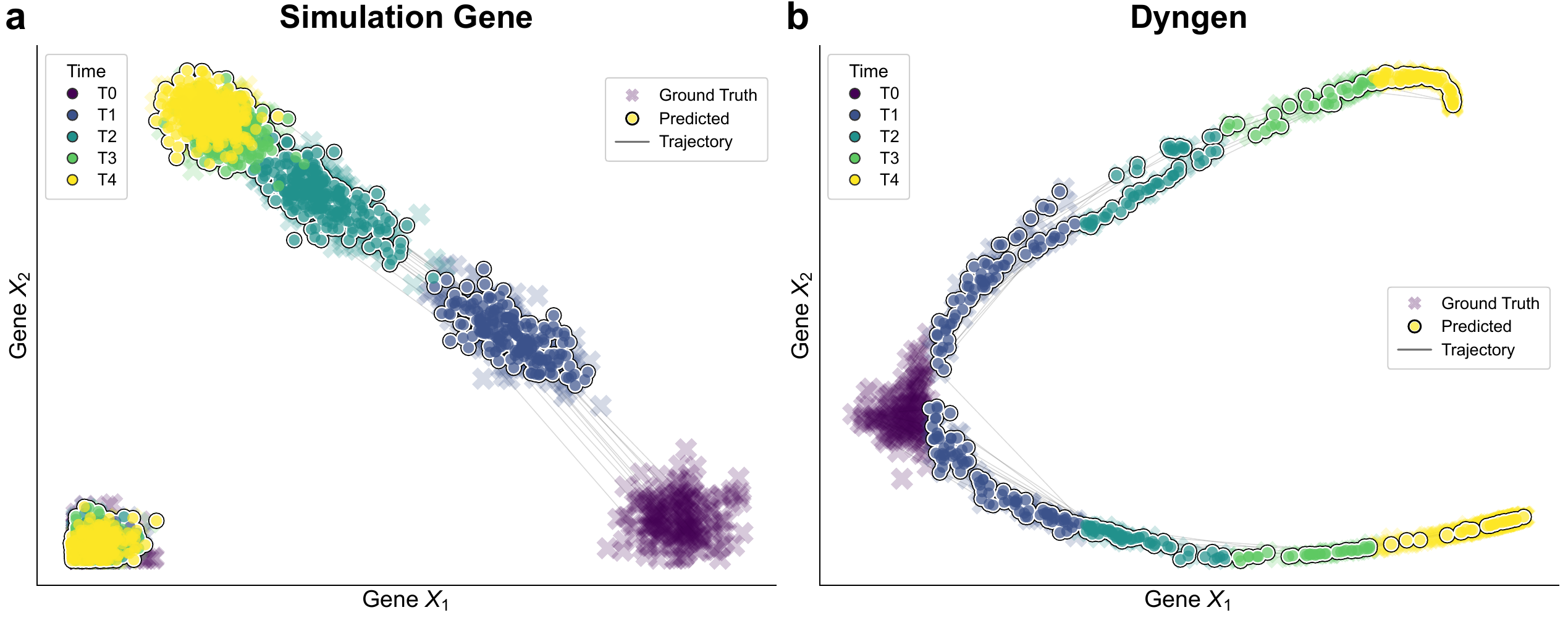}
    \caption{
    Qualitative trajectory reconstruction by ESFM on synthetic benchmarks.
    Crosses denote observed target samples, circles denote ESFM predictions,
    and gray curves denote propagated source trajectories.
    }
    \label{fig:esfm-synthetic-trajectories}
\end{figure}

\cref{fig:esfm-synthetic-trajectories} visualizes the learned dynamics on Gene
and Dyngen. The predicted endpoint distributions closely match the observed
targets, while the propagated source trajectories illustrate the learned state
evolution.

\paragraph{Generalization to held-out single-cell snapshots (Q4).}
Finally, we evaluate ESFM on unseen snapshots using hold-one-out interpolation
on EMT, embryoid body (EB), CITE-seq, and mouse hematopoiesis. We report the
mean 1-Wasserstein distance $\mathcal W_1$ over held-out time points. \textcolor{black}{Additional
results, including held-out growth maps and trajectory comparisons with
WFR-FM, VGFM, DeepRUOT and Var-RUOT, are provided in
\cref{app:real-benchmarks,app:eb100-efficiency}.}

As shown in \cref{tab:esfm-real}, ESFM achieves the lowest reported mean on EMT
and the second-lowest on EB and Mouse. These results
show that ESFM maintains competitive distributional prediction and generalizes
well to unseen snapshots across diverse real single-cell systems.

{\textcolor{black}%
{Computational efficiency on the 100D EB dataset, including training time and memory, is reported in Appendix~\ref{app:eb100-efficiency} (Figure~\ref{fig:esfm-efficiency}).}
}%
\begin{table}[t]
\centering
\captionsetup{skip=3pt}
\caption{
Hold-one-out interpolation performance on real single-cell datasets.
}
\label{tab:esfm-real}

\setlength{\tabcolsep}{4pt}
\renewcommand{\arraystretch}{0.88}

\resizebox{0.70\textwidth}{!}{
\begin{tabular}{lcccc}
\toprule
\textbf{Method}
& EMT (10D)
& EB (50D)
& CITE (50D)
& Mouse (50D) \\
\midrule
MMFM & 0.323 & 11.213 & 38.521 & 8.263 \\
SF2M & 0.308{\tiny$\pm 0.001$} & 10.986{\tiny$\pm 0.006$} & 38.333{\tiny$\pm 0.002$} & 8.646{\tiny$\pm 0.004$} \\
MIOFlow & 0.325 & 10.960 & 39.574 & 7.779 \\
TIGON & 0.360 & 11.080 & 38.159 & 6.868 \\
DeepRUOT & 0.323{\tiny$\pm 0.002$} & \best{10.075}{\tiny$\pm 0.004$} & 37.892{\tiny$\pm 0.002$} & 6.847{\tiny$\pm 0.003$} \\
Var-RUOT & 0.320{\tiny$\pm 0.003$} & 11.035{\tiny$\pm 0.017$} & 38.393{\tiny$\pm 0.029$} & 8.672{\tiny$\pm 0.040$} \\
UOT-FM & 0.322 & 11.344 & 38.649 & 9.332 \\
VGFM & 0.301 & 10.370 & 37.386 & 8.496 \\
WFR-FM & \second{0.298} & 10.157 & 37.221 & \best{6.586} \\
USB & \second{0.298}{\tiny$\pm 0.0001$} & 10.177{\tiny$\pm 0.0001$} & 37.087{\tiny$\pm 0.0001$} & 6.988{\tiny$\pm 0.0001$} \\
 \textcolor{black}{OT-MFM} & \textcolor{black}{0.314} & \textcolor{black}{10.726} & \textcolor{black}{\best{36.217}} & \textcolor{black}{7.753} \\
 \textcolor{black}{WLF-UOT} & \textcolor{black}{0.314} & \textcolor{black}{11.616} & \textcolor{black}{\second{36.990}} & \textcolor{black}{9.040} \\
 \textcolor{black}{OTP-FM} & \textcolor{black}{0.315} & \textcolor{black}{10.953} & \textcolor{black}{37.160} & \textcolor{black}{8.286} \\
\textbf{ESFM (ours)} & \best{0.295} & \second{10.134} & 37.147 & \second{6.714} \\
\bottomrule
\end{tabular}
}
\end{table}

\section{Conclusion, Limitations, and Discussion}
\label{sec:discussion}

We introduced ESFM, an extended space flow matching framework for unbalanced
dynamics from population snapshots. ESFM lifts nonconservative dynamics to
conservative transport in a joint space of state and log mass, where state
transport and relative growth are modeled jointly. For the linear state and
log mass paths studied here, we use a compatible quadratic action and derive
the corresponding static transport problem. The terminal log mass coordinate
can be eliminated analytically, yielding a tractable reduced coupling problem
and a simulation-free flow matching objective. Experiments demonstrate improved
recovery of state-dependent growth dynamics while maintaining competitive
predictive performance.

Several limitations remain. The prescribed path is an inductive prior rather
than a uniquely identifiable biological trajectory, and solving the static
transport problem can be costly at scale. Minibatch and entropic approximations
improve efficiency but require further study. Future work could explore more biologically informed path and action designs
within the extended space framework.

\section*{Acknowledgements}
This work was supported by the National Key R\&D Program of China (No.\ 2021YFA1003301 to T.L.) and the National Natural Science Foundation of China (NSFC No.\ 12288101 to T.L.). We acknowledge support from the High-performance Computing Platform of Peking University for computation.

\bibliographystyle{unsrtnat}
\bibliography{example_paper}

\begin{thebibliography}{52}
\providecommand{\natexlab}[1]{#1}
\providecommand{\url}[1]{\texttt{#1}}
\expandafter\ifx\csname urlstyle\endcsname\relax
  \providecommand{\doi}[1]{doi: #1}\else
  \providecommand{\doi}{doi: \begingroup \urlstyle{rm}\Url}\fi

\bibitem[Zheng et~al.(2017)Zheng, Terry, Belgrader, Ryvkin, Bent, Wilson,
  Ziraldo, Wheeler, McDermott, Zhu, et~al.]{zheng2017massively}
Grace X.~Y. Zheng, Jessica~M. Terry, Phillip Belgrader, Paul Ryvkin, Zachary~W.
  Bent, Ryan Wilson, Solongo~B. Ziraldo, Tobias~D. Wheeler, Geoff~P. McDermott,
  Junjie Zhu, et~al.
\newblock Massively parallel digital transcriptional profiling of single cells.
\newblock \emph{Nature Communications}, 8\penalty0 (1):\penalty0 14049, 2017.
\newblock \doi{10.1038/ncomms14049}.

\bibitem[Bunne et~al.(2024)Bunne, Schiebinger, Krause, Regev, and
  Cuturi]{bunne2024optimal}
Charlotte Bunne, Geoffrey Schiebinger, Andreas Krause, Aviv Regev, and Marco
  Cuturi.
\newblock Optimal transport for single-cell and spatial omics.
\newblock \emph{Nature Reviews Methods Primers}, 4\penalty0 (1):\penalty0 58,
  2024.
\newblock \doi{10.1038/s43586-024-00334-2}.

\bibitem[Zhang et~al.(2025{\natexlab{a}})Zhang, Sun, Peng, Li, and
  Zhou]{zhang2025review}
Zhenyi Zhang, Yuhao Sun, Qiangwei Peng, Tiejun Li, and Peijie Zhou.
\newblock Integrating dynamical systems modeling with spatiotemporal
  {scRNA-Seq} data analysis.
\newblock \emph{Entropy}, 27\penalty0 (5):\penalty0 453, 2025{\natexlab{a}}.
\newblock \doi{10.3390/e27050453}.

\bibitem[Schiebinger et~al.(2019)Schiebinger, Shu, Tabaka, Cleary, Subramanian,
  Solomon, Gould, Liu, Lin, Berube, Lee, Chen, Brumbaugh, Rigollet,
  Hochedlinger, Jaenisch, Regev, and Lander]{waddingot}
Geoffrey Schiebinger, Jian Shu, Marcin Tabaka, Brian Cleary, Vidya Subramanian,
  Aryeh Solomon, Joshua Gould, Siyan Liu, Stacie Lin, Peter Berube, Lia Lee,
  Jenny Chen, Justin Brumbaugh, Philippe Rigollet, Konrad Hochedlinger, Rudolf
  Jaenisch, Aviv Regev, and Eric~S. Lander.
\newblock Optimal-transport analysis of single-cell gene expression identifies
  developmental trajectories in reprogramming.
\newblock \emph{Cell}, 176\penalty0 (4):\penalty0 928--943.e22, 2019.
\newblock \doi{10.1016/j.cell.2019.01.006}.

\bibitem[Tong et~al.(2020)Tong, Huang, Wolf, van Dijk, and
  Krishnaswamy]{trajectorynet}
Alexander Tong, Jessie Huang, Guy Wolf, David van Dijk, and Smita Krishnaswamy.
\newblock {TrajectoryNet}: A dynamic optimal transport network for modeling
  cellular dynamics.
\newblock In \emph{Proceedings of the 37th International Conference on Machine
  Learning}, volume 119 of \emph{Proceedings of Machine Learning Research},
  pages 9526--9536, 2020.
\newblock URL \url{https://proceedings.mlr.press/v119/tong20a.html}.

\bibitem[Chen et~al.(2018)Chen, Rubanova, Bettencourt, and
  Duvenaud]{chen2018neural}
Ricky T.~Q. Chen, Yulia Rubanova, Jesse Bettencourt, and David~K. Duvenaud.
\newblock Neural ordinary differential equations.
\newblock In \emph{Advances in Neural Information Processing Systems},
  volume~31, pages 6571--6583, 2018.
\newblock URL
  \url{https://proceedings.neurips.cc/paper_files/paper/2018/hash/69386f6bb1dfed68692a24c8686939b9-Abstract.html}.

\bibitem[Huguet et~al.(2022)Huguet, Magruder, Tong, Fasina, Kuchroo, Wolf, and
  Krishnaswamy]{mioflow}
Guillaume Huguet, Daniel~Sumner Magruder, Alexander Tong, Oluwadamilola Fasina,
  Manik Kuchroo, Guy Wolf, and Smita Krishnaswamy.
\newblock Manifold interpolating optimal-transport flows for trajectory
  inference.
\newblock In \emph{Advances in Neural Information Processing Systems},
  volume~35, pages 29705--29718, 2022.
\newblock \doi{10.52202/068431-2154}.

\bibitem[Benamou and Brenier(2000)]{benamou2000computational}
Jean-David Benamou and Yann Brenier.
\newblock A computational fluid mechanics solution to the {Monge--Kantorovich}
  mass transfer problem.
\newblock \emph{Numerische Mathematik}, 84\penalty0 (3):\penalty0 375--393,
  2000.
\newblock \doi{10.1007/s002110050002}.

\bibitem[Lipman et~al.(2023)Lipman, Chen, Ben-Hamu, Nickel, and Le]{cfm_lipman}
Yaron Lipman, Ricky T.~Q. Chen, Heli Ben-Hamu, Maximilian Nickel, and Matthew
  Le.
\newblock Flow matching for generative modeling.
\newblock In \emph{The Eleventh International Conference on Learning
  Representations}, 2023.
\newblock URL \url{https://openreview.net/forum?id=PqvMRDCJT9t}.

\bibitem[Liu et~al.(2023)Liu, Gong, and Liu]{liu2022flow}
Xingchao Liu, Chengyue Gong, and Qiang Liu.
\newblock Flow straight and fast: Learning to generate and transfer data with
  rectified flow.
\newblock In \emph{The Eleventh International Conference on Learning
  Representations}, 2023.
\newblock URL \url{https://openreview.net/forum?id=XVjTT1nw5z}.

\bibitem[Albergo and Vanden-Eijnden(2023)]{albergo2022building}
Michael~S. Albergo and Eric Vanden-Eijnden.
\newblock Building normalizing flows with stochastic interpolants.
\newblock In \emph{The Eleventh International Conference on Learning
  Representations}, 2023.
\newblock URL \url{https://openreview.net/forum?id=li7qeBbCR1t}.

\bibitem[Klein et~al.(2024)Klein, Uscidda, Theis, and Cuturi]{klein2024genot}
Dominik Klein, Th{\'e}o Uscidda, Fabian Theis, and Marco Cuturi.
\newblock {GENOT}: Entropic ({Gromov}) {Wasserstein} flow matching with
  applications to single-cell genomics.
\newblock In \emph{Advances in Neural Information Processing Systems},
  volume~37, pages 103897--103944, 2024.
\newblock \doi{10.52202/079017-3301}.

\bibitem[Rohbeck et~al.(2025)Rohbeck, De~Brouwer, Bunne, Huetter, Biton, Chen,
  Regev, and Lopez]{rohbeck2025modeling}
Martin Rohbeck, Edward De~Brouwer, Charlotte Bunne, Jan-Christian Huetter, Anne
  Biton, Kelvin~Y. Chen, Aviv Regev, and Romain Lopez.
\newblock Modeling complex system dynamics with flow matching across time and
  conditions.
\newblock In \emph{The Thirteenth International Conference on Learning
  Representations}, 2025.
\newblock URL
  \url{https://proceedings.iclr.cc/paper_files/paper/2025/hash/206b764b00195b8c896c5af8bde80726-Abstract-Conference.html}.

\bibitem[Tong et~al.(2024{\natexlab{a}})Tong, Fatras, Malkin, Huguet, Zhang,
  Rector-Brooks, Wolf, and Bengio]{tong2023conditional}
Alexander Tong, Kilian Fatras, Nikolay Malkin, Guillaume Huguet, Yanlei Zhang,
  Jarrid Rector-Brooks, Guy Wolf, and Yoshua Bengio.
\newblock Improving and generalizing flow-based generative models with
  minibatch optimal transport.
\newblock \emph{Transactions on Machine Learning Research}, 2024{\natexlab{a}}.
\newblock URL \url{https://openreview.net/forum?id=CD9Snc73AW}.

\bibitem[Pooladian et~al.(2023)Pooladian, Ben-Hamu, Domingo-Enrich, Amos,
  Lipman, and Chen]{pooladian2023multisample}
Aram-Alexandre Pooladian, Heli Ben-Hamu, Carles Domingo-Enrich, Brandon Amos,
  Yaron Lipman, and Ricky T.~Q. Chen.
\newblock Multisample flow matching: Straightening flows with minibatch
  couplings.
\newblock In \emph{Proceedings of the 40th International Conference on Machine
  Learning}, volume 202 of \emph{Proceedings of Machine Learning Research},
  pages 28100--28127, 2023.
\newblock URL \url{https://proceedings.mlr.press/v202/pooladian23a.html}.

\bibitem[Sha et~al.(2024)Sha, Qiu, Zhou, and Nie]{TIGON}
Yutong Sha, Yuchi Qiu, Peijie Zhou, and Qing Nie.
\newblock Reconstructing growth and dynamic trajectories from single-cell
  transcriptomics data.
\newblock \emph{Nature Machine Intelligence}, 6\penalty0 (1):\penalty0 25--39,
  2024.
\newblock \doi{10.1038/s42256-023-00763-w}.

\bibitem[Zhang et~al.(2025{\natexlab{b}})Zhang, Li, and Zhou]{DeepRUOT}
Zhenyi Zhang, Tiejun Li, and Peijie Zhou.
\newblock Learning stochastic dynamics from snapshots through regularized
  unbalanced optimal transport.
\newblock In \emph{The Thirteenth International Conference on Learning
  Representations}, 2025{\natexlab{b}}.
\newblock URL
  \url{https://proceedings.iclr.cc/paper_files/paper/2025/hash/32b8a612105de5c22db337b774ce7b61-Abstract-Conference.html}.

\bibitem[Sun et~al.(2025)Sun, Zhang, Wang, Li, and Zhou]{sun2025variational}
Yuhao Sun, Zhenyi Zhang, Zihan Wang, Tiejun Li, and Peijie Zhou.
\newblock Variational regularized unbalanced optimal transport: Single network,
  least action.
\newblock In \emph{Advances in Neural Information Processing Systems},
  volume~38, pages 19150--19192, 2025.
\newblock \doi{10.52202/085713-0572}.

\bibitem[Peng et~al.(2026{\natexlab{a}})Peng, Zhou, and Li]{peng2026stvcr}
Qiangwei Peng, Peijie Zhou, and Tiejun Li.
\newblock {stVCR}: Spatiotemporal dynamics of single cells.
\newblock \emph{Nature Methods}, 23\penalty0 (3):\penalty0 542--553,
  2026{\natexlab{a}}.
\newblock \doi{10.1038/s41592-026-03010-3}.

\bibitem[Liero et~al.(2016)Liero, Mielke, and Savar{\'e}]{liero2016optimal}
Matthias Liero, Alexander Mielke, and Giuseppe Savar{\'e}.
\newblock Optimal transport in competition with reaction: The
  {Hellinger--Kantorovich} distance and geodesic curves.
\newblock \emph{SIAM Journal on Mathematical Analysis}, 48\penalty0
  (4):\penalty0 2869--2911, 2016.
\newblock \doi{10.1137/15M1041420}.

\bibitem[Chizat et~al.(2018{\natexlab{a}})Chizat, Peyr{\'e}, Schmitzer, and
  Vialard]{chizat2018interpolating}
L{\'e}na{\"i}c Chizat, Gabriel Peyr{\'e}, Bernhard Schmitzer, and
  Fran{\c{c}}ois-Xavier Vialard.
\newblock An interpolating distance between optimal transport and {Fisher--Rao}
  metrics.
\newblock \emph{Foundations of Computational Mathematics}, 18\penalty0
  (1):\penalty0 1--44, 2018{\natexlab{a}}.
\newblock \doi{10.1007/s10208-016-9331-y}.

\bibitem[Chizat et~al.(2018{\natexlab{b}})Chizat, Peyr{\'e}, Schmitzer, and
  Vialard]{chizat2018unbalanced}
L{\'e}na{\"i}c Chizat, Gabriel Peyr{\'e}, Bernhard Schmitzer, and
  Fran{\c{c}}ois-Xavier Vialard.
\newblock Unbalanced optimal transport: Dynamic and {Kantorovich} formulations.
\newblock \emph{Journal of Functional Analysis}, 274\penalty0 (11):\penalty0
  3090--3123, 2018{\natexlab{b}}.
\newblock \doi{10.1016/j.jfa.2018.03.008}.

\bibitem[Eyring et~al.(2024)Eyring, Klein, Uscidda, Palla, Kilbertus, Akata,
  and Theis]{eyring2024unbalancedness}
Luca Eyring, Dominik Klein, Th{\'e}o Uscidda, Giovanni Palla, Niki Kilbertus,
  Zeynep Akata, and Fabian~J. Theis.
\newblock Unbalancedness in neural {Monge} maps improves unpaired domain
  translation.
\newblock In \emph{The Twelfth International Conference on Learning
  Representations}, 2024.
\newblock URL
  \url{https://proceedings.iclr.cc/paper_files/paper/2024/hash/022a39052abf9ca467e268923057dfc0-Abstract-Conference.html}.

\bibitem[Wang et~al.(2025)Wang, Jiang, Zhang, Gu, Zhou, and Sun]{wang2025joint}
Dongyi Wang, Yuanwei Jiang, Zhenyi Zhang, Xiang Gu, Peijie Zhou, and Jian Sun.
\newblock Joint velocity-growth flow matching for single-cell dynamics
  modeling.
\newblock In \emph{Advances in Neural Information Processing Systems},
  volume~38, pages 178180--178216, 2025.
\newblock \doi{10.52202/085713-5363}.

\bibitem[Peng et~al.(2026{\natexlab{b}})Peng, Wang, Ying, Sun, Nie, Zhang, Li,
  and Zhou]{peng2026wfrfm}
Qiangwei Peng, Zihan Wang, Junda Ying, Yuhao Sun, Qing Nie, Lei Zhang, Tiejun
  Li, and Peijie Zhou.
\newblock {WFR-FM}: Simulation-free dynamic unbalanced optimal transport.
\newblock In \emph{The Fourteenth International Conference on Learning
  Representations}, 2026{\natexlab{b}}.
\newblock URL \url{https://openreview.net/forum?id=1nqu7bK1mm}.

\bibitem[Ying et~al.(2026{\natexlab{a}})Ying, Wang, Yang, Zhou, and
  Zhang]{ying2026sudo}
Junda Ying, Yuxuan Wang, Bowen Yang, Peijie Zhou, and Lei Zhang.
\newblock Simulation-free unbalanced dynamic optimal transport with general
  growth penalty.
\newblock \emph{arXiv preprint arXiv:2609.04710}, 2026{\natexlab{a}}.

\bibitem[Kapusniak et~al.(2024)Kapusniak, Potaptchik, Reu, Zhang, Tong,
  Bronstein, Bose, and Di~Giovanni]{kapusniak2024metric}
Kacper Kapusniak, Peter Potaptchik, Teodora Reu, Leo Zhang, Alexander Tong,
  Michael Bronstein, Joey Bose, and Francesco Di~Giovanni.
\newblock Metric flow matching for smooth interpolations on the data manifold.
\newblock In \emph{Advances in Neural Information Processing Systems},
  volume~37, pages 135011--135042, 2024.
\newblock \doi{10.52202/079017-4291}.

\bibitem[Lee et~al.(2025)Lee, Moradijamei, and Shakeri]{lee2025mmsfm}
Justin Lee, Behnaz Moradijamei, and Heman Shakeri.
\newblock Multi-marginal stochastic flow matching for high-dimensional snapshot
  data at irregular time points.
\newblock In \emph{Proceedings of the 42nd International Conference on Machine
  Learning}, volume 267 of \emph{Proceedings of Machine Learning Research},
  pages 33476--33498, 2025.
\newblock URL \url{https://proceedings.mlr.press/v267/lee25x.html}.

\bibitem[Kviman et~al.(2026)Kviman, Tamogashev, Branchini, Elvira, Lagergren,
  and Malkin]{kviman2026alicfm}
Oskar Kviman, Kirill Tamogashev, Nicola Branchini, V{\'i}ctor Elvira, Jens
  Lagergren, and Nikolay Malkin.
\newblock Multi-marginal flow matching with adversarially learnt interpolants.
\newblock In \emph{The Fourteenth International Conference on Learning
  Representations}, 2026.

\bibitem[Pu et~al.(2026)Pu, Long, Huang, Chen, Wang, Xiao, Luo, Zhu, Zhou, and
  Wang]{pu2026snapshots}
Siyu Pu, Qingqing Long, Xiaohan Huang, Haotian Chen, Jiajia Wang, Meng Xiao,
  Xiao Luo, Hengshu Zhu, Yuanchun Zhou, and Xuezhi Wang.
\newblock From snapshots to trajectories: Learning single-cell gene expression
  dynamics via conditional flow matching.
\newblock \emph{arXiv preprint arXiv:2605.22340}, 2026.

\bibitem[Albergo et~al.(2024)Albergo, Goldstein, Boffi, Ranganath, and
  Vanden-Eijnden]{albergo2024datadependent}
Michael~Samuel Albergo, Mark Goldstein, Nicholas~Matthew Boffi, Rajesh
  Ranganath, and Eric Vanden-Eijnden.
\newblock Stochastic interpolants with data-dependent couplings.
\newblock In \emph{Proceedings of the 41st International Conference on Machine
  Learning}, volume 235 of \emph{Proceedings of Machine Learning Research},
  pages 921--937, 2024.
\newblock URL \url{https://proceedings.mlr.press/v235/albergo24a.html}.

\bibitem[Kantorovich(1942)]{kantorovich1942translocation}
Leonid~V. Kantorovich.
\newblock On the translocation of masses.
\newblock \emph{Doklady Akademii Nauk SSSR}, 37:\penalty0 199--201, 1942.

\bibitem[Tong et~al.(2024{\natexlab{b}})Tong, Malkin, Fatras, Atanackovic,
  Zhang, Huguet, Wolf, and Bengio]{sflowmatch}
Alexander~Y. Tong, Nikolay Malkin, Kilian Fatras, Lazar Atanackovic, Yanlei
  Zhang, Guillaume Huguet, Guy Wolf, and Yoshua Bengio.
\newblock Simulation-free {Schr{\"o}dinger} bridges via score and flow
  matching.
\newblock In \emph{Proceedings of the 27th International Conference on
  Artificial Intelligence and Statistics}, volume 238 of \emph{Proceedings of
  Machine Learning Research}, pages 1279--1287, 2024{\natexlab{b}}.
\newblock URL \url{https://proceedings.mlr.press/v238/tong24a.html}.

\bibitem[Du et~al.(2026)Du, Zhang, and Li]{du2026lagrangian}
Shukai Du, Junzhe Zhang, and Yiming Li.
\newblock Lagrangian flow matching: A least-action framework for principled
  path design.
\newblock \emph{arXiv preprint arXiv:2605.15419}, 2026.

\bibitem[Benamou(2003)]{benamou2003numerical}
Jean-David Benamou.
\newblock Numerical resolution of an unbalanced mass transport problem.
\newblock \emph{ESAIM: Mathematical Modelling and Numerical Analysis},
  37\penalty0 (5):\penalty0 851--868, 2003.
\newblock \doi{10.1051/m2an:2003058}.

\bibitem[Kondratyev et~al.(2016)Kondratyev, Monsaingeon, and
  Vorotnikov]{kondratyev2016}
Stanislav Kondratyev, L{\'e}onard Monsaingeon, and Dmitry Vorotnikov.
\newblock A new optimal transport distance on the space of finite {Radon}
  measures.
\newblock \emph{Advances in Differential Equations}, 21\penalty0
  (11/12):\penalty0 1117--1164, 2016.
\newblock \doi{10.57262/ade/1476369298}.

\bibitem[Sun et~al.(2026)Sun, Rocha, Phelan, Bhaskar, Huguet, Zhang, Magruder,
  Tong, Xu, Fasina, Gerstein, Wolf, Ivanova, Chaffer, and
  Krishnaswamy]{sun2026mioflow2}
Xingzhi Sun, Jo{\~a}o~Felipe Rocha, Brett Phelan, Dhananjay Bhaskar, Guillaume
  Huguet, Yanlei Zhang, Daniel~Sumner Magruder, Alexander Tong, Ke~Xu,
  Oluwadamilola Fasina, Mark Gerstein, Guy Wolf, Natalia Ivanova, Christine~L.
  Chaffer, and Smita Krishnaswamy.
\newblock {MIOFlow} 2.0: A unified framework for inferring cellular stochastic
  dynamics from single cell and spatial transcriptomics data.
\newblock \emph{arXiv preprint arXiv:2603.22564}, 2026.

\bibitem[Maddu et~al.(2026)Maddu, Chard{\`e}s, and
  Shelley]{maddu2026probability}
Suryanarayana Maddu, Victor Chard{\`e}s, and Michael~J. Shelley.
\newblock Learning biophysical models of gene regulation with probability flow
  matching.
\newblock \emph{arXiv preprint arXiv:2604.25062}, 2026.
\newblock URL \url{https://arxiv.org/abs/2604.25062}.

\bibitem[Ying et~al.(2026{\natexlab{b}})Ying, Wang, Yang, Zhou, and
  Zhang]{ying2026beyond}
Junda Ying, Yuxuan Wang, Bowen Yang, Peijie Zhou, and Lei Zhang.
\newblock Beyond continuity: Simulation-free reconstruction of discrete
  branching dynamics from single-cell snapshots.
\newblock In \emph{Forty-Third International Conference on Machine Learning},
  2026{\natexlab{b}}.
\newblock URL \url{https://openreview.net/forum?id=IABjbJlinz}.

\bibitem[Peng et~al.(2026{\natexlab{c}})Peng, Chen, and Zhou]{peng2026mustfm}
Qiangwei Peng, Lezhi Chen, and Peijie Zhou.
\newblock Multiscale supervised unbalanced optimal transport flow matching.
\newblock \emph{arXiv preprint arXiv:2605.16529}, 2026{\natexlab{c}}.
\newblock URL \url{https://arxiv.org/abs/2605.16529}.

\bibitem[Wang et~al.(2026)Wang, Wang, Peng, Zhou, and Li]{wang2026wfrmfm}
Xinyu Wang, Ruoyu Wang, Qiangwei Peng, Peijie Zhou, and Tiejun Li.
\newblock {WFR-MFM}: One-step inference for dynamic unbalanced optimal
  transport.
\newblock In \emph{Forty-Third International Conference on Machine Learning},
  2026.
\newblock URL \url{https://openreview.net/forum?id=yVGbbB3ERW}.

\bibitem[Neklyudov et~al.(2024)Neklyudov, Brekelmans, Tong, Atanackovic, Liu,
  and Makhzani]{tong_action}
Kirill Neklyudov, Rob Brekelmans, Alexander Tong, Lazar Atanackovic, Qiang Liu,
  and Alireza Makhzani.
\newblock A computational framework for solving {Wasserstein} {Lagrangian}
  flows.
\newblock In \emph{Proceedings of the 41st International Conference on Machine
  Learning}, volume 235 of \emph{Proceedings of Machine Learning Research},
  pages 37461--37485, 2024.
\newblock URL \url{https://proceedings.mlr.press/v235/neklyudov24a.html}.

\bibitem[Kansal et~al.(2026)Kansal, Crair, Nguyen, Pope, and
  Parry]{kansal2026otpfm}
Raghav Kansal, David Crair, Nghia Nguyen, Scott Pope, and Bradley Parry.
\newblock Multimarginal flow matching with optimal transport potentials.
\newblock In \emph{International Conference on Machine Learning}, 2026.
\newblock URL \url{https://arxiv.org/abs/2606.05327}.

\bibitem[Liero et~al.(2018)Liero, Mielke, and Savar{\'e}]{liero2018optimal}
Matthias Liero, Alexander Mielke, and Giuseppe Savar{\'e}.
\newblock Optimal entropy-transport problems and a new {Hellinger--Kantorovich}
  distance between positive measures.
\newblock \emph{Inventiones Mathematicae}, 211\penalty0 (3):\penalty0
  969--1117, 2018.
\newblock \doi{10.1007/s00222-017-0759-8}.

\bibitem[Klein et~al.(2025)Klein, Fleck, Bobrovskiy, Zimmermann, Becker, Palma,
  Dony, Tejada-Lapuerta, Huguet, Lin, Azbukina, Sanch{\'i}s-Calleja, Uscidda,
  Szalata, Gander, Regev, Treutlein, Camp, and Theis]{klein2025cellflow}
Dominik Klein, Jonas~Simon Fleck, Daniil Bobrovskiy, Lea Zimmermann, S{\"o}ren
  Becker, Alessandro Palma, Leander Dony, Alejandro Tejada-Lapuerta, Guillaume
  Huguet, Hsiu-Chuan Lin, Nadezhda Azbukina, F{\'a}tima Sanch{\'i}s-Calleja,
  Th{\'e}o Uscidda, Artur Szalata, Manuel Gander, Aviv Regev, Barbara
  Treutlein, J.~Gray Camp, and Fabian~J. Theis.
\newblock {CellFlow} enables generative single-cell phenotype modeling with
  flow matching.
\newblock \emph{bioRxiv}, 2025.
\newblock \doi{10.1101/2025.04.11.648220}.
\newblock URL \url{https://www.biorxiv.org/content/10.1101/2025.04.11.648220}.

\bibitem[Paszke et~al.(2017)Paszke, Gross, Chintala, Chanan, Yang, DeVito, Lin,
  Desmaison, Antiga, and Lerer]{pytorch}
Adam Paszke, Sam Gross, Soumith Chintala, Gregory Chanan, Edward Yang, Zachary
  DeVito, Zeming Lin, Alban Desmaison, Luca Antiga, and Adam Lerer.
\newblock Automatic differentiation in {PyTorch}.
\newblock In \emph{NIPS 2017 Workshop on Autodiff}, 2017.
\newblock URL \url{https://openreview.net/forum?id=BJJsrmfCZ}.

\bibitem[Cannoodt et~al.(2021)Cannoodt, Saelens, Deconinck, and
  Saeys]{cannoodt2021spearheading}
Robrecht Cannoodt, Wouter Saelens, Louise Deconinck, and Yvan Saeys.
\newblock Spearheading future omics analyses using {dyngen}, a multi-modal
  simulator of single cells.
\newblock \emph{Nature Communications}, 12\penalty0 (1):\penalty0 3942, 2021.
\newblock \doi{10.1038/s41467-021-24152-2}.

\bibitem[Moon et~al.(2019)Moon, van Dijk, Wang, Gigante, Burkhardt, Chen, Yim,
  van~den Elzen, Hirn, Coifman, Ivanova, Wolf, and
  Krishnaswamy]{moon2019visualizing}
Kevin~R. Moon, David van Dijk, Zheng Wang, Scott Gigante, Daniel~B. Burkhardt,
  William~S. Chen, Kristina Yim, Antonia van~den Elzen, Matthew~J. Hirn,
  Ronald~R. Coifman, Natalia~B. Ivanova, Guy Wolf, and Smita Krishnaswamy.
\newblock Visualizing structure and transitions in high-dimensional biological
  data.
\newblock \emph{Nature Biotechnology}, 37\penalty0 (12):\penalty0 1482--1492,
  2019.
\newblock \doi{10.1038/s41587-019-0336-3}.

\bibitem[Cook and Vanderhyden(2020)]{cook2020context}
David~P. Cook and Barbara~C. Vanderhyden.
\newblock Context specificity of the {EMT} transcriptional response.
\newblock \emph{Nature Communications}, 11\penalty0 (1):\penalty0 2142, 2020.
\newblock \doi{10.1038/s41467-020-16066-2}.

\bibitem[Lance et~al.(2022)Lance, Luecken, Burkhardt, Cannoodt, Rautenstrauch,
  Laddach, Ubingazhibov, Cao, Deng, Khan, Liu, Russkikh, Ryazantsev, Ohler,
  {{NeurIPS 2021 Multimodal Data Integration Competition Participants}}, Pisco,
  Bloom, Krishnaswamy, and Theis]{lance2022multimodal}
Christopher Lance, Malte~D. Luecken, Daniel~B. Burkhardt, Robrecht Cannoodt,
  Pia Rautenstrauch, Anna Laddach, Aidyn Ubingazhibov, Zhi-Jie Cao, Kaiwen
  Deng, Sumeer Khan, Qiao Liu, Nikol Russkikh, Gleb Ryazantsev, Uwe Ohler,
  {{NeurIPS 2021 Multimodal Data Integration Competition Participants}},
  Angela~Oliveira Pisco, Jonathan Bloom, Smita Krishnaswamy, and Fabian~J.
  Theis.
\newblock Multimodal single-cell data integration challenge: Results and
  lessons learned.
\newblock In \emph{Proceedings of the NeurIPS 2021 Competitions and
  Demonstrations Track}, volume 176 of \emph{Proceedings of Machine Learning
  Research}, pages 162--176, 2022.
\newblock URL \url{https://proceedings.mlr.press/v176/lance22a.html}.

\bibitem[Weinreb et~al.(2020)Weinreb, Rodriguez-Fraticelli, Camargo, and
  Klein]{weinreb2020lineage}
Caleb Weinreb, Alejo Rodriguez-Fraticelli, Fernando~D. Camargo, and Allon~M.
  Klein.
\newblock Lineage tracing on transcriptional landscapes links state to fate
  during differentiation.
\newblock \emph{Science}, 367\penalty0 (6479):\penalty0 eaaw3381, 2020.
\newblock \doi{10.1126/science.aaw3381}.

\bibitem[Baradat and Lavenant(2025)]{branRUOT}
Aymeric Baradat and Hugo Lavenant.
\newblock \emph{Regularized Unbalanced Optimal Transport as Entropy
  Minimization with Respect to Branching {Brownian} Motion}, volume 458 of
  \emph{Ast{\'e}risque}.
\newblock Soci{\'e}t{\'e} Math{\'e}matique de France, 2025.
\newblock ISBN 978-2-37905-211-8.
\newblock \doi{10.24033/ast.1247}.

\end{thebibliography}

\appendix
\crefalias{section}{appendix}
\crefalias{subsection}{appendix}
\crefalias{subsubsection}{appendix}

\newpage
\section{Mathematical Derivations and Proofs}
\label{app:proofs}
\subsection{Additional Background on Flow Matching and Dynamic OT}
\label{app:background-fm-ot}

This section provides additional background, including marginal flow matching and the
dynamic--static correspondence underlying OT-based conditional flow matching.

\paragraph{Flow matching.}
Consider normalized endpoint distributions
$\rho_0,\rho_1\in\mathcal P(\R^d)$. Flow matching learns a time-dependent
vector field whose ODE transports samples from $\rho_0$ to $\rho_1$:
\begin{equation}
    \frac{\dd \bm{x}_t}{\dd t}
    =
    \bm u_t(\bm{x}_t).
    \label{eq:appendix-background-ode}
\end{equation}
Given a probability path $\{\rho_t\}_{t\in[0,1]}$ and its velocity field
$\bm u_t$, the marginal flow-matching objective is
\begin{equation*}
    \mathcal L_{\mathrm{FM}}(\theta)
    =
    \mathbb E_{t,\bm{x}\sim\rho_t}
    \left[
    \norm{
    \bm u_\theta(\bm{x},t)-\bm u_t(\bm{x})
    }^2
    \right].
\end{equation*}
The learned vector field is used for inference by integrating
\cref{eq:appendix-background-ode}. In practice, the marginal path
$\rho_t$ and marginal velocity $\bm u_t$ are generally unavailable in closed form,
which motivates the conditional formulation used in
\cref{eq:background-cfm}.

\paragraph{Dynamic optimal transport.}
In the balanced setting, a density $\rho_t$ and velocity field $\bm u_t$
satisfy the continuity equation
\begin{equation*}
    \partial_t\rho_t
    +
    \nabla\cdot(\rho_t\bm u_t)
    =
    0.
\end{equation*}
The Benamou--Brenier formulation selects the interpolation of minimum kinetic
energy \citep{benamou2000computational}:
\begin{equation*}
\begin{aligned}
    W_2^2(\rho_0,\rho_1)
    =
    \inf_{\rho,\bm u}\quad
    &
    \int_0^1
    \int_{\R^d}
    \norm{\bm u_t(\bm{x})}^2
    \rho_t(\bm{x})
    \dd\bm{x}\dd t,
    \\
    \text{s.t.}\quad
    &
    \partial_t\rho_t
    +
    \nabla\cdot(\rho_t\bm u_t)
    =
    0,
    \\
    &
    \rho_{t=0}=\rho_0,
    \qquad
    \rho_{t=1}=\rho_1.
\end{aligned}
\end{equation*}

For a pair of Dirac measures at $\bm{x}_0$ and $\bm{x}_1$, the minimizing
particle path is
\begin{equation}
    \bm{x}_t
    =
    (1-t)\bm{x}_0+t\bm{x}_1,
    \qquad
    \bm u_t
    =
    \bm{x}_1-\bm{x}_0.
    \label{eq:appendix-dirac-geodesic}
\end{equation}
Thus, the dynamic OT geometry determines the conditional path and its velocity
at the particle level.

\paragraph{Static OT and endpoint coupling.}
The Benamou--Brenier dynamic problem is equivalent to the quadratic
Kantorovich problem
\begin{equation*}
    \pi^\star
    \in
    \arg\min_{\pi\in\Pi(\rho_0,\rho_1)}
    \int_{\R^d\times\R^d}
    \norm{\bm{x}_1-\bm{x}_0}^2
    \dd\pi(\bm{x}_0,\bm{x}_1).
\end{equation*}
The static formulation determines which source and target states are paired,
whereas the corresponding dynamic formulation determines how each pair moves
between its endpoints. Aggregating the Dirac-to-Dirac geodesics in
\cref{eq:appendix-dirac-geodesic} over the optimal coupling $\pi^\star$
recovers the balanced displacement interpolation. This dynamic--static
correspondence is the geometric basis of OT-CFM
\citep{tong2023conditional}.

\paragraph{Extension to unbalanced dynamics.}
For mass-changing systems, the conservative continuity equation is replaced by
\begin{equation*}
    \partial_t\rho_t
    +
    \nabla\cdot(\rho_t\bm u_t)
    =
    \rho_t g_t,
\end{equation*}
where $g_t$ is the growth rate defined in
\cref{eq:background-growth-rate}. Dynamic unbalanced OT formulations,
including WFR, augment the transport action with a growth penalty and admit
corresponding static formulations based on relaxed couplings or semi-couplings
\citep{chizat2018interpolating,liero2018optimal}.

The same construction principle can therefore be used in unbalanced flow
matching: a dynamic transport geometry determines the single-particle state
and mass paths, while its static formulation determines the endpoint pairing.
Examples include relaxed-coupling approaches such as UOT-FM and CellFlow
\citep{eyring2024unbalancedness,klein2025cellflow}, semi-relaxed growth
estimation as in VGFM \citep{wang2025joint}, and WFR-geodesic constructions
such as WFR-FM and MUST-FM
\citep{peng2026wfrfm,peng2026mustfm}. Related methods model stochastic
branching or mechanistic gene-regulatory dynamics, including USB and PFM
\citep{ying2026beyond,maddu2026probability}.

\subsection{Proof of \texorpdfstring{\cref{thm:extended-projection}}{the Projection Theorem}}
\label{app:projection}

Let $y=\ln m$ and let $\hat\rho_t(\bm{x},y)$ be a particle number density. The physical density is
\begin{equation*}
    \rho_t(\bm{x})
    =
    \int_{\R}\e^{y}\hat\rho_t(\bm{x},y)\dd y.
\end{equation*}
If the extended velocity is $(\hat{\bm u}_t,\hat g_t)$, then the physical velocity and growth rate are weighted averages:
\begin{equation*}
\begin{gathered}
    \bm u_t(\bm{x})
    =
    \frac{
    \int_{\R}
    \hat{\bm u}_t(\bm{x},y)
    \hat\rho_t(\bm{x},y)
    \e^y\dd y
    }{
    \rho_t(\bm{x})
    },\;\;\;
    g_t(\bm{x})
    =
    \frac{
    \int_{\R}
    \hat g_t(\bm{x},y)
    \hat\rho_t(\bm{x},y)
    \e^y\dd y
    }{
    \rho_t(\bm{x})
    }.
\end{gathered}
\end{equation*}
The conservative extended equation is
\begin{equation}
    \partial_t\hat\rho_t
    +
    \nabla_{\bm x}\cdot(\hat\rho_t\hat{\bm u}_t)
    +
    \partial_y(\hat\rho_t\hat g_t)
    =
    0.
    \label{eq:appendix-extended-continuity}
\end{equation}
We now derive the physical nonconservative continuity equation in \cref{eq:physical-continuity-main}. Multiplying \cref{eq:appendix-extended-continuity} by $\e^{y}$ and integrating over $y$ gives
\begin{equation}
    \int_{\R}\e^{y}\partial_t\hat\rho_t\dd y
    +
    \int_{\R}\e^{y}\nabla_{\bm x}\cdot(\hat\rho_t\hat{\bm u}_t)\dd y
    +
    \int_{\R}\e^{y}\partial_y(\hat\rho_t\hat g_t)\dd y
    =
    0.
    \label{eq:appendix-projection-integrated}
\end{equation}
The first two terms are the projected time derivative and flux:
\begin{align}
    \int_{\R}\e^{y}\partial_t\hat\rho_t(\bm{x},y)\dd y
    &=
    \partial_t\rho_t(\bm{x}),
    \label{eq:appendix-projection-time}
    \\
    \int_{\R}\e^{y}\nabla_{\bm x}\cdot
    \left(\hat\rho_t(\bm{x},y)\hat{\bm u}_t(\bm{x},y)\right)\dd y
    &=
    \nabla_{\bm x}\cdot\left(\rho_t(\bm{x})\bm u_t(\bm{x})\right),
    \label{eq:appendix-projection-flux}
\end{align}
where the second identity follows from the definition of the weighted physical velocity.
For the $y$-transport term, integration by parts gives
\begin{equation}
\begin{aligned}
    \int_{\R}\e^{y}\partial_y
    \left(\hat\rho_t(\bm{x},y)\hat g_t(\bm{x},y)\right)\dd y
    &=
    \left[\e^{y}\hat\rho_t(\bm{x},y)\hat g_t(\bm{x},y)\right]_{-\infty}^{+\infty}
    -
    \int_{\R}\e^{y}\hat\rho_t(\bm{x},y)\hat g_t(\bm{x},y)\dd y
    \\
    &=
    -\rho_t(\bm{x})g_t(\bm{x}),
\end{aligned}
\label{eq:appendix-projection-growth}
\end{equation}
where the boundary term vanishes by the assumption in \cref{thm:extended-projection}.
Since $y=\ln m$, the two limits correspond to particle masses approaching zero and infinity.
Substituting \cref{eq:appendix-projection-time,eq:appendix-projection-flux,eq:appendix-projection-growth} into \cref{eq:appendix-projection-integrated} yields
\[
    \partial_t\rho_t+\nabla_{\bm x}\cdot(\rho_t\bm u_t)=\rho_t g_t,
\]
which is the physical unbalanced continuity equation in \cref{eq:physical-continuity-main}.

\subsection{Proof of \texorpdfstring{\cref{thm:extended-dynamic-static}}{the Dynamic--Static Equivalence Theorem}}
\label{app:dynamic-static-equivalence}

Following the dynamic--static equivalence arguments of \citet{benamou2000computational} and \citet{chizat2018unbalanced}, we present the proof from a particle-trajectory perspective in the extended space
\[
    \mathcal W=\R^d\times\R,
    \qquad
    \norm{(\bm a,b)}_\delta^2
    :=
    \norm{\bm a}^2+\delta^2\abs{b}^2.
\]
\noindent{\normalsize\textbf{Single-particle optimality.}}\par

Indeed, for any absolutely continuous path connecting $\bm w_0$ and $\bm w_1$, the Cauchy--Schwarz inequality gives
\[
    \int_0^1\norm{\dot{\bm{x}}_t}^2\dd t
    \ge
    \left\|\int_0^1\dot{\bm{x}}_t\dd t\right\|^2
    =
    \norm{\bm{x}_1-\bm{x}_0}^2,
    \qquad
    \int_0^1\abs{\dot y_t}^2\dd t
    \ge
    \abs{y_1-y_0}^2,
\]
with equality when $\dot{\bm{x}}_t$ and $\dot y_t$ are constant in time. Thus the path in \cref{eq:main-path} minimizes
\cref{eq:main-action}, with minimum \cref{eq:main-pair-cost}.

Let $\mathcal B_\delta$ and $\mathcal K_\delta$ denote the values of \cref{eq:main-dynamic-esfm,eq:main-extended-static}, respectively. We prove the two inequalities separately.

\noindent{\normalsize\textbf{Dynamic cost is no smaller than static cost: \(\mathcal B_\delta\ge\mathcal K_\delta\).}}\par

For a dynamic solution represented by a particle flow, let $\hat{\bm s}_t=(\hat{\bm u}_t,\hat g_t)$. Let $\bm W_t(\bm w_0)$ denote the extended state reached at time $t$ by a particle that starts from $\bm w_0$ and follows the extended velocity field $\hat{\bm s}_t$; equivalently, $\bm W_t(\bm w_0)$ satisfies
\[
    \dot{\bm W}_t(\bm w_0)
    =
    \hat{\bm s}_t\bigl(\bm W_t(\bm w_0)\bigr),
    \qquad
    \bm W_0(\bm w_0)=\bm w_0.
\]
The extended continuity equation implies
\[
    \hat\rho_t=(\bm W_t)_\#\hat\rho_0.
\]
Hence
\[
    \hat\pi
    :=
    (\mathrm{id},\bm W_1)_\#\hat\rho_0
\]
is a coupling between $\hat\rho_0$ and $\hat\rho_1$. Its terminal marginal is the same $\hat\rho_1$ as in the dynamic problem and therefore satisfies the required weighted projection onto $\rho_1$.

Using the pushforward identity and then Cauchy--Schwarz along each particle path gives
\[
\begin{aligned}
    &\int_0^1\int_{\mathcal W}
    \norm{\hat{\bm s}_t(\bm w)}_\delta^2
    \dd\hat\rho_t(\bm w)\dd t
    \\
    &=
    \int_{\mathcal W}\int_0^1
    \norm{\dot{\bm W}_t(\bm w_0)}_\delta^2
    \dd t\dd\hat\rho_0(\bm w_0)
    \\
    &\ge
    \int_{\mathcal W}
    \norm{\bm W_1(\bm w_0)-\bm w_0}_\delta^2
    \dd\hat\rho_0(\bm w_0)
    \\
    &=
    \iint_{\mathcal W^2}
    \norm{\bm w_1-\bm w_0}_\delta^2
    \dd\hat\pi(\bm w_0,\bm w_1)
    \ge
    \mathcal K_\delta(\rho_0,\rho_1).
\end{aligned}
\]
Since this holds for every feasible dynamic solution, taking the infimum gives
\[
    \mathcal B_\delta(\rho_0,\rho_1)
    \ge
    \mathcal K_\delta(\rho_0,\rho_1).
\]

\noindent{\normalsize\textbf{Static cost is no smaller than dynamic cost: \(\mathcal K_\delta\ge\mathcal B_\delta\).}}\par

Conversely, let $(\hat\pi,\hat\rho_1)$ be any feasible static solution and define the straight conditional path
\[
    \bm W_t(\bm w_0,\bm w_1)
    :=
    (1-t)\bm w_0+t\bm w_1,
    \qquad
    \hat\rho_t:=(\bm W_t)_\#\hat\pi.
\]
The boundary conditions follow from the two marginals of $\hat\pi$. At an intermediate state $\bm w$, define the macroscopic extended velocity as the conditional average
\[
    \hat{\bm s}_t(\bm w)
    :=
    \mathbb E_{\hat\pi}
    \left[
    \bm w_1-\bm w_0
    \mid
    \bm W_t=\bm w
    \right].
\]
Applying the chain rule to a smooth test function along every straight path and then integrating over $\hat\pi$ shows that $(\hat\rho_t,\hat{\bm s}_t)$ satisfies the extended continuity equation in the weak sense.

Indeed, let $\varphi\in C_c^1([0,1]\times\mathcal W)$. Along each path,
\[
    \frac{\dd}{\dd t}
    \varphi\bigl(t,\bm W_t(\bm w_0,\bm w_1)\bigr)
    =
    \partial_t\varphi(t,\bm W_t)
    +
    \nabla_{\bm w}\varphi(t,\bm W_t)
    \cdot(\bm w_1-\bm w_0).
\]
Integrating first over $t$ and then over the endpoint pairs gives
\[
\begin{aligned}
    &\int_{\mathcal W}\varphi(1,\bm w)\dd\hat\rho_1(\bm w)
    -
    \int_{\mathcal W}\varphi(0,\bm w)\dd\hat\rho_0(\bm w)
    \\
    &=
    \int_0^1\iint_{\mathcal W^2}
    \Big[
    \partial_t\varphi(t,\bm W_t)
    +
    \nabla_{\bm w}\varphi(t,\bm W_t)
    \cdot(\bm w_1-\bm w_0)
    \Big]
    \dd\hat\pi\dd t
    \\
    &=
    \int_0^1\int_{\mathcal W}
    \Big[
    \partial_t\varphi(t,\bm w)
    +
    \nabla_{\bm w}\varphi(t,\bm w)
    \cdot\hat{\bm s}_t(\bm w)
    \Big]
    \dd\hat\rho_t(\bm w)\dd t.
\end{aligned}
\]
The last equality uses $\hat\rho_t=(\bm W_t)_\#\hat\pi$ and the definition of $\hat{\bm s}_t$ as the conditional mean. This identity is precisely the weak form of
\[
    \partial_t\hat\rho_t
    +
    \operatorname{div}_{\bm w}
    \bigl(\hat\rho_t\hat{\bm s}_t\bigr)
    =0,
\]
which is the extended continuity equation in \cref{eq:main-dynamic-esfm}.

We next compare the costs. For fixed $t$, the definition of the conditional mean and Jensen's inequality imply
\[
\begin{aligned}
    \norm{\hat{\bm s}_t(\bm w)}_\delta^2
    &=
    \left\|
    \mathbb E_{\hat\pi}
    \left[
    \bm w_1-\bm w_0
    \mid
    \bm W_t=\bm w
    \right]
    \right\|_\delta^2
    \\
    &\le
    \mathbb E_{\hat\pi}
    \left[
    \norm{\bm w_1-\bm w_0}_\delta^2
    \mid
    \bm W_t=\bm w
    \right].
\end{aligned}
\]
Since $\hat\rho_t=(\bm W_t)_\#\hat\pi$, averaging this inequality over the intermediate states $\bm w$ recovers the expectation over all endpoint pairs:
\[
\begin{aligned}
    \int_{\mathcal W}
    \norm{\hat{\bm s}_t(\bm w)}_\delta^2
    \dd\hat\rho_t(\bm w)
    &\le
    \int_{\mathcal W}
    \mathbb E_{\hat\pi}
    \left[
    \norm{\bm w_1-\bm w_0}_\delta^2
    \mid
    \bm W_t=\bm w
    \right]
    \dd\hat\rho_t(\bm w)
    \\
    &=
    \mathbb E_{\hat\pi}
    \left[
    \norm{\bm w_1-\bm w_0}_\delta^2
    \right]
    \\
    &=
    \iint_{\mathcal W^2}
    \norm{\bm w_1-\bm w_0}_\delta^2
    \dd\hat\pi.
\end{aligned}
\]
Integrating over $t\in[0,1]$ therefore yields
\[
    \int_0^1\int_{\mathcal W}
    \norm{\hat{\bm s}_t(\bm w)}_\delta^2
    \dd\hat\rho_t(\bm w)\dd t
    \le
    \iint_{\mathcal W^2}
    \norm{\bm w_1-\bm w_0}_\delta^2
    \dd\hat\pi.
\]
Thus every feasible static solution generates a feasible dynamic solution with no larger cost. Taking the infimum gives
\[
    \mathcal B_\delta(\rho_0,\rho_1)
    \le
    \mathcal K_\delta(\rho_0,\rho_1).
\]

Combining the two inequalities gives
\[
    \mathcal B_\delta(\rho_0,\rho_1)
    =
    \mathcal K_\delta(\rho_0,\rho_1),
\]
which proves the equivalence of \cref{eq:main-dynamic-esfm,eq:main-extended-static}.

\subsection{Proof of \texorpdfstring{\cref{thm:local-log-mass}}{the Reduced Static OT Theorem}}
\label{app:local-log-mass-proof}

This section proves \cref{thm:local-log-mass} by first eliminating $y_0$
and then solving the local optimization over $y_1$.

\noindent{\normalsize\textbf{Source reduction and the local problem.}}\par

Since $\hat\rho_0(\bm{x}_0,y_0)=\rho_0(\bm{x}_0)\Dirac_0(y_0)$,
every feasible extended plan is concentrated on $y_0=0$. After imposing
$y_1\le M$, the static problem becomes
\begin{equation}
\begin{aligned}
    \inf_{\pi\ge0}\quad&
    \iiint
    \left[
    \norm{\bm{x}_1-\bm{x}_0}^2+\delta^2y_1^2
    \right]\pi(\bm{x}_0,\bm{x}_1,y_1)
    \dd\bm{x}_0\dd\bm{x}_1\dd y_1,
    \\
    \text{s.t.}\quad&
    \operatorname{supp}\pi
    \subseteq
    \R^d\times\R^d\times(-\infty,M],
    \\
    &
    \iint \pi(\bm{x}_0,\bm{x}_1,y_1)\dd\bm{x}_1\dd y_1
    =
    \rho_0(\bm{x}_0),
    \\
    &
    \iint \e^{y_1}\pi(\bm{x}_0,\bm{x}_1,y_1)\dd\bm{x}_0\dd y_1
    =
    \rho_1(\bm{x}_1).
\end{aligned}
\label{eq:main-source-reduced-static}
\end{equation}

For a fixed endpoint state pair, write
$A=\gamma_0(\bm{x}_0,\bm{x}_1)$, $B=\gamma_1(\bm{x}_0,\bm{x}_1)$,
$r=B/A$, and $f(y_1)=\pi(\bm{x}_0,\bm{x}_1,y_1)$.
We consider $A>0$ and $B>0$; pairs with $A=B=0$ contribute nothing.
The state cost is fixed, and the two moments of $f$ must remain equal
to $A$ and $B$. Removing the common coefficient $\delta^2$ gives the local problem:
\begin{equation}
\begin{aligned}
    \inf_{f\in\mathcal M_+\left((-\infty,M]\right)}\quad&
    \int_{-\infty}^{M} y^2 f(y)\dd y,
    \\
    \text{s.t.}\quad&
    \int_{-\infty}^{M} f(y)\dd y=A,
    \qquad
    \int_{-\infty}^{M}\e^{y} f(y)\dd y=B.
\end{aligned}
\label{eq:appendix-local-log-mass}
\end{equation}
Let $r=B/A$. Feasibility requires $0<r\le\e^M$.

\noindent{\normalsize\textbf{Lagrangian and dual problem.}}\par

The Lagrangian associated with \cref{eq:appendix-local-log-mass} is
\[
    \mathcal L
    =
    \int_{-\infty}^{M}
    \left(
    y^2-\lambda_1-\lambda_2\e^y
    \right)f(y)\dd y
    +
    \lambda_1A+\lambda_2B.
\]
Minimization over $f\in\mathcal M_+\bigl((-\infty,M]\bigr)$ requires
\[
    y^2-\lambda_1-\lambda_2\e^y\ge0,
    \qquad
    \text{for}\: y\le M.
\]
Hence
\[
    \lambda_1
    \le
    \inf_{y\le M}
    \left(
    y^2-\lambda_2\e^y
    \right).
\]
Define
\[
    g_{\lambda_2}(y):=y^2-\lambda_2\e^y.
\]
For fixed $\lambda_2$, the dual objective is maximized by taking
\[
    \lambda_1
    =
    \inf_{y\le M}g_{\lambda_2}(y).
\]
Therefore, the dual problem can be written as
\begin{equation}
    \sup_{\lambda_2\in\R}
    \left\{
    A\inf_{y\le M}g_{\lambda_2}(y)
    +
    \lambda_2B
    \right\}.
\label{eq:appendix-local-dual}
\end{equation}

\noindent{\normalsize\textbf{Solving the dual problem and determining \(\yp\).}}\par

To solve the dual problem in \cref{eq:appendix-local-dual}, we first examine the minimum of $g_{\lambda_2}$ over $(-\infty,M]$. Its derivative is
\[
    g_{\lambda_2}'(y)
    =
    2y-\lambda_2\e^y.
\]
An interior stationary point $\yp(\lambda_2)$ satisfies
\[
    2\yp(\lambda_2)\e^{-\yp(\lambda_2)}
    =
    \lambda_2.
\]
The extrema of $g_{\lambda_2}$ have the following structure:
\begin{itemize}
    \item If $\lambda_2\le0$, then $g_{\lambda_2}'(y)=0$ has a unique solution, and the minimizer satisfies $\yp(\lambda_2)\le0$.
    \item If $0<\lambda_2<\frac{2}{\e}$, the function
    \[
        \phi(y):=2y\e^{-y}
    \]
    reaches its maximum $\frac{2}{\e}$ at $y=1$. In this case, $g_{\lambda_2}'(y)=0$ has two roots, and its unique local minimum satisfies
    \[
        \yp(\lambda_2)\in(0,1).
    \]
    The global minimum may occur either at this interior point or at the boundary point $M$.
    \item If $\lambda_2\ge\frac{2}{\e}$, then $g_{\lambda_2}'(y)\le0$ for all $y$. Thus, $g_{\lambda_2}$ is decreasing and reaches its minimum at $M$.
\end{itemize}
Consequently,
\[
    \inf_{y\le M}g_{\lambda_2}(y)
    =
    \min\left\{
    g_{\lambda_2}\bigl(\yp(\lambda_2)\bigr),
    g_{\lambda_2}(M)
    \right\}.
\]
Moreover,
\[
    g_{\lambda_2}\bigl(\yp(\lambda_2)\bigr)-g_{\lambda_2}(M)
    =
    \yp(\lambda_2)^2-M^2
    +
    \lambda_2
    \left(
    \e^M-\e^{\yp(\lambda_2)}
    \right).
\]
By the envelope theorem,
\[
    \frac{\dd}{\dd\lambda_2}
    \left[
    g_{\lambda_2}\bigl(\yp(\lambda_2)\bigr)-g_{\lambda_2}(M)
    \right]
    =
    \e^M-\e^{\yp(\lambda_2)}
    >
    0.
\]
Thus, as $\lambda_2$ increases, the global minimizer switches from the interior point $\yp(\lambda_2)$ to the boundary point $M$. Let $\lambda_2^*$ be the switching value and denote the corresponding interior minimizer by
\[
    \yp:=\yp(\lambda_2^*).
\]
At the switching point,
\begin{equation}
    g_{\lambda_2^*}(\yp)
    =
    g_{\lambda_2^*}(M).
\label{eq:appendix-switching-condition}
\end{equation}
Therefore,
\[
    y_{\min}(\lambda_2)
    =
    \begin{cases}
    \yp(\lambda_2),
    &\lambda_2<\lambda_2^*,
    \\[0.4em]
    \yp\text{ or }M,
    &\lambda_2=\lambda_2^*,
    \\[0.4em]
    M,
    &\lambda_2>\lambda_2^*.
    \end{cases}
\]

We now determine $\yp$. Since it is an interior local minimizer,
\[
    \lambda_2^*
    =
    2\yp\e^{-\yp}.
\]
Substituting this identity into \cref{eq:appendix-switching-condition} gives
\[
    \yp^2
    -
    \left(2\yp\e^{-\yp}\right)\e^{\yp}
    =
    M^2
    -
    \left(2\yp\e^{-\yp}\right)\e^M.
\]
After simplification,
\begin{equation}
    M^2-\yp^2+2\yp-2\yp\e^{M-\yp}
    =
    0.
\label{eq:appendix-y-star}
\end{equation}
Define
\[
    h(y)
    :=
    M^2-y^2+2y-2y\e^{M-y}.
\]
Then
\[
    h'(y)
    =
    2(y-1)\left(\e^{M-y}-1\right).
\]
For $M>1$,
\[
    h(0)=M^2>0,
    \qquad
    h(1)=M^2+1-2\e^{M-1}<0,
    \qquad
    h(M)=0.
\]
Moreover, $h$ is strictly decreasing on $(0,1)$ and strictly increasing on $(1,M)$. Hence, besides the trivial root $y=M$, the equation $h(y)=0$ has a unique nontrivial solution $\yp\in(0,1)$, which is uniquely determined by $M$.

\noindent{\normalsize\textbf{Maximizing the dual objective and recovering the optimizer.}}\par

Define the dual objective
\[
    D(\lambda_2)
    :=
    A g_{\lambda_2}\bigl(y_{\min}(\lambda_2)\bigr)
    +
    \lambda_2B.
\]
By the envelope theorem,
\[
    D'(\lambda_2)
    =
    B-A\e^{y_{\min}(\lambda_2)}.
\]
Therefore, if $D'(\lambda_2)=0$, the desired minimizer satisfies
\[
    y_{\min}(\lambda_2)
    =
    \ln\frac{B}{A}.
\]
The solution has two cases, depending on the position of $\ln(B/A)$ relative to $\yp$.

\noindent\textbf{Case 1: \(\ln(B/A)\le\yp\).}
In this case, $D'(\lambda_2)=0$ can be satisfied before $\lambda_2$ reaches $\lambda_2^*$. Hence
\[
    y_{\min}(\lambda_2)
    =
    \yp(\lambda_2)
    =
    \ln\frac{B}{A}.
\]
The optimal plan is concentrated at the single point $\ln(B/A)$:
\[
    f^*(y)
    =
    A\Dirac\left(y-\ln\frac{B}{A}\right).
\]
The corresponding minimum is
\[
    \inf_f
    \int_{-\infty}^{M}y^2f(y)\dd y
    =
    A\left(\ln\frac{B}{A}\right)^2.
\]

\noindent\textbf{Case 2: \(\yp\le\ln(B/A)\le M\).}
When $\lambda_2<\lambda_2^*$,
\[
    y_{\min}(\lambda_2)
    =
    \yp(\lambda_2)
    <
    \yp
    \le
    \ln\frac{B}{A},
\]
and therefore
\[
    D'(\lambda_2)
    =
    B-A\e^{y_{\min}(\lambda_2)}
    >
    0.
\]
When $\lambda_2>\lambda_2^*$,
\[
    y_{\min}(\lambda_2)
    =
    M
    \ge
    \ln\frac{B}{A},
\]
and therefore
\[
    D'(\lambda_2)
    =
    B-A\e^M
    \le
    0.
\]
Thus, the dual objective reaches its maximum at $\lambda_2^*$. At this point,
\[
    g_{\lambda_2^*}(\yp)
    =
    g_{\lambda_2^*}(M),
\]
and the optimal plan is supported on $\yp$ and $M$:
\[
    f^*(y)
    =
    w_1\Dirac(y-\yp)
    +
    w_2\Dirac(y-M).
\]
The weights are uniquely determined by
\begin{align*}
    w_1+w_2&=A,
    \\
    w_1\e^{\yp}+w_2\e^M&=B,
\end{align*}
which gives
\[
    w_1
    =
    \frac{A\e^M-B}{\e^M-\e^{\yp}},
    \qquad
    w_2
    =
    \frac{B-A\e^{\yp}}{\e^M-\e^{\yp}}.
\]
The corresponding minimum is
\begin{align*}
    \inf_f
    \int_{-\infty}^{M}y^2f(y)\dd y
    &=
    w_1\yp^2+w_2M^2
    \notag
    \\
    &=
    \frac{A\e^M-B}{\e^M-\e^{\yp}}\,\yp^2
    +
    \frac{B-A\e^{\yp}}{\e^M-\e^{\yp}}\,M^2.
\end{align*}

Writing $r=B/A$, an optimizer is
\begin{equation}
    f^\star(y_1)
    =
    \begin{cases}
    A\Dirac\!\left(y_1-\ln r\right),
    &0<r\le\e^{\yp},
    \\[0.8em]
    \displaystyle
    A\left[
    \frac{\e^M-r}{\e^M-\e^{\yp}}\Dirac(y_1-\yp)
    +
    \frac{r-\e^{\yp}}{\e^M-\e^{\yp}}\Dirac(y_1-M)
    \right],
    &\e^{\yp}<r\le\e^M.
    \end{cases}
\label{eq:main-local-optimizer}
\end{equation}
The minimum is denoted by $A\Phi_M(r)$, where
\[
    \Phi_M(r)
    =
    \begin{cases}
    (\ln r)^2,
    &0<r\le\e^{\yp},
    \\[0.6em]
    \displaystyle
    \frac{\e^M-r}{\e^M-\e^{\yp}}\yp^2
    +
    \frac{r-\e^{\yp}}{\e^M-\e^{\yp}}M^2,
    &\e^{\yp}<r\le\e^M.
    \end{cases}
\]
Using the critical equation,
\[
    \frac{M^2-\yp^2}{\e^M-\e^{\yp}}
    =
    \frac{2\yp}{\e^{\yp}},
    \qquad
    \frac{\e^M\yp^2-\e^{\yp} M^2}{\e^M-\e^{\yp}}
    =
    \yp^2-2\yp.
\]
Therefore,
\begin{equation}
    \Phi_M(r)
    =
    \begin{cases}
    (\ln r)^2,
    &0<r\le\e^{\yp},
    \\[0.6em]
    \displaystyle
    \frac{2\yp}{\e^{\yp}}r+\yp^2-2\yp,
    &\e^{\yp}<r\le\e^M.
    \end{cases}
\label{eq:appendix-phi}
\end{equation}
This function is continuous and continuously differentiable at $r=\e^{\yp}$.

\noindent{\normalsize\textbf{Recovering the reduced problem.}}\par

For any feasible plan in \cref{eq:main-source-reduced-static}, the local
bound gives
\[
    \int_{-\infty}^{M}y_1^2\pi(\bm{x}_0,\bm{x}_1,y_1)\dd y_1
    \ge \gamma_0(\bm{x}_0,\bm{x}_1)\Phi_M(R(\bm{x}_0,\bm{x}_1)).
\]
Integrating over endpoint state pairs bounds its cost from below by the
reduced objective in \cref{eq:main-reduced-static}.
Conversely, for any feasible $(\gamma_0,\gamma_1)$ in that reduced problem,
use \cref{eq:main-local-optimizer} at each pair with $A=\gamma_0$ and $r=R$.
The resulting plan has marginals
$\gamma_0$ and $\gamma_1$, satisfies the growth cap, and attains the local
bound. Thus the two problems have the same infimum. This proves
\cref{thm:local-log-mass}.

\subsection{Proof of \texorpdfstring{\cref{prop:entropic-solver}}{the Entropic Solver Proposition} and Update Rules}
\label{app:solver}

This section derives the Sinkhorn-like updates and proves \cref{prop:entropic-solver}.

\noindent{\normalsize\textbf{Discrete reduced problem.}}\par

Let the empirical source and target measures be
\[
    \rho_0=\sum_{i=1}^n a_i\Dirac_{\bm{x}_i^0},
    \qquad
    \rho_1=\sum_{j=1}^m b_j\Dirac_{\bm{x}_j^1},
\]
Define the state space cost
\[
    C_{ij}=\norm{\bm{x}_i^0-\bm{x}_j^1}^2.
\]
We write $\Gamma_{ij}$ for the discrete base coupling $\gamma_0$ and $R_{ij}$ for the local growth ratio $\dd\gamma_1/\dd\gamma_0$. Then \cref{eq:main-reduced-static} becomes
\begin{equation}
\begin{aligned}
    \min_{\Gamma\ge0,\;0<R\le\e^M}\quad&
    \sum_{i=1}^n\sum_{j=1}^m
    \left[
    C_{ij}+\delta^2\Phi_M(R_{ij})
    \right]\Gamma_{ij},
    \\
    \text{s.t.}\quad&
    \sum_{j=1}^m\Gamma_{ij}=a_i,
    \qquad i=1,\ldots,n,
    \\
    &
    \sum_{i=1}^nR_{ij}\Gamma_{ij}=b_j,
    \qquad j=1,\ldots,m.
\end{aligned}
\label{eq:appendix-discrete}
\end{equation}

\noindent{\normalsize\textbf{Entropic regularization and Lagrangian.}}\par

For computation we add an entropy term to the base coupling:
\begin{equation}
\begin{aligned}
    \min_{\Gamma\ge0,\;0<R\le\e^M}\quad
    \sum_{ij}
    \Big[
    C_{ij}\Gamma_{ij}
    +
    \delta^2\Gamma_{ij}\Phi_M(R_{ij})
    +
    \varepsilon\Gamma_{ij}(\ln\Gamma_{ij}-1)
    \Big],
\end{aligned}
\label{eq:appendix-entropic-primal}
\end{equation}
subject to the same constraints as in \cref{eq:appendix-discrete}. Introduce row dual variables $u_i$ and column dual variables 
$v_j$. The Lagrangian of \cref{eq:appendix-entropic-primal} is
\[
\begin{aligned}
\mathcal L(\Gamma,R;u,v)
&={}
\sum_{ij}
\Gamma_{ij}
\left[
C_{ij}
+\delta^2\Phi_M(R_{ij})
+\varepsilon(\ln\Gamma_{ij}-1)
-u_i-v_jR_{ij}
\right]
\\
&\quad +
\sum_i u_i a_i
+
\sum_j v_j b_j .
\end{aligned}
\]
For fixed $R$, stationarity with respect to $\Gamma_{ij}$ gives
\begin{equation}
    \Gamma_{ij}
    =
    \exp\left(
    \frac{u_i+v_jR_{ij}-C_{ij}-\delta^2\Phi_M(R_{ij})}{\varepsilon}
    \right).
\label{eq:appendix-gamma-fixed-r}
\end{equation}
Substituting \cref{eq:appendix-gamma-fixed-r} into the Lagrangian gives the dual function
\[
\begin{aligned}
    \inf_{\Gamma}\mathcal L(\Gamma,R;u,v)
    =
    \sum_i u_i a_i+\sum_j v_j b_j
    -
    \varepsilon
    \sum_{ij}
    \exp\left(
    \frac{u_i+v_jR_{ij}-C_{ij}-\delta^2\Phi_M(R_{ij})}{\varepsilon}
    \right).
\end{aligned}
\]

Thus the remaining local step is a scalar maximization over the growth ratio. We make this step explicit next.

\noindent{\normalsize\textbf{The scalar \(R\) subproblem.}}\par

For a fixed target index $j$ and dual variable $v_j$, define the scalar objective for the growth-ratio update as
\[
    h_j(R;v_j)
    :=
    v_jR-\delta^2\Phi_M(R),
    \qquad
    0<R\le\e^M.
\]
Define the column value function as the optimal value of this one-dimensional problem:
\[
    H_j(v_j)
    :=
    \max_{0<R\le\e^M}
    h_j(R;v_j).
\]

Using the piecewise expression of $\Phi_M$ in \cref{eq:appendix-phi}, the objective $h_j$ becomes
\[
h_j(R;v_j)
=
\begin{cases}
v_jR-\delta^2(\ln R)^2,
&0<R\le\e^{\yp},
\\[0.4em]
\left(
v_j-\delta^2\dfrac{2\yp}{\e^{\yp}}
\right)R
+\delta^2(2\yp-\yp^2),
&\e^{\yp}<R\le\e^M.
\end{cases}
\]
The slope of the second branch is
\[
    v_j-\delta^2\frac{2\yp}{\e^{\yp}},
\]
and its sign determines the following three regimes.

\noindent\textbf{Interior branch.}
If $v_j<\delta^2 2\yp/\e^{\yp}$, the second branch is decreasing, so the maximizer lies on the first branch. Setting $p=\ln R$ and differentiating gives
\[
    v_j-2\delta^2p\e^{-p}=0.
\]
Thus the interior optimizer is
\[
    R^\star(v_j)
    =
    \exp\left[
    -W_0\left(-\frac{v_j}{2\delta^2}\right)
    \right],
\]
where $W_0$ denotes the principal branch of the Lambert $W$ function, defined by
$W_0(z)\e^{W_0(z)}=z$. The principal branch $W_0$ is increasing on
$[-1/\e,\infty)$.
Since the argument $-v_j/(2\delta^2)$ decreases as $v_j$ increases, the
optimizer
$R^\star(v_j)=\exp[-W_0(-v_j/(2\delta^2))]$
is increasing in $v_j$.

\noindent\textbf{Transition branch.}
If $v_j=\delta^2 2\yp/\e^{\yp}$, the derivative on the first branch is
\[
    \frac{\partial}{\partial R}h_j(R;v_j)
    =
    2\delta^2
    \left(
    \frac{\yp}{\e^{\yp}}
    -
    \frac{\ln R}{R}
    \right).
\]
For $0<R\le\e^{\yp}$ and $\yp<1$, we have
$\ln R/R\le\yp/\e^{\yp}$, so the first branch is nondecreasing and reaches its
maximum at $R=\e^{\yp}$. The second branch is flat and has the same value as
this boundary value. Every point in $[\e^{\yp},\e^M]$ is therefore optimal.

\noindent\textbf{Capped branch.}
If $v_j>\delta^2 2\yp/\e^{\yp}$, the second branch is increasing, so the maximizer is attained at the growth cap $R=\e^M$.

Combining the three regimes gives the optimal set
\[
\mathcal R^\star(v_j)
=
\operatorname*{arg\,max}_{0<R\le\e^M}
h_j(R;v_j)
=
\begin{cases}
\left\{
\exp\left[-W_0\left(-\dfrac{v_j}{2\delta^2}\right)\right]
\right\},
&v_j<\delta^2\dfrac{2\yp}{\e^{\yp}},
\\[1.0em]
[\e^{\yp},\e^M],
&v_j=\delta^2\dfrac{2\yp}{\e^{\yp}},
\\[0.6em]
\{\e^M\},
&v_j>\delta^2\dfrac{2\yp}{\e^{\yp}}.
\end{cases}
\]
The corresponding value function is
\[
H_j(v_j)
=
\begin{cases}
\delta^2
\left[
2\ln R^\star(v_j)
-
\bigl(\ln R^\star(v_j)\bigr)^2
\right],
&v_j<\delta^2\dfrac{2\yp}{\e^{\yp}},
\\[1.0em]
\delta^2(2\yp-\yp^2),
&v_j=\delta^2\dfrac{2\yp}{\e^{\yp}},
\\[0.8em]
\left(
v_j-\delta^2\dfrac{2\yp}{\e^{\yp}}
\right)\e^M
+
\delta^2(2\yp-\yp^2),
&v_j>\delta^2\dfrac{2\yp}{\e^{\yp}}.
\end{cases}
\]
After this elimination, the dual problem can be written as
\begin{equation}
\begin{aligned}
    \max_{u,v}\quad
    \mathcal D(u,v)
    =
    \sum_i u_i a_i
    +
    \sum_j v_j b_j
    -
    \varepsilon
    \sum_{ij}
    \exp\left(
    \frac{u_i+H_j(v_j)-C_{ij}}{\varepsilon}
    \right).
\end{aligned}
\label{eq:appendix-dual-final}
\end{equation}
In what follows, $R_j$ denotes a selected element of $\mathcal R^\star(v_j)$. By the envelope theorem, $H_j'(v_j)=R_j$ when the maximizer is unique; on the transition branch, the same relation is understood as a subgradient condition with $R_j\in[\e^{\yp},\e^M]$.
Taking derivatives of $\mathcal D$ in \cref{eq:appendix-dual-final} with respect to $u_i$ and $v_j$ gives
\[
    \frac{\partial \mathcal D}{\partial u_i}
    =
    a_i
    -
    \sum_j
    \exp\left(
    \frac{u_i+H_j-C_{ij}}{\varepsilon}
    \right),
    \qquad
    \frac{\partial \mathcal D}{\partial v_j}
    =
    b_j
    -
    R_j
    \sum_i
    \exp\left(
    \frac{u_i+H_j-C_{ij}}{\varepsilon}
    \right),
\]
where $H_j=H_j(v_j)$. Setting these conditions to zero yields
\begin{subequations}
\begin{align}
    a_i
    &=
    \sum_j
    \exp\left(
    \frac{u_i+H_j-C_{ij}}{\varepsilon}
    \right),
    \label{eq:appendix-row-opt}
    \\
    b_j
    &=
    R_j
    \sum_i
    \exp\left(
    \frac{u_i+H_j-C_{ij}}{\varepsilon}
    \right).
    \label{eq:appendix-column-opt}
\end{align}
\end{subequations}

\noindent{\normalsize\textbf{Alternating updates in \((u,H)\).}}\par

The explicit solver uses $u_i$ and the column values $H_j$ as update variables. Given $H$, the row condition \cref{eq:appendix-row-opt} gives the closed-form update
\begin{equation}
    u_i
    =
    \varepsilon\ln a_i
    -
    \varepsilon
    \log
    \sum_j
    \exp\left(
    \frac{H_j-C_{ij}}{\varepsilon}
    \right).
    \label{eq:appendix-u-update}
\end{equation}
Given $u$, define
\begin{equation}
    K_j
    :=
    \frac{b_j}
    {\sum_i\exp((u_i-C_{ij})/\varepsilon)}.
    \label{eq:appendix-kj}
\end{equation}
The column condition \cref{eq:appendix-column-opt} becomes
\begin{equation}
    R_j\exp(H_j/\varepsilon)=K_j,
    \qquad\text{or}\qquad
    \ln R_j+\frac{H_j}{\varepsilon}=\ln K_j.
\label{eq:appendix-r-h-k}
\end{equation}
We solve this scalar relation by cases. The only nontrivial case is the interior branch, where the left-hand side is a 
strictly increasing function of $\ln R_j$ because $\ln R_j\le\yp<1$; the transition and capped branches give explicit values. 
Thus, once the branch is selected by $K_j$, $\ln R_j$ and $H_j$ is unique. Define
\begin{equation}
    T_1
    :=
    \yp+\frac{\delta^2(2\yp-\yp^2)}{\varepsilon},
    \qquad
    T_2
    :=
    M+\frac{\delta^2(2\yp-\yp^2)}{\varepsilon}.
\label{eq:appendix-thresholds}
\end{equation}
Using the thresholds in \cref{eq:appendix-thresholds}, the three regimes give the following updates.

\noindent\textbf{Interior branch.}
If $\ln K_j\le T_1$, set $p_j=\ln R_j$. On this branch,
\[
    H_j=\delta^2(2p_j-p_j^2),
\]
so \cref{eq:appendix-r-h-k} gives
\[
    p_j+\frac{\delta^2}{\varepsilon}(2p_j-p_j^2)=\ln K_j.
\]
Solving the quadratic and taking the root with $p_j\le\yp$ yields
\[
    p_j
    =
    \frac{
    (\varepsilon+2\delta^2)
    -
    \sqrt{(\varepsilon+2\delta^2)^2-4\delta^2\varepsilon\ln K_j}
    }
    {2\delta^2},
\]
and then
\[
    R_j=\e^{p_j},
    \qquad
    H_j=\delta^2(2p_j-p_j^2).
\]

\noindent\textbf{Transition branch.}
If $T_1<\ln K_j\le T_2$, the column value is constant:
\[
    H_j=\delta^2(2\yp-\yp^2).
\]
Then \cref{eq:appendix-r-h-k} gives
\[
    R_j
    =
    K_j
    \exp\left(
    -\frac{\delta^2(2\yp-\yp^2)}{\varepsilon}
    \right).
\]
The threshold condition is exactly the statement that $R_j$ lies in $[\e^{\yp},\e^M]$.

\noindent\textbf{Capped branch.}
If $\ln K_j>T_2$, the maximizer is $R_j=\e^M$. \cref{eq:appendix-r-h-k} gives
\[
    R_j=\e^M,
    \qquad
    H_j=\varepsilon(\ln K_j-M).
\]

\noindent{\normalsize\textbf{Recovery and sampling quantities.}}\par

After convergence, the base coupling and the target-mass coupling are recovered as
\begin{equation}
    \Gamma^{(0)}_{ij}
    =
    \exp\left(
    \frac{u_i+H_j-C_{ij}}{\varepsilon}
    \right),
    \qquad
    \Gamma^{(1)}_{ij}
    =
    R_j\Gamma^{(0)}_{ij}.
\label{eq:appendix-gamma-recovery}
\end{equation}
The recovered base coupling $\Gamma^{(0)}$ is used to sample  $(\bm{x}_0,\bm{x}_1)$ 
and fix the initial log-mass as $y_0=0$. The terminal log-mass $y_1$ is then sampled according to the normalized 
version of the local optimizer in \cref{eq:main-local-optimizer}, using the recovered growth ratio $R_j$:
\begin{equation}
    f^*(y_1)
    =
    \begin{cases}
    \Dirac(y_1-\ln R_j),
    &0<R_j\le\e^{\yp},
    \\[0.8em]
    \displaystyle
    \frac{\e^M-R_j}{\e^M-\e^{\yp}}\Dirac(y_1-\yp)
    +
    \frac{R_j-\e^{\yp}}{\e^M-\e^{\yp}}\Dirac(y_1-M),
    &\e^{\yp}<R_j\le\e^M.
    \end{cases}
\label{eq:appendix-log-mass-sampling}
\end{equation}
Thus, if $R_j\le\e^{\yp}$, we set $y_1=\ln R_j$; otherwise we sample $y_1=\yp$ with probability $(\e^M-R_j)/(\e^M-\e^{\yp})$ and $y_1=M$ with probability $(R_j-\e^{\yp})/(\e^M-\e^{\yp})$.

The complete Sinkhorn-like solver is summarized below; it computes the recovered couplings and growth ratios used for endpoint and terminal log-mass sampling.

\begin{algorithm}[htbp]
\caption{Entropic Solver for the ESFM Static Coupling}
\label{alg:appendix-esfm-static-coupling}
\begin{algorithmic}[1]
\STATE \textbf{Input:} source weights and states $\{a_i,\bm{x}_i^0\}_{i=1}^n$, target weights and states $\{b_j,\bm{x}_j^1\}_{j=1}^m$, parameters $\delta,M,\varepsilon$.
\STATE Compute $C_{ij}=\norm{\bm{x}_i^0-\bm{x}_j^1}^2$ and solve for $\yp\in(0,1)$ using \cref{eq:appendix-y-star}.
\STATE Check $\sum_j b_j\le\e^M\sum_i a_i$.
\STATE Initialize $H_j$.
\REPEAT
    \STATE Update every $u_i$ using \cref{eq:appendix-u-update}.
    \STATE Compute every $K_j$ using \cref{eq:appendix-kj}.
    \STATE Update every $H_j$ and $R_j$ using the three branch formulas above.
\UNTIL{the marginal residuals or iterate changes are below tolerance}
\STATE Recover $\Gamma^{(0)}$ and $\Gamma^{(1)}$ using \cref{eq:appendix-gamma-recovery}.
\STATE \textbf{Output:} $\Gamma^{(0)}$, $\Gamma^{(1)}$, and $R$.
\end{algorithmic}
\end{algorithm}

\subsection{Proof of \texorpdfstring{\cref{prop:particle-splitting}}{the Source-Particle Splitting Proposition}}
\label{app:particle-splitting}

 Consider the discrete problem
\begin{equation}
\begin{aligned}
\min_{\Gamma\geq 0,\;0<R\leq \e^M}\quad &
    \sum_{i,j}
    \left[
        C_{ij}+\delta^2\Phi_M(R_{ij})
    \right]\Gamma_{ij},
\\
\text{s.t.}\quad &
    \sum_j \Gamma_{ij}=a_i,
    \qquad
    \sum_i R_{ij}\Gamma_{ij}=b_j.
\end{aligned}
\label{eq:discrete-splitting}
\end{equation}
Let $a_i=a_{i_1}+a_{i_2}$ with $a_{i_1},a_{i_2}>0$.

\begin{proof}
Because the two split particles remain at the same location as the original particle,
\begin{equation*}
    C_{i_1j}=C_{i_2j}=C_{ij}.
\end{equation*}
First consider any feasible solution of the split problem, allowing
$R_{i_1j}$ and $R_{i_2j}$ to be chosen independently. For each target particle
$j$, define
\begin{equation*}
    \Gamma_{ij}
    =
    \Gamma_{i_1j}+\Gamma_{i_2j},
\end{equation*}
and, whenever $\Gamma_{ij}>0$,
\begin{equation*}
    R_{ij}
    =
    \frac{
        R_{i_1j}\Gamma_{i_1j}
        +
        R_{i_2j}\Gamma_{i_2j}
    }{
        \Gamma_{i_1j}+\Gamma_{i_2j}
    }.
\end{equation*}
If $\Gamma_{ij}=0$, choose any $R_{ij}\in(0,\e^M]$; the corresponding
transport and growth contributions are zero. 
This aggregation preserves the source and target constraints as well as the
bound $0<R_{ij}\leq \e^M$. The spatial contribution is unchanged because the
split particles are at the same location. Since $\Phi_M$ is convex, Jensen's inequality gives
\begin{equation*}
\begin{aligned}
    \Gamma_{i_1j}\Phi_M(R_{i_1j})
    +
    \Gamma_{i_2j}\Phi_M(R_{i_2j})
\geq
    \Gamma_{ij}
    \Phi_M(R_{ij}).
\end{aligned}
\end{equation*}
Therefore aggregation cannot increase the objective. The optimal value of the
original problem is consequently no larger than that of the split problem.

To attain equality, take any feasible solution $(\Gamma,R)$ of the original
problem and define
\begin{equation*}
    \alpha=\frac{a_{i_1}}{a_i}.
\end{equation*}
Construct a feasible split solution by setting
\begin{equation*}
\begin{aligned}
    \Gamma_{i_1j} &= \alpha\Gamma_{ij},
    &
    \Gamma_{i_2j} &= (1-\alpha)\Gamma_{ij},
    \\
    R_{i_1j} &= R_{ij},
    &
    R_{i_2j} &= R_{ij},
\end{aligned}
\end{equation*}
This choice gives equality in Jensen's inequality and preserves both marginal
constraints and the objective value. 
Therefore, the original and split problems have the same optimal value.
\end{proof}

\subsection{Proof of \texorpdfstring{\cref{thm:conditional-marginal-esfm}}{the Gradient Equivalence Theorem}}
\label{app:conditional-marginal-equivalence}

The endpoint distribution $q(\bm z)$ is defined on
$\mathcal Z=\mathcal W\times\mathcal W$, with $\bm z=(\bm w_0,\bm w_1)$.
Write the conditional field as
$\textcolor{black}{\hat{\bm s}_t}(\bm w\mid\bm z)=(\textcolor{black}{\hat{\bm u}_t}(\bm w\mid\bm z),\textcolor{black}{\hat g_t}(\bm w\mid\bm z))$.
The marginal path and fields used in
\cref{eq:main-marginal-loss,thm:conditional-marginal-esfm} are
\begin{align}
    \hat\rho_t(\bm w)
    &=
    \int_{\mathcal Z}
    \hat\rho_t(\bm w\mid\bm z)q(\bm z)\dd\bm z,
    \label{eq:main-marginal-density}
    \\
    \textcolor{black}{\hat{\bm u}_t}(\bm w)
    &=
    \frac{
    \int_{\mathcal Z}
    \textcolor{black}{\hat{\bm u}_t}(\bm w\mid\bm z)
    \hat\rho_t(\bm w\mid\bm z)q(\bm z)\dd\bm z
    }{
    \hat\rho_t(\bm w)
    },
    \label{eq:main-marginal-u}
    \\
    \textcolor{black}{\hat g_t}(\bm w)
    &=
    \frac{
    \int_{\mathcal Z}
    \textcolor{black}{\hat g_t}(\bm w\mid\bm z)
    \hat\rho_t(\bm w\mid\bm z)q(\bm z)\dd\bm z
    }{
    \hat\rho_t(\bm w)
    }.
    \label{eq:main-marginal-g}
\end{align}
\begin{proof}[Proof of \cref{thm:conditional-marginal-esfm}]
We prove the claim for the state velocity component; the log mass growth component is identical with scalar products. The conditional velocity part of \cref{eq:main-loss} is
\[
\begin{aligned}
    \mathcal L_{\bm v}^{\mathrm{cond}}(\theta)
    =
    \int_0^1\int_{\mathcal W}\int_{\mathcal Z}
    \norm{\bm v_\theta(\bm w,t)-\textcolor{black}{\hat{\bm u}_t}(\bm w\mid\bm z)}^2
    \hat\rho_t(\bm w\mid\bm z)q(\bm z)
    \dd\bm z\dd\bm w\dd t .
\end{aligned}
\]
Expanding the square gives
\[
\begin{aligned}
    \mathcal L_{\bm v}^{\mathrm{cond}}(\theta)
    =
    \int_0^1\int_{\mathcal W}
    \Bigg[
    &\norm{\bm v_\theta(\bm w,t)}^2
    \int_{\mathcal Z}
    \hat\rho_t(\bm w\mid\bm z)q(\bm z)
    \dd\bm z
    \\
    &-
    2\left\langle
    \bm v_\theta(\bm w,t),
    \int_{\mathcal Z}
    \textcolor{black}{\hat{\bm u}_t}(\bm w\mid\bm z)
    \hat\rho_t(\bm w\mid\bm z)q(\bm z)
    \dd\bm z
    \right\rangle
    \\
    &+
    \int_{\mathcal Z}
    \norm{\textcolor{black}{\hat{\bm u}_t}(\bm w\mid\bm z)}^2
    \hat\rho_t(\bm w\mid\bm z)q(\bm z)\dd\bm z
    \Bigg]
    \dd\bm w\dd t .
\end{aligned}
\]
Expanding the marginal velocity loss in \cref{eq:main-marginal-loss} gives
\[
\begin{aligned}
    \mathcal L_{\bm v}^{\mathrm{marg}}(\theta)
    =
    \int_0^1\int_{\mathcal W}
    \Big[
    &\norm{\bm v_\theta(\bm w,t)}^2
    -
    2\left\langle
    \bm v_\theta(\bm w,t),
    \textcolor{black}{\hat{\bm u}_t}(\bm w)
    \right\rangle
    +
    \norm{\textcolor{black}{\hat{\bm u}_t}(\bm w)}^2
    \Big]
    \hat\rho_t(\bm w)\dd\bm w\dd t,
\end{aligned}
\]

By \cref{eq:main-marginal-density}, the first square term in the conditional expansion equals the corresponding term in the marginal expansion. By \cref{eq:main-marginal-u},
\[
    \int_{\mathcal Z}
    \textcolor{black}{\hat{\bm u}_t}(\bm w\mid\bm z)
    \hat\rho_t(\bm w\mid\bm z)q(\bm z)\dd\bm z
    =
    \textcolor{black}{\hat{\bm u}_t}(\bm w)\hat\rho_t(\bm w),
\]
so the cross term also matches. The remaining target-field square terms do not depend on $\theta$. Therefore,
\[
    \mathcal L_{\bm v}^{\mathrm{cond}}(\theta)
    =
    \mathcal L_{\bm v}^{\mathrm{marg}}(\theta)
    +
    C_{\bm v}.
\]

The same expansion applies to the scalar growth component $g_\phi$, using \cref{eq:main-marginal-g} in place of \cref{eq:main-marginal-u}. Hence the conditional and marginal growth losses differ only by a $\phi$-independent constant. Since $\lambda$ is fixed during training, the full losses satisfy
\[
    \mathcal L_{\mathrm{ESFM}}^{\mathrm{cond}}(\theta,\phi)
    =
    \mathcal L_{\mathrm{ESFM}}^{\mathrm{marg}}(\theta,\phi)
    +
    C,
\]
with $C$ independent of $(\theta,\phi)$, and taking gradients with respect to both parameter sets gives the result.
\end{proof}
\section{Dataset Details and Additional Numerical Results}
\label{app:additional-results}

\subsection{Experimental Details}
\label{app:experimental-details}

Experiments were run on an NVIDIA A100 GPU cluster with 128 CPU cores or a laptop with an NVIDIA RTX 5060 Laptop GPU. 
We use two separate multilayer perceptrons for velocity and growth, with five and three hidden layers, respectively. 
Both take $(t,\bm{x},y)$ as input and use LeakyReLU activations. The prediction experiments use 256 hidden units per layer,
 while held-out experiments use 512. The models are implemented in PyTorch \citep{pytorch} 
 and trained with Adam.

\paragraph{Coupling parameter $\delta$.}
The parameter $\delta>0$ weights the mass change cost relative to the state transport cost when solving the static coupling problem in \cref{eq:main-reduced-static}:
\[
\begin{aligned}
    \inf_{\gamma_0,\gamma_1\ge0}\quad&
    \iint
    \left[
    \norm{\bm{x}_1-\bm{x}_0}^2
    +\delta^2\Phi_M\!\left(R(\bm{x}_0,\bm{x}_1)\right)
    \right]\dd\gamma_0,\\
    \text{s.t.}\quad&
    \int\gamma_0(\bm{x}_0,\bm{x}_1)\dd\bm{x}_1=\rho_0(\bm{x}_0),
    \qquad
    \int\gamma_1(\bm{x}_0,\bm{x}_1)\dd\bm{x}_0=\rho_1(\bm{x}_1),\\
    &0<R(\bm{x}_0,\bm{x}_1)\le\e^M.
\end{aligned}
\]
Increasing $\delta$ penalizes growth or decay more strongly, favoring state transport when matching the endpoint distributions.
 Decreasing $\delta$ makes mass variation less costly. This balance determines the optimal coupling, including the endpoint
  pairing and terminal log mass allocation, and hence the conditional samples used for training.

\paragraph{Training weight $\lambda$.}
The parameter $\lambda>0$ balances the velocity and growth regression errors in the conditional ESFM loss from \cref{eq:main-loss}:
\[
\begin{aligned}
    \mathcal L_{\mathrm{ESFM}}^{\mathrm{cond}}(\theta,\phi)
    =
    \int_0^1\int_{\mathcal Z}\int_{\mathcal W}
    \Big[
    &\norm{\bm v_\theta(\bm w,t)-\bm u_t(\bm w\mid\bm z)}^2\\
    &+\lambda\abs{g_\phi(\bm w,t)-g_t(\bm w\mid\bm z)}^2
    \Big]
    \hat\rho_t(\bm w\mid\bm z)q(\bm z)
    \dd\bm w\dd\bm z\dd t.
\end{aligned}
\]
A larger $\lambda$ gives greater weight to the growth regression error relative to the velocity error. This parameter acts 
only during training and does not change the transport action, coupling, or conditional targets. 

\paragraph{Choice of $M$.}
The parameter $M$ sets the upper bound on the local growth ratio through $R\leq\e^M$ and determines the mass cost function $\Phi_M$ in \cref{eq:main-reduced-static}, thereby affecting the computation of the optimal coupling.

Increasing $M$ enlarges the feasible set of the local mass problem, so
$\Phi_M(r)$ is nonincreasing in $M$ for any fixed feasible growth ratio $r$.
By \cref{eq:appendix-y-star}, the threshold $\yp$ tends to zero as
$M\to\infty$. For any fixed $r>1$, the second branch of $\Phi_M$
in \cref{thm:local-log-mass} applies for sufficiently large $M$, giving
\[
    \Phi_M(r)
    =\frac{2\yp}{\e^{\yp}}r+\yp^2-2\yp
    \longrightarrow0
    \qquad\text{as }M\to\infty.
\]
A finite growth bound therefore limits individual particle growth and keeps
the mass cost strictly positive for any fixed feasible growth ratio $r>1$.

For two adjacent snapshots containing $n_0$ and $n_1$ cells, the observed average growth ratio is $n_1/n_0$.
We choose the maximum allowed growth ratio as a multiple of this average:
\[
    \e^M=\kappa\frac{n_1}{n_0},
    \qquad \kappa>1.
\]

Thus, the maximum growth ratio of a single particle is $\kappa$ times the population average $n_1/n_0$. To maintain the
assumption $M>1$ used in the derivation, our implementation sets
\[
    M=\max\left\{1.05,\,\log\left(\kappa\frac{n_1}{n_0}\right)\right\}.
\]
We use $\kappa=2$ as the default choice for most datasets, which allows individual particles to grow by up to twice the observed average population growth ratio. This is an empirical choice that provides room for heterogeneous growth while limiting extreme mass allocation. The values used for individual datasets are reported below.

\subsection{Evaluation Metrics}
\label{app:metrics}

We use two metrics to evaluate prediction results: the 1-Wasserstein distance ($\mathcal W_1$) to measure the difference between normalized predicted and observed distributions, and relative mass error (RME) to assess the accuracy of total mass prediction. They are defined as
\[
    \mathcal W_1(p,q)
    =\min_{\pi\in\Pi(p,q)}
    \int\|\bm x-\bm x'\|_2\,\dd\pi(\bm x,\bm x'),
\]
\[
    \mathrm{RME}(t_k)
    =\frac{\left|\sum_i m_i(t_k)-n_k\right|}{n_k}.
\]
Here $p$ and $q$ are the normalized predicted and observed empirical distributions, and $\Pi(p,q)$ is the set of couplings between them. The target cell count at time $t_k$ is denoted by $n_k$, and $m_i(t_k)$ is the predicted mass of particle $i$ at that time. Lower values of both metrics indicate better performance. We report RME only for methods that model mass variation.

For ESFM, each prediction starts from the previous observed snapshot with $m_i=1$,
corresponding to $y_i=0$, consistent with the source condition of the interval coupling.
We solve the learned ODE in the extended space to obtain the predicted cell states $\bm x_i(t_k)$
and log masses $y_i(t_k)$. The prediction in the original state space is the weighted particle
distribution $\sum_i \e^{y_i(t_k)}\Dirac_{\bm x_i(t_k)}$: each particle contributes at its predicted
cell state with mass $m_i(t_k)=\e^{y_i(t_k)}$. This is the weighted projection of the extended
distribution. We compute $\mathcal W_1$ between its normalized form and the normalized observed
distribution using POT, with distances measured in the cell state space. We compute RME by
comparing its total mass $\sum_i m_i(t_k)$ with the observed cell count $n_k$.
For the held-out experiments, we exclude one intermediate snapshot from training and predict it
from the previous observed snapshot. We report the mean $\mathcal W_1$ over the held-out time points.

To compare ESFM with other methods, we run the available reference implementations with method and dataset specific configurations.
 For WFR-FM \citep{peng2026wfrfm}, VGFM \citep{wang2025joint} and USB \citep{ying2026beyond}, we use their reference 
 configurations as the starting point and retain their respective network architectures and training procedures. 
 For MMFM \citep{rohbeck2025modeling}, Metric FM \citep{kapusniak2024metric}, SF2M \citep{sflowmatch} and UOT-FM 
 \citep{eyring2024unbalancedness}, we use the corresponding released model and matching implementations, with learning rates, 
 training budgets and sample sizes set separately for each method and dataset. Cosine annealing is used in WFR-FM, 
 USB and MMFM training scripts. For MIOFlow \citep{mioflow}, TIGON \citep{TIGON}, DeepRUOT \citep{DeepRUOT} and 
 Var-RUOT \citep{sun2025variational}, we retain the method specific optimization procedures, including local and
  global training in MIOFlow, density estimation in TIGON and the staged training pipeline in DeepRUOT. For methods with 
   stochastic inference, we report the mean and standard deviation over five inference runs using a fixed trained model.

\subsection{Synthetic Benchmarks}
\label{app:synthetic-benchmarks}

We evaluate ESFM on Simulation Gene, Saturating-Growth Data, Dyngen and a Gaussian mixture benchmark.
The aggregate prediction results are reported in \cref{tab:esfm-synthetic}; dataset descriptions and
 additional results are provided below. In the Dyngen prediction table, entries with $\pm$ report the mean and standard 
 deviation over five inference runs with different random seeds using a fixed trained model. 

\paragraph{Simulation Gene.}
The Simulation Gene dataset is generated from a synthetic gene regulatory network \citep{DeepRUOT}. The expression levels of three genes evolve according to
\begin{equation}
\begin{aligned}
\frac{\dd X_1}{\dd t}
&=
\frac{\alpha_1X_1^2+\beta}
{1+\alpha_1X_1^2+\gamma_2X_2^2+\gamma_3X_3^2+\beta}
-\delta_1X_1+\eta_1\xi_{1,t},\\
\frac{\dd X_2}{\dd t}
&=
\frac{\alpha_2X_2^2+\beta}
{1+\gamma_1X_1^2+\alpha_2X_2^2+\gamma_3X_3^2+\beta}
-\delta_2X_2+\eta_2\xi_{2,t},\\
\frac{\dd X_3}{\dd t}
&=
\frac{\alpha_3X_3^2}{1+\alpha_3X_3^2}
-\delta_3X_3+\eta_3\xi_{3,t}.
\end{aligned}
\label{eq:saturating-gene-dynamics}
\end{equation}
Here $X_i(t)$ is the expression level of gene $i$. Genes $X_1$ and $X_2$ form a toggle switch:
each gene activates itself and inhibits the other. The external signal $\beta$ activates both genes,
while $X_3$ inhibits them. The parameter $\alpha_i$ controls self-activation, $\gamma_i$ controls inhibition
and $\delta_i$ controls degradation. The term $\eta_i\xi_{i,t}$ represents
random noise with strength $\eta_i$. Cells divide randomly,
with a growth rate determined by $X_2$:
\[
    g(\bm X)=\alpha_g\frac{X_2^2}{1+X_2^2},
\]
where $\alpha_g$ controls the growth scale. After division, daughter cells inherit the parent's
gene expression state with small random perturbations. The simulation starts from two cell populations:
one changes its state and proliferates, while the other remains a steady state. This setting tests whether a
 model can distinguish population growth from cell state
transitions. Snapshots are collected at times $0$, $8$, $16$, $24$ and $32$. 
We use $X_1$ and $X_2$ as the model inputs and set the hyperparameters to
$\delta=2.0$, $\lambda=4.0$ and $\kappa=2.0$. The sensitivity analysis is reported in
\cref{tab:esfm-sensitivity-gene}. The results show that $\delta$ affects prediction performance,
whereas the tested values of $\kappa$ and $\lambda$ give similar results.

\begin{table}[t]
\centering
\caption{Sensitivity analysis of $\delta$, $\lambda$ and $\kappa$ on the Simulation Gene dataset.}
\label{tab:esfm-sensitivity-gene}
\resizebox{0.55\textwidth}{!}{
\begin{tabular}{lccc}
\toprule
\textbf{Parameter} & \textbf{Value} & $\mathcal W_1$ ($\downarrow$) & RME ($\downarrow$) \\
\midrule
$\delta$ & 1.0 & 0.027 & 0.008 \\
$\delta$ & 2.0 & 0.018 & 0.001 \\
$\delta$ & 3.0 & 0.024 & 0.013 \\
\midrule
$\kappa$ & 1.2 & 0.018 & 0.001 \\
$\kappa$ & 1.5 & 0.018 & 0.001 \\
$\kappa$ & 3.0 & 0.018 & 0.001 \\
\midrule
$\lambda$ & 1.0 & 0.018 & 0.001 \\
$\lambda$ & 4.0 & 0.018 & 0.001 \\
$\lambda$ & 8.0 & 0.017 & 0.001 \\
\bottomrule
\end{tabular}
}
\end{table}

\paragraph{Saturating-Growth Data.}
\label{app:saturating-growth-data}
This variant follows the same gene regulatory dynamics as Simulation Gene in
\cref{eq:saturating-gene-dynamics}, but replaces the growth law with
\begin{equation*}
    g(\bm X)
    =\alpha_g\frac{X_2^8}{0.2^8+X_2^8},
\end{equation*}
where $\alpha_g$ denotes the maximum growth rate. The growth rate increases with $X_2$ and then stays close to $\alpha_g$. One 
population grows at a nearly constant rate, while the other shows almost no growth. This experiment tests whether ESFM can 
identify which cells grow and accurately predict their growth rates. Snapshots are collected at times $0$, $4$, $8$, $16$ and $32$. We retain all three gene coordinates for training and
use PCA for visualization. We set the hyperparameters to $\delta=2.0$, $\lambda=1.0$ and $\kappa=2.0$. 
To compute the coefficient of determination $R^2$ reported in Q1, we pool all $N=2772$ observed cells across
the five snapshots and evaluate each cell at its own observation time $t_i$.
Let $(\bm X_{j,k},Y_{j,k})$ denote the resulting particles at $t_k$.
At the initial time, the particles are the observed source cells with $Y_{j,0}=0$.

\noindent\textbf{Physical growth and its particle estimate.}
The projection in \cref{thm:extended-projection} defines the physical density
and growth field as
\begin{equation*}
\begin{aligned}
    \rho_t(\bm x)
    &=\int_{\R}\e^y\hat\rho_t(\bm x,y)\dd y,
    \\
    g_t(\bm x)
    &=\frac{
        \int_{\R}\hat g_t(\bm x,y)\hat\rho_t(\bm x,y)\e^y\dd y
    }{\rho_t(\bm x)}.
\end{aligned}
\end{equation*}
The ODE provides particles $\{(\bm X_j(t),Y_j(t))\}_{j=1}^{n}$ whose states
generally do not coincide with the observed cells. To evaluate growth at an
observed state $\bm x$, we use nearby particles to form the local kernel estimate
\[
    \hat\rho_t^{\,h}(\bm x,\dd y)
    =\frac1n\sum_{j=1}^{n}
    K_h(\bm x-\bm X_j(t))\Dirac_{Y_j(t)}(\dd y),
    \qquad
    K_h(\bm z)=\exp\!\left(-\frac{\norm{\bm z}^2}{2h^2}\right).
\]
The kernel acts only on the state coordinates; common normalization factors
cancel in the ratio. Substituting this local measure into the projection and
using the learned growth field $g_\phi$ gives
\[
\begin{aligned}
    \widehat g_t^{\,h}(\bm x)
    &:=\frac{
        \int_{\R}g_\phi(\bm x,y,t)\e^y\hat\rho_t^{\,h}(\bm x,\dd y)
    }{\int_{\R}\e^y\hat\rho_t^{\,h}(\bm x,\dd y)}
    \\
    &=\frac{
        \sum_j K_h(\bm x-\bm X_j(t))\e^{Y_j(t)}
        g_\phi(\bm x,Y_j(t),t)
    }{
        \sum_j K_h(\bm x-\bm X_j(t))\e^{Y_j(t)}
    }.
\end{aligned}
\]
We set $h$ to the distance to the 16th nearest generated state, without using
growth labels, and include all particles in the estimate.
WFR-FM is evaluated directly through its physical growth head at $(\bm x,t_k)$.
Let $g_i=g(\bm X_i)$ be its true
growth rate and $\widehat g_i$ its predicted rate. We compute
\[
    R^2=1-\frac{\sum_{i=1}^{N}(\widehat g_i-g_i)^2}
    {\sum_{i=1}^{N}(g_i-\bar g)^2},
    \qquad
    \bar g=\frac{1}{N}\sum_{i=1}^{N}g_i.
\]
As shown in \cref{fig:saturating-growth-g-recovery}, ESFM predicts growth rates close to the true values for both populations. 
WFR-FM predicts lower growth rates, including negative values for some cells even though their true growth rates are nonnegative.

{\color{black}%
\noindent\textbf{Additional growth-field comparisons.}
We extend the comparison to VGFM, DeepRUOT and Var-RUOT, using the same dataset.
ESFM is evaluated through the physical growth projection described above;
the baselines are evaluated through their physical growth heads, expressed
in the same time units. Growth labels are used only for evaluation.
\Cref{fig:app-saturating-growth-comparison} shows the predicted fields.
ESFM captures both the proliferative and
approximately quiescent populations. VGFM predicts appreciable positive
growth in part of the approximately quiescent population.

\begin{figure}[H]
    \centering
    \includegraphics[width=\textwidth]{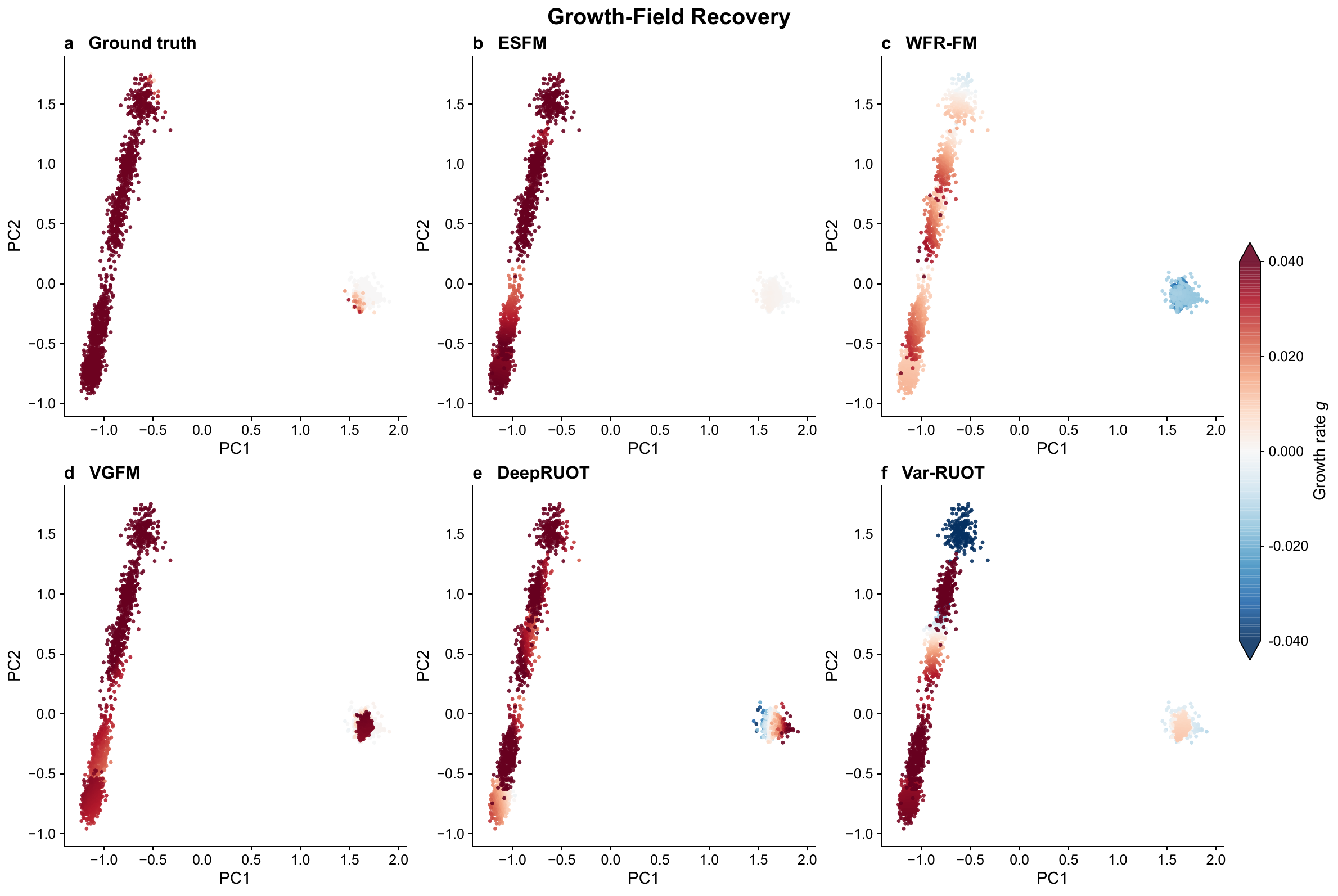}
    \caption{\textcolor{black}{Growth-field recovery on Saturating-Growth Data using all five
    observed snapshots. Ground truth and the predictions of ESFM, WFR-FM,
    VGFM, DeepRUOT and Var-RUOT are shown on the same 2,772 cell coordinates.}}
    \label{fig:app-saturating-growth-comparison}
\end{figure}

\noindent\textbf{Growth recovery on held-out snapshots.}
For each method, we mask the snapshots at $t=4,8,16$ one at a time and train
a separate hold-out model for each masked time point using the remaining
snapshots. We compare the predicted growth rates at these time points with
the ground truth and display the results together in
\cref{fig:app-saturating-heldout-growth}. The results show that ESFM recovers
the ground-truth growth rates at the held-out time points.

\begin{figure}[H]
    \centering
    \includegraphics[width=\textwidth]{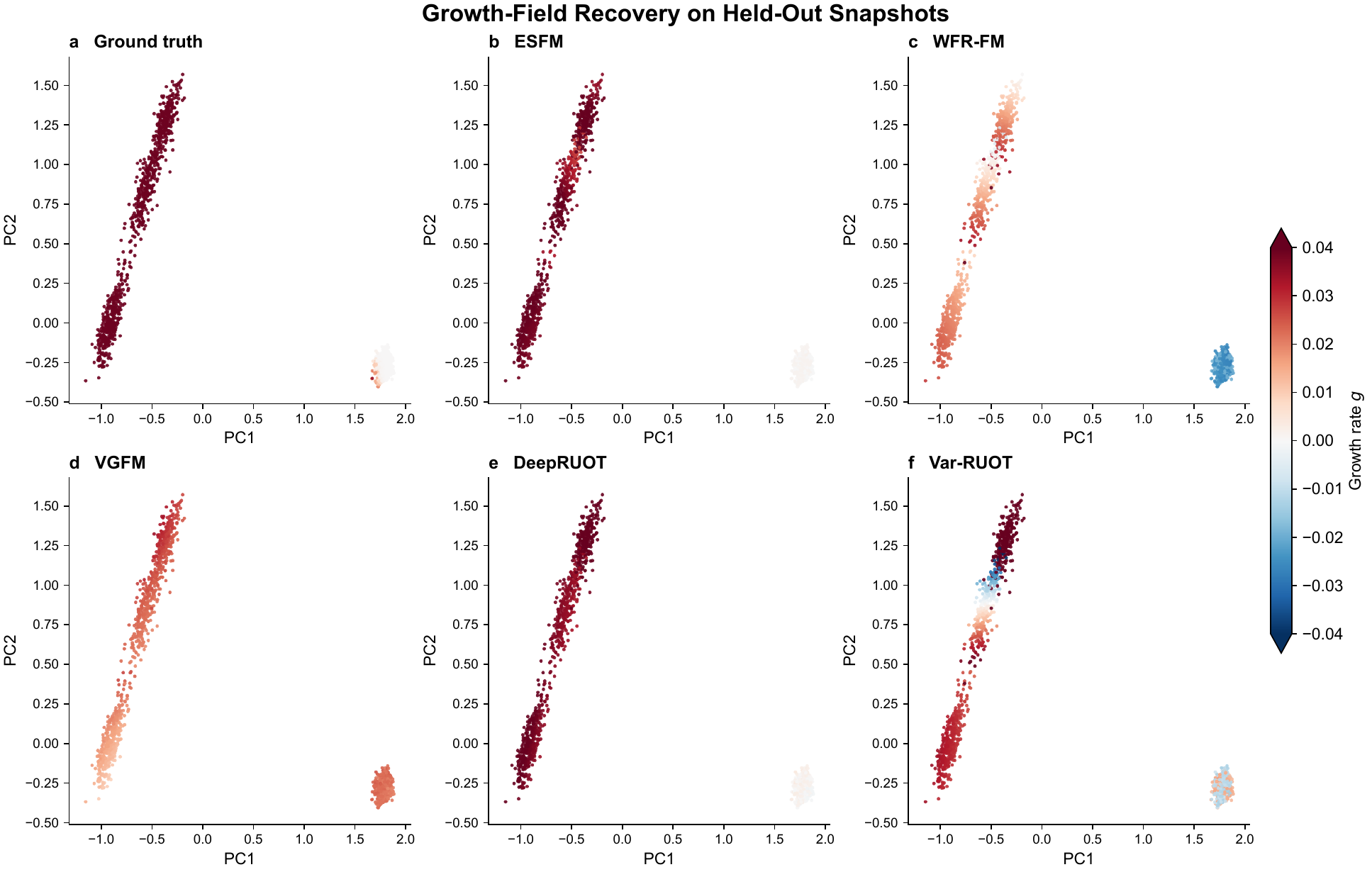}
    \caption{\textcolor{black}{Growth-field recovery on held-out snapshots at $t=4,8,16$.
    Each prediction uses a model trained without the evaluated snapshot.}}
    \label{fig:app-saturating-heldout-growth}
\end{figure}

\noindent\textbf{Trajectory recovery on held-out snapshots.}
Using the same three hold-out models trained by masking $t=4,8,16$ one at
a time, we predict trajectories and cell positions at the corresponding
held-out time points. We compare the predicted positions with the observed
snapshots and display the results together with the model-generated
trajectories in \cref{fig:app-saturating-heldout-trajectories}.

\begin{figure}[H]
    \centering
    \includegraphics[width=\textwidth]{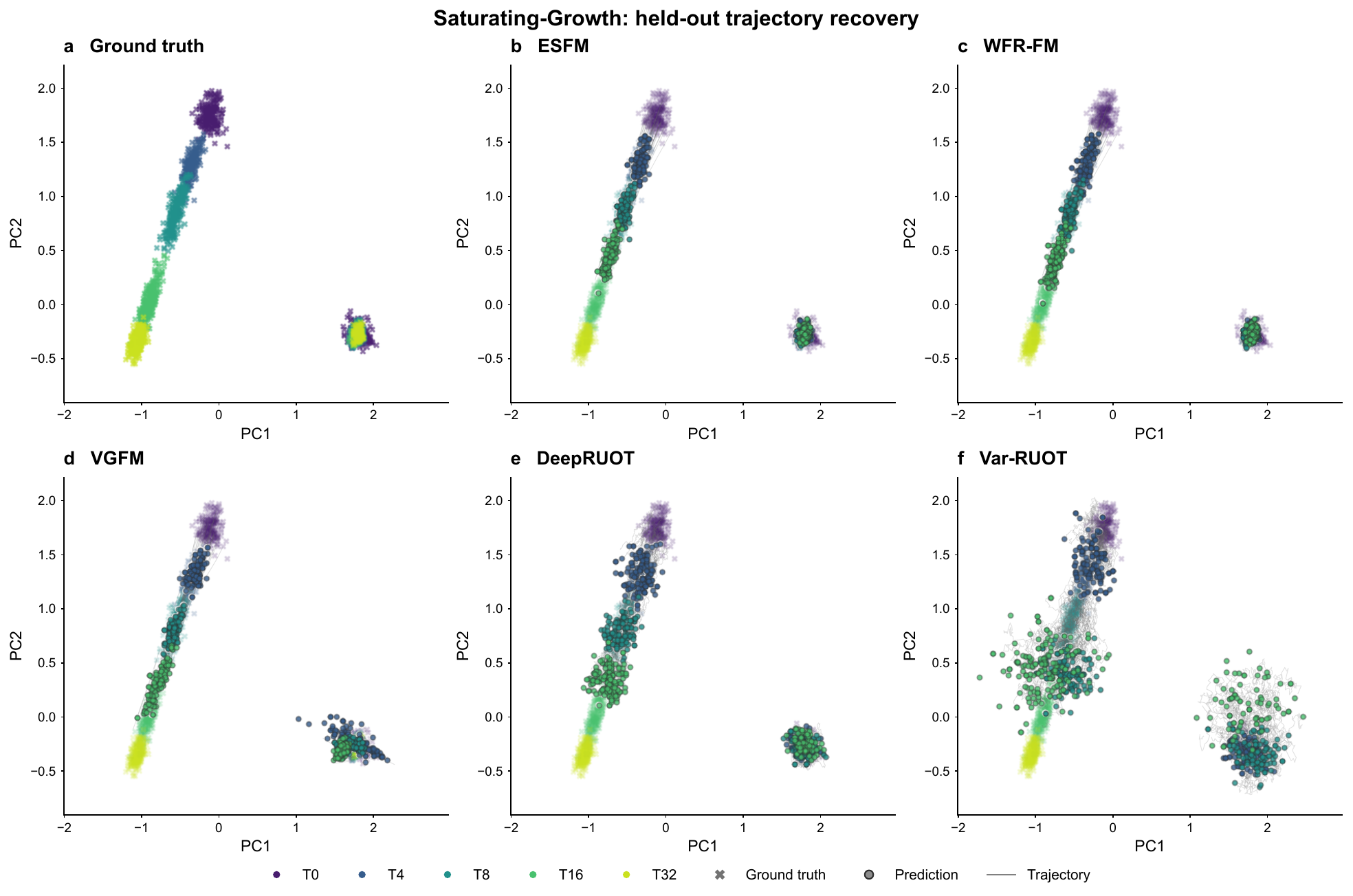}
    \caption{\textcolor{black}{Held-out trajectory and marginal recovery on Saturating-Growth Data.}}
    \label{fig:app-saturating-heldout-trajectories}
\end{figure}

}%
\paragraph{Dyngen.}
The Dyngen dataset is generated using the dyngen single-cell simulator \citep{cannoodt2021spearheading} and contains 728 cells across five snapshots. The simulated 
trajectories bifurcate into two branches with unequal cell abundance, providing a test of joint state and mass prediction. 
We use a five-dimensional PHATE representation \citep{moon2019visualizing} and set the hyperparameters to $\delta=2.5$, 
$\lambda=6.25$ and $\kappa=1.5$. The prediction results in \cref{tab:esfm-synthetic-dygen-detailed} show that ESFM achieves the 
lowest distribution and mass errors at most prediction times. 
\begin{table}[H]
\centering
\caption{Comparison of method performance over time on the Dyngen dataset.}
\label{tab:esfm-synthetic-dygen-detailed}
\resizebox{\textwidth}{!}{
\begin{tabular}{lcccccccc}
\toprule
\textbf{Method} & \multicolumn{2}{c}{$t=1$} & \multicolumn{2}{c}{$t=2$} & \multicolumn{2}{c}{$t=3$} & \multicolumn{2}{c}{$t=4$} \\
\cmidrule(lr){2-3}\cmidrule(lr){4-5}\cmidrule(lr){6-7}\cmidrule(lr){8-9}
& $\mathcal W_1$ & RME & $\mathcal W_1$ & RME & $\mathcal W_1$ & RME & $\mathcal W_1$ & RME \\
\midrule
MMFM & 0.574 & --- & 1.704 & --- & 1.499 & --- & 1.706 & --- \\
Metric FM & 0.892 & --- & 2.347 & --- & 2.030 & --- & 1.799 & --- \\
SF2M & 0.637 & --- & 1.266 & --- & 1.415 & --- & 1.790 & --- \\
MIOFlow & 0.420 & --- & 0.640 & --- & 1.537 & --- & 1.263 & --- \\
TIGON & 0.446 & 0.033 & 0.584 & 0.060 & 0.415 & 0.023 & 0.603 & 0.071 \\
DeepRUOT & 0.454 & 0.011 & 0.481 & 0.070 & 0.870 & 0.104 & 0.688 & 0.074 \\
Var-RUOT & 0.315 & 0.128 & 0.548 & 0.336 & 0.630 & 0.222 & 0.593 & 0.023 \\
UOT-FM & 0.652 & 0.008 & 0.780 & 0.077 & 1.252 & 0.090 & 2.130 & 0.213 \\
VGFM & 0.335 & \second{0.001} & 0.312 & 0.073 & 1.109 & 0.041 & 0.634 & 0.033 \\
WFR-FM & \second{0.110} & 0.003 & \second{0.098} & \second{0.007} & 0.211 & \second{0.008} & \second{0.121} & \second{0.002} \\
USB & \best{0.109}{\tiny$\pm 0.0002$} & 0.002{\tiny$\pm 0.0000$} & \best{0.093}{\tiny$\pm 0.0002$} & \best{0.002}{\tiny$\pm 0.0001$} & \second{0.180}{\tiny$\pm 0.0013$} & 0.015{\tiny$\pm 0.0004$} & 0.142{\tiny$\pm 0.0040$} & 0.008{\tiny$\pm 0.0008$} \\
\textbf{ESFM (ours)} & 0.135 & \best{0.0002} & \best{0.093} & 0.008 & \best{0.132} & \best{0.0028} & \best{0.085} & \best{0.0003} \\
\bottomrule
\end{tabular}
}
\end{table}

{\color{black}%
As an additional visualization, we realize the learned population dynamics on the Dyngen dataset through a birth--death process.
Cells follow the learned velocity field, with division and apoptosis events
determined by the inferred growth rates. As shown in
\cref{fig:app-dyngen-birth-death}, this realization produces explicit division
and death events, complementing the continuous population-level representation.

\begin{figure}[H]
    \centering
    \includegraphics[width=0.65\textwidth]{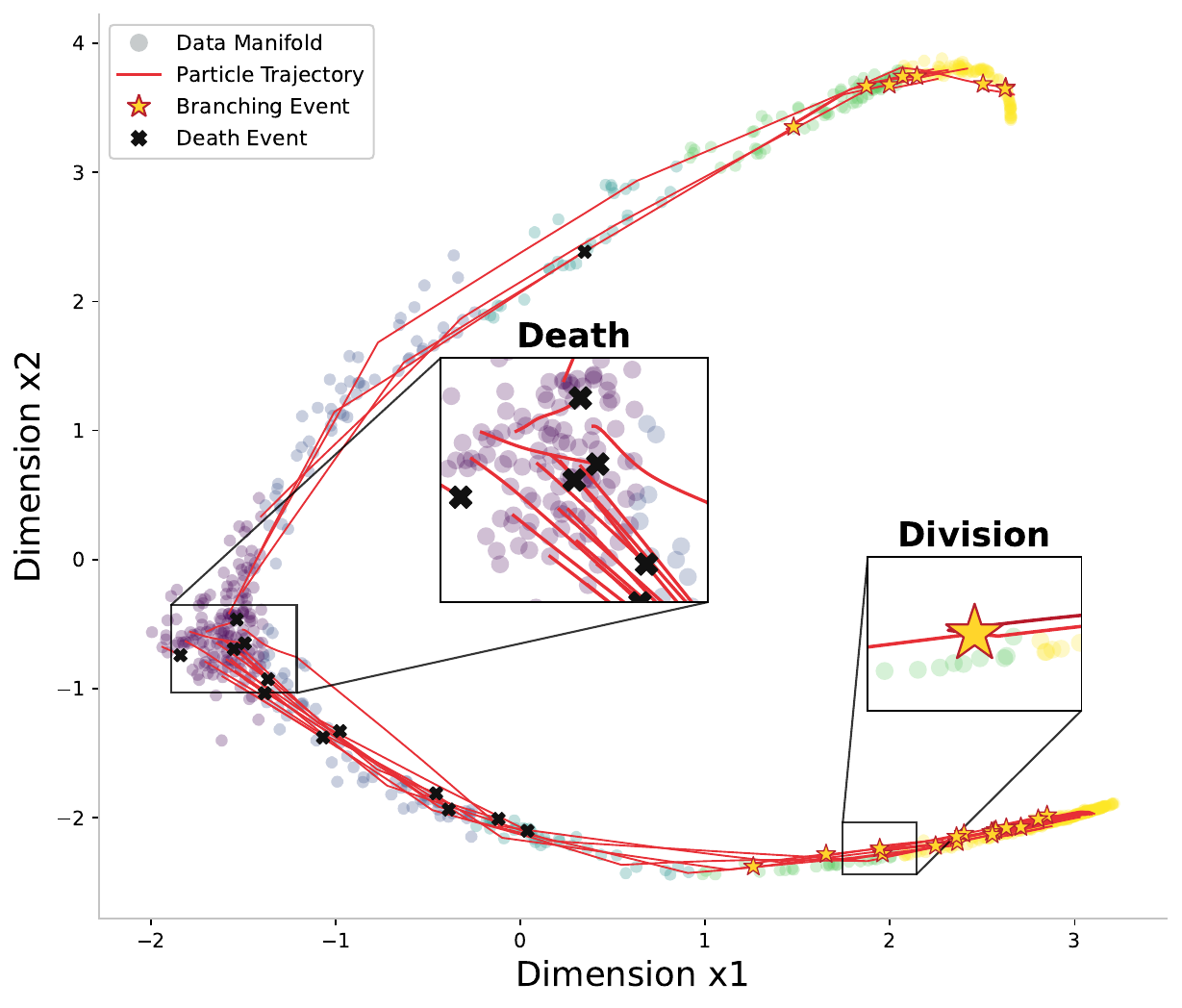}
    \caption{\textcolor{black}{\textbf{Birth--death simulation on the Dyngen dataset.}
    Discrete particle trajectories generated from the learned ESFM dynamics, with
 division and apoptosis events determined by the inferred growth rates.}}
    \label{fig:app-dyngen-birth-death}
\end{figure}
}%

\begin{figure}[H]
    \centering
    \includegraphics[width=0.6\textwidth]{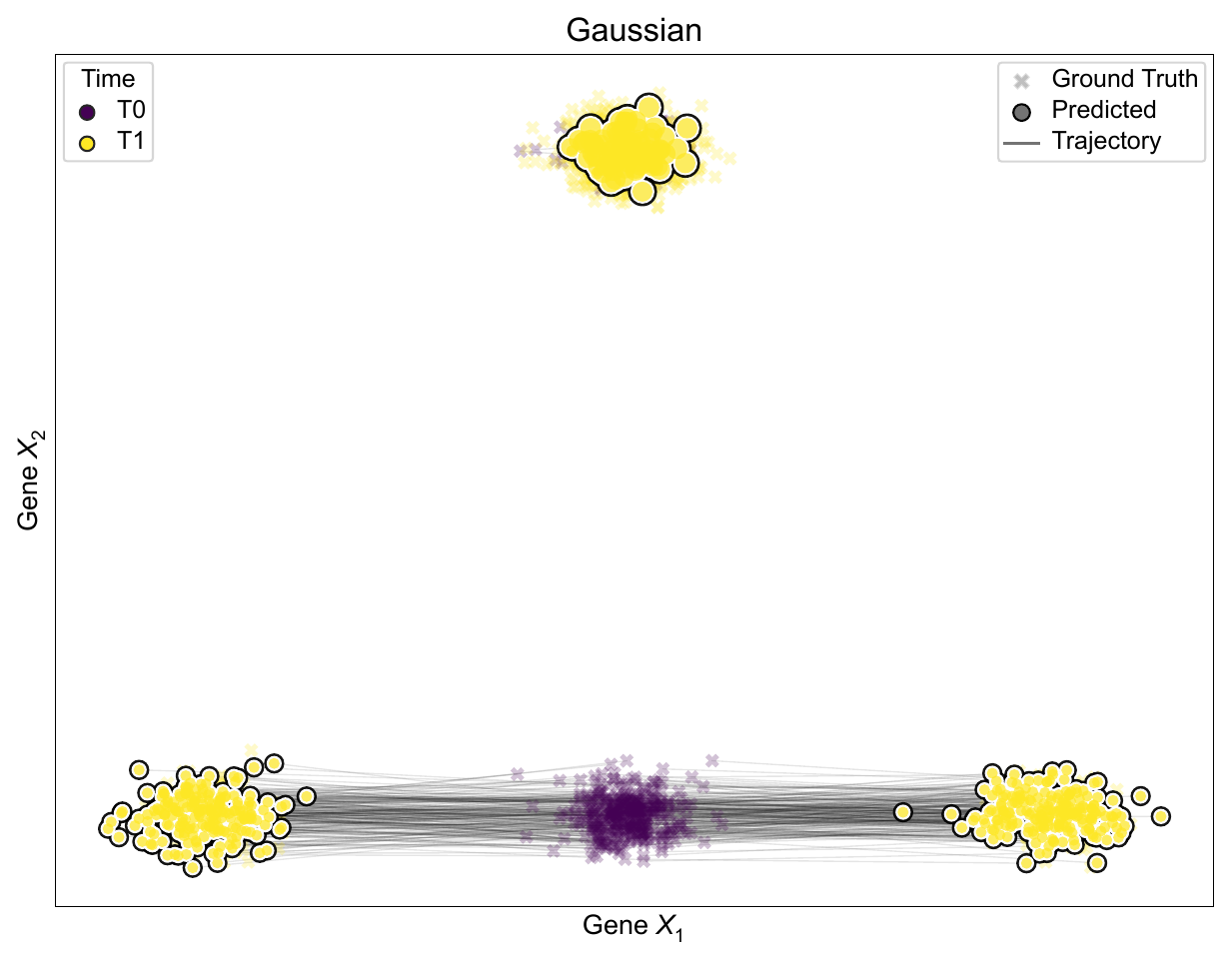}
    \caption{ESFM trajectory prediction on the 1000D Gaussian mixture benchmark.}
    \label{fig:esfm-gaussian-trajectory}
\end{figure}

\paragraph{Gaussian mixture.}
The Gaussian mixture benchmark contains two snapshots in 1000 dimensions \citep{wang2025joint}. The source contains 500 samples, with 100 in an upper component and 400 in a lower component. In the target, the upper component expands to 1000 samples, while the lower population splits into two components of 200 samples each. The benchmark therefore combines population growth with branching transport in a high dimensional state space. We set the hyperparameters to $\delta=1.2$, $\lambda=1.44$ and $\kappa=4.0$. As shown in \cref{fig:esfm-gaussian-trajectory}, the predicted endpoints capture the three target modes, with little movement in the upper population and splitting trajectories in the lower population.

\subsection{Real Single-Cell Benchmarks}
\label{app:real-benchmarks}

We evaluate ESFM on four real single-cell datasets: EMT, embryoid body differentiation (EB), CITE-seq, and mouse hematopoiesis. The aggregate held-out results are shown in \cref{tab:esfm-real}. Detailed dataset descriptions and prediction results are provided below.
For the prediction tables below, entries with $\pm$ report the mean and standard deviation over five inference runs with different 
random seeds using a fixed trained model. 

{\color{black}%

\noindent\textbf{Additional held-out growth and trajectory visualizations.}
We supplement the held-out experiments on EMT, EB, CITE-seq and Mouse with
ESFM growth-field visualizations and trajectory comparisons with WFR-FM,
VGFM, DeepRUOT and Var-RUOT. The results are presented below for each dataset.

}%

\paragraph{EMT.}
The epithelial--mesenchymal transition (EMT) dataset contains A549 lung cancer cells collected at four time points during the
 transition \citep{cook2020context}. We use an autoencoder to reduce the gene expression data to 10 dimensions.
  These snapshots capture changes in both transcriptional state and observed cell abundance.
  We evaluate ESFM on this dataset with hyperparameters set to $\delta=1.5$, $\lambda=2.25$, and $\kappa=2.0$. As shown in \cref{tab:esfm-emt-detailed}, ESFM achieves
   the lowest distribution error at $t=2,3$ and the second-lowest at $t=1$, while maintaining low mass error.

\begin{table}[H]
\centering
\caption{Comparison of method performance over time on the 10D EMT dataset.}
\label{tab:esfm-emt-detailed}
\resizebox{\textwidth}{!}{
\begin{tabular}{lcccccc}
\toprule
\textbf{Method} & \multicolumn{2}{c}{$t=1$} & \multicolumn{2}{c}{$t=2$} & \multicolumn{2}{c}{$t=3$} \\
\cmidrule(lr){2-3}\cmidrule(lr){4-5}\cmidrule(lr){6-7}
& $\mathcal W_1$ & RME & $\mathcal W_1$ & RME & $\mathcal W_1$ & RME \\
\midrule
MMFM & 0.258 & --- & 0.287 & --- & 0.310 & --- \\
Metric FM & 0.261 & --- & 0.297 & --- & 0.305 & --- \\
SF2M & 0.257 & --- & 0.281 & --- & 0.290 & --- \\
MIOFlow & 0.244 & --- & 0.267 & --- & 0.284 & --- \\
TIGON & 0.243 & 0.002 & 0.266 & 0.003 & 0.285 & \second{0.001} \\
DeepRUOT & 0.290 & \second{0.001} & 0.319 & 0.011 & 0.329 & 0.002 \\
Var-RUOT & 0.254 & 0.075 & 0.267 & 0.014 & 0.268 & 0.041 \\
UOT-FM & 0.254 & 0.002 & 0.270 & 0.013 & 0.277 & 0.010 \\
VGFM & 0.235 & 0.016 & 0.242 & 0.011 & 0.245 & 0.018 \\
WFR-FM & 0.210 & \second{0.001} & 0.227 & \second{0.002} & 0.235 & \second{0.001} \\
USB & \best{0.183}{\tiny$\pm 0.000$} & \best{0.000}{\tiny$\pm 0.000$} & \second{0.216}{\tiny$\pm 0.000$} & 0.009{\tiny$\pm 0.000$} & \second{0.231}{\tiny$\pm 0.000$} & 0.009{\tiny$\pm 0.000$} \\
\textbf{ESFM (ours)} & \second{0.197} & \best{0.000} & \best{0.189} & \best{0.001} & \best{0.198} & \best{0.000} \\
\bottomrule
\end{tabular}
}
\end{table}

{\color{black}%

\noindent\textbf{Held-out visualizations for EMT.}
\Cref{fig:app-emt-heldout-growth} shows ESFM's physical growth predictions
pooled over the held-out snapshots at $t=1,2$.
Using the corresponding hold-out models, we predict trajectories and cell
positions at these time points. We compare the predicted positions with the
observed snapshots and display the results together with the model-generated
trajectories in \cref{fig:app-emt-heldout-trajectories}.

\begin{figure}[H]
    \centering
    \begin{subfigure}[b]{0.32\textwidth}
        \centering
        \includegraphics[width=\linewidth]{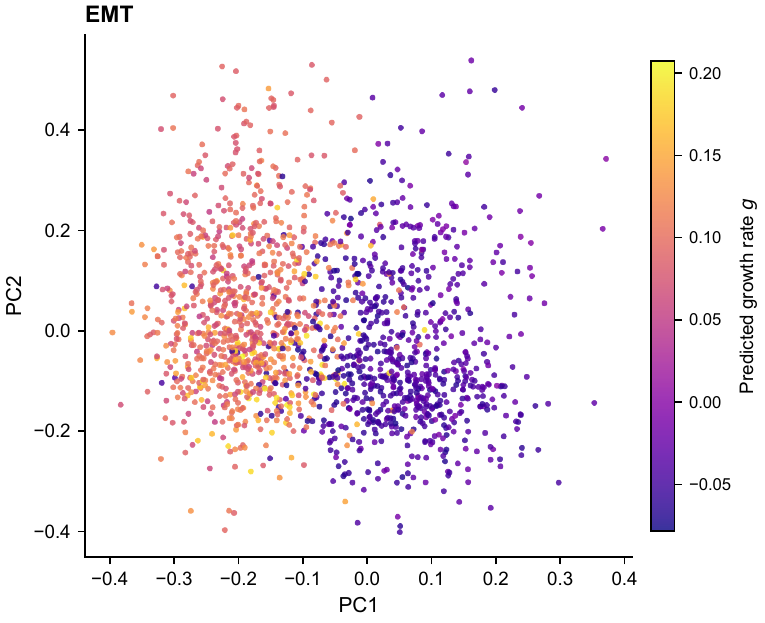}
        \caption{\textcolor{black}{ESFM growth predictions.}}
        \label{fig:app-emt-heldout-growth}
    \end{subfigure}\hfill
    \begin{subfigure}[b]{0.66\textwidth}
        \centering
        \includegraphics[width=\linewidth]{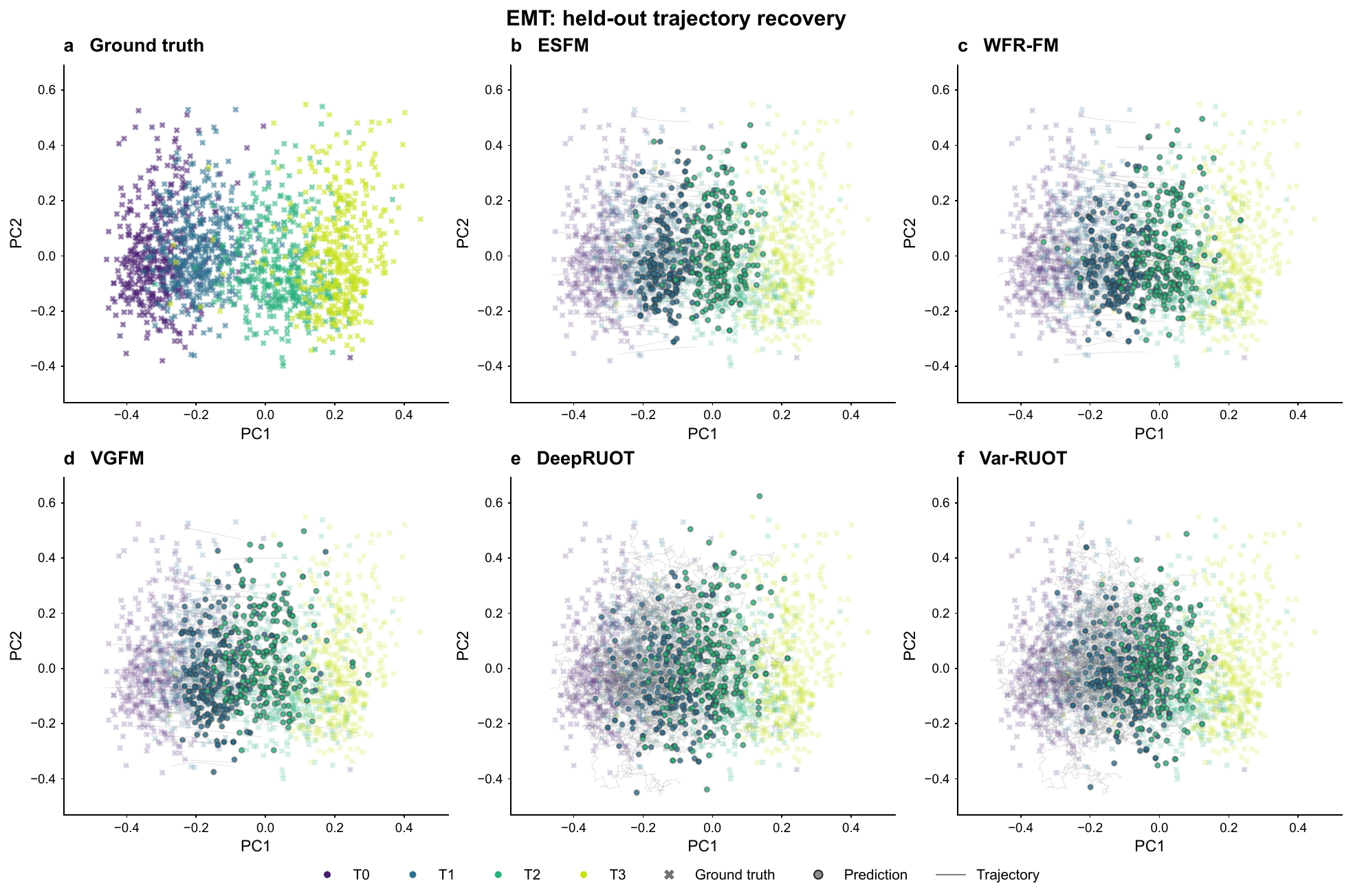}
        \caption{\textcolor{black}{Trajectory and marginal comparison.}}
        \label{fig:app-emt-heldout-trajectories}
    \end{subfigure}
    \caption{\textcolor{black}{Growth-field and trajectory predictions on the held-out 10D EMT
    snapshots at $t=1,2$.}}
    \label{fig:app-emt-heldout-comparison}
\end{figure}
}%

\paragraph{Embryoid body.}
The human embryoid body (EB) dataset contains 16,819 cells sampled at five time points over a 27-day differentiation experiment
 \citep{moon2019visualizing}. This experimental system models early embryonic development and captures the emergence of diverse
  cell states. We evaluate PCA representations with 5, 50 and 100 dimensions to examine prediction performance at different
  dimensions. The hyperparameter settings are $\delta=2.0$, $\lambda=1.0$ and $\kappa=2.0$ for 5D;
    $\delta=24.0$, $\lambda=5.0$ and $\kappa=2.0$ for 50D; and $\delta=18.0$, $\lambda=3.0$ and $\kappa=2.0$ for 100D.
    The prediction results in \cref{tab:esfm-eb-5d,tab:esfm-eb-50d,tab:esfm-eb-100d} show that ESFM achieves the lowest 
    $\mathcal W_1$ across all tested dimensions and prediction times. It also achieves the lowest mass error at three of the four prediction times in the 5D representation.

\begin{table}[H]
\centering
\caption{Comparison of method performance over time on the 5D EB dataset.}
\label{tab:esfm-eb-5d}
\resizebox{\textwidth}{!}{
\begin{tabular}{lcccccccc}
\toprule
\textbf{Method} & \multicolumn{2}{c}{$t=1$} & \multicolumn{2}{c}{$t=2$} & \multicolumn{2}{c}{$t=3$} & \multicolumn{2}{c}{$t=4$} \\
\cmidrule(lr){2-3}\cmidrule(lr){4-5}\cmidrule(lr){6-7}\cmidrule(lr){8-9}
& $\mathcal W_1$ & RME & $\mathcal W_1$ & RME & $\mathcal W_1$ & RME & $\mathcal W_1$ & RME \\
\midrule
MMFM & 0.477 & --- & 0.554 & --- & 0.781 & --- & 0.872 & --- \\
Metric FM & 0.449 & --- & 0.552 & --- & 0.583 & --- & 0.597 & --- \\
SF2M & 0.556 & --- & 0.715 & --- & 0.750 & --- & 0.650 & --- \\
MIOFlow & 0.442 & --- & 0.585 & --- & 0.651 & --- & 0.670 & --- \\
TIGON & 0.386 & \best{0.002} & 0.502 & 0.015 & 0.602 & 0.021 & 0.600 & 0.027 \\
DeepRUOT & 0.386 & 0.005 & 0.497 & 0.017 & 0.591 & 0.021 & 0.585 & 0.030 \\
Var-RUOT & 0.416 & 0.111 & 0.486 & 0.144 & 0.509 & 0.054 & 0.511 & 0.022 \\
UOT-FM & 0.544 & 0.032 & 0.670 & 0.029 & 0.729 & 0.016 & 0.852 & 0.041 \\
VGFM & 0.402 & 0.046 & 0.494 & 0.018 & 0.525 & 0.035 & 0.573 & 0.021 \\
WFR-FM & \second{0.324} & \second{0.003} & \second{0.401} & \second{0.001} & 0.431 & \second{0.005} & 0.510 & \second{0.005} \\
USB & 0.331{\tiny$\pm 2\times 10^{-5}$} & 0.006{\tiny$\pm 2\times 10^{-6}$} & 0.408{\tiny$\pm 2\times 10^{-5}$} & 0.023{\tiny$\pm 4\times 10^{-6}$} & \second{0.427}{\tiny$\pm 2\times 10^{-5}$} & 0.018{\tiny$\pm 6\times 10^{-6}$} & \second{0.450}{\tiny$\pm 4\times 10^{-5}$} & 0.019{\tiny$\pm 1\times 10^{-5}$} \\
\textbf{ESFM (ours)} & \best{0.313} & 0.005 & \best{0.310} & \best{0.000} & \best{0.330} & \best{0.003} & \best{0.343} & \best{0.001} \\
\bottomrule
\end{tabular}
}
\end{table}

\begin{table}[H]
\centering
\caption{Comparison of method performance over time on the 50D EB dataset.}
\label{tab:esfm-eb-50d}
\resizebox{\textwidth}{!}{
\begin{tabular}{lcccccccc}
\toprule
\textbf{Method} & \multicolumn{2}{c}{$t=1$} & \multicolumn{2}{c}{$t=2$} & \multicolumn{2}{c}{$t=3$} & \multicolumn{2}{c}{$t=4$} \\
\cmidrule(lr){2-3}\cmidrule(lr){4-5}\cmidrule(lr){6-7}\cmidrule(lr){8-9}
& $\mathcal W_1$ & RME & $\mathcal W_1$ & RME & $\mathcal W_1$ & RME & $\mathcal W_1$ & RME \\
\midrule
MMFM & 9.124 & --- & 10.474 & --- & 11.022 & --- & 11.480 & --- \\
Metric FM & 8.506 & --- & 9.795 & --- & 10.621 & --- & 12.042 & --- \\
SF2M & 9.247 & --- & 10.882 & --- & 11.650 & --- & 12.154 & --- \\
MIOFlow & 8.447 & --- & 9.229 & --- & 9.436 & --- & 10.123 & --- \\
TIGON & 8.433 & 0.067 & 9.275 & 0.022 & 9.802 & 0.179 & 10.148 & 0.101 \\
DeepRUOT & 8.169 & \best{0.003} & 9.049 & 0.038 & 9.378 & 0.088 & 9.733 & \best{0.004} \\
Var-RUOT & 9.442 & 0.128 & 9.709 & 0.081 & 10.482 & 0.031 & 10.735 & 0.030 \\
UOT-FM & 8.717 & 0.063 & 10.858 & \second{0.009} & 11.813 & 0.022 & 12.733 & 0.018 \\
VGFM & 7.951 & 0.089 & 8.747 & 0.042 & 9.244 & 0.019 & \second{9.620} & 0.044 \\
WFR-FM & \second{7.664} & \second{0.008} & \second{8.659} & \best{0.006} & \second{9.182} & \best{0.004} & 9.914 & \best{0.004} \\
USB & 7.992{\tiny$\pm 2\times 10^{-5}$} & \second{0.008}{\tiny$\pm 7\times 10^{-7}$} & 8.973{\tiny$\pm 5\times 10^{-5}$} & 0.019{\tiny$\pm 1\times 10^{-6}$} & 9.421{\tiny$\pm 2\times 10^{-5}$} & 0.008{\tiny$\pm 2\times 10^{-6}$} & 9.963{\tiny$\pm 2\times 10^{-5}$} & 0.017{\tiny$\pm 3\times 10^{-6}$} \\
\textbf{ESFM (ours)} & \best{7.419} & 0.013 & \best{7.634} & 0.040 & \best{7.716} & \second{0.007} & \best{7.918} & \second{0.015} \\
\bottomrule
\end{tabular}
}
\end{table}

\begin{table}[H]
\centering
\caption{Comparison of method performance over time on the 100D EB dataset.}
\label{tab:esfm-eb-100d}
\resizebox{\textwidth}{!}{
\begin{tabular}{lcccccccc}
\toprule
\textbf{Method} & \multicolumn{2}{c}{$t=1$} & \multicolumn{2}{c}{$t=2$} & \multicolumn{2}{c}{$t=3$} & \multicolumn{2}{c}{$t=4$} \\
\cmidrule(lr){2-3}\cmidrule(lr){4-5}\cmidrule(lr){6-7}\cmidrule(lr){8-9}
& $\mathcal W_1$ & RME & $\mathcal W_1$ & RME & $\mathcal W_1$ & RME & $\mathcal W_1$ & RME \\
\midrule
MMFM & 11.460 & --- & 13.879 & --- & 14.441 & --- & 14.907 & --- \\
Metric FM & 10.806 & --- & 12.348 & --- & 13.622 & --- & 16.801 & --- \\
SF2M & 11.333 & --- & 12.982 & --- & 13.718 & --- & 14.945 & --- \\
MIOFlow & 11.387 & --- & 12.331 & --- & 11.905 & --- & 12.908 & --- \\
TIGON & 10.547 & 0.014 & 12.926 & 0.052 & 13.897 & 0.107 & 14.945 & 0.096 \\
DeepRUOT & 10.256 & \best{0.002} & 11.103 & 0.074 & 11.529 & 0.136 & \second{12.406} & 0.047 \\
Var-RUOT & 11.746 & 0.091 & 12.237 & 0.024 & 12.957 & 0.150 & 13.335 & 0.074 \\
UOT-FM & 10.757 & 0.056 & 12.799 & 0.037 & 13.761 & 0.044 & 15.657 & 0.022 \\
VGFM & 10.313 & 0.048 & 11.278 & 0.035 & 11.703 & 0.028 & 12.637 & 0.066 \\
WFR-FM & \second{9.941} & \second{0.009} & \second{11.040} & \best{0.006} & \second{11.516} & \second{0.008} & 12.664 & \second{0.005} \\
USB & 10.206{\tiny$\pm 5\times 10^{-5}$} & 0.016{\tiny$\pm 5\times 10^{-7}$} & 11.297{\tiny$\pm 6\times 10^{-5}$} & \second{0.010}{\tiny$\pm 8\times 10^{-7}$} & 11.758{\tiny$\pm 9\times 10^{-6}$} & \best{0.002}{\tiny$\pm 2\times 10^{-6}$} & 12.888{\tiny$\pm 5\times 10^{-5}$} & \best{0.003}{\tiny$\pm 3\times 10^{-6}$} \\
\textbf{ESFM (ours)} & \best{9.825} & 0.020 & \best{10.728} & 0.014 & \best{11.130} & 0.009 & \best{11.768} & \second{0.005} \\
\bottomrule
\end{tabular}
}
\end{table}

{\color{black}%
\noindent\textbf{Held-out visualizations for EB.}
We use the 50D representation for the additional visualizations.
\Cref{fig:app-eb-heldout-growth} pools ESFM growth predictions from the three
models withholding $t=1,2,3$, respectively.
Using the corresponding hold-out models, we predict trajectories and cell
positions at these time points. We compare the predicted positions with the
observed snapshots and display the results together with the model-generated
trajectories in \cref{fig:app-eb-heldout-trajectories}.

\begin{figure}[H]
    \centering
    \begin{subfigure}[b]{0.32\textwidth}
        \centering
        \includegraphics[width=\linewidth]{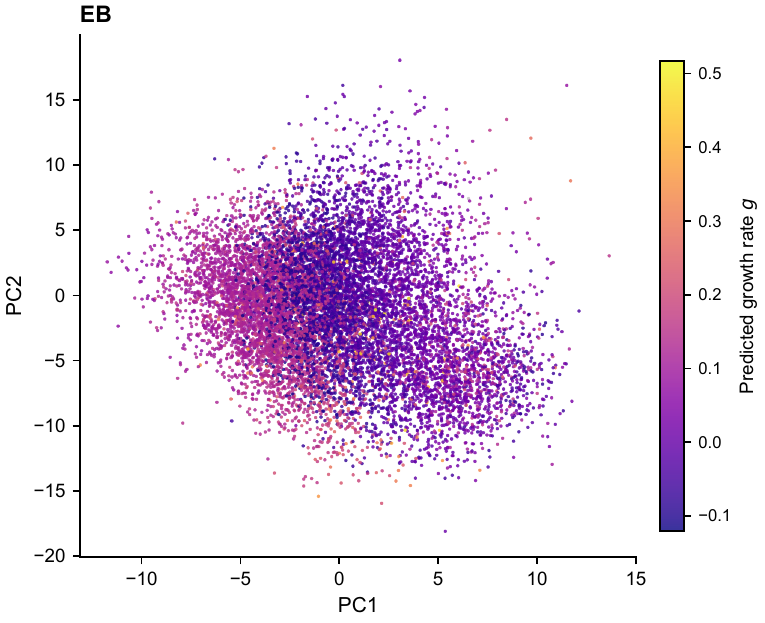}
        \caption{\textcolor{black}{ESFM growth predictions.}}
        \label{fig:app-eb-heldout-growth}
    \end{subfigure}\hfill
    \begin{subfigure}[b]{0.66\textwidth}
        \centering
        \includegraphics[width=\linewidth]{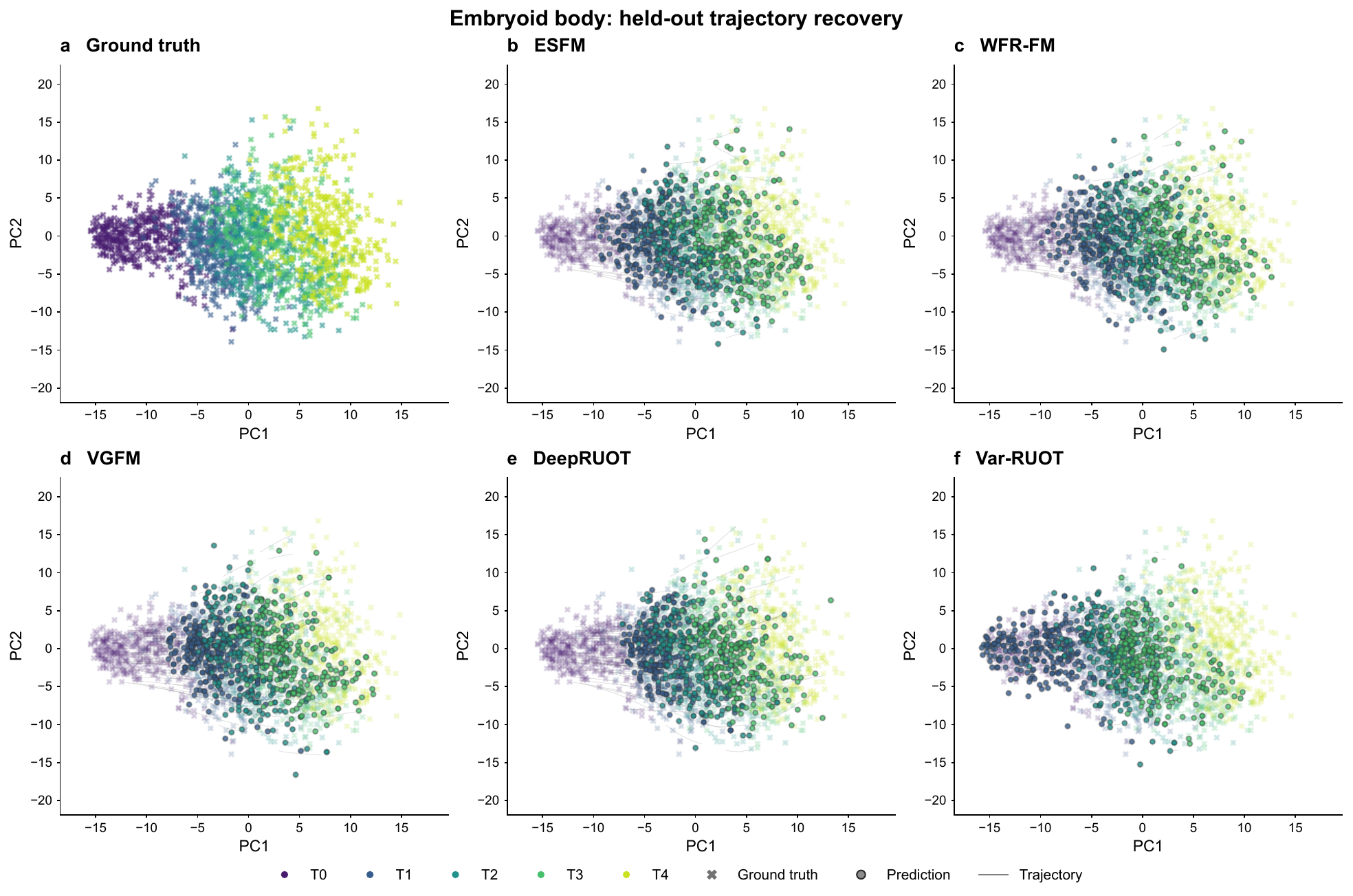}
        \caption{\textcolor{black}{Trajectory and marginal comparison.}}
        \label{fig:app-eb-heldout-trajectories}
    \end{subfigure}
    \caption{\textcolor{black}{Growth-field and trajectory predictions on the held-out 50D EB
    snapshots at $t=1,2,3$.}}
    \label{fig:app-eb-heldout-comparison}
\end{figure}

}%

\paragraph{CITE-seq.}
The CITE-seq dataset contains 31,240 cells collected at four time points \citep{lance2022multimodal}. Although the original data 
include both gene expression and surface protein measurements, our experiments use only the gene expression matrix, reduced to 50 
dimensions by PCA. The large number of cells and variation in snapshot size provide a test of prediction on unbalanced single-cell
 data. We set the hyperparameters to $\delta=1.5$, $\lambda=1.0$ and $\kappa=3.0$. As shown in \cref{tab:esfm-cite-detailed}, ESFM achieves the lowest
  distribution error at $t=2,3$ and the third-lowest at $t=1$, although some baselines attain lower mass errors.

\begin{table}[H]
\centering
\caption{Comparison of method performance over time on the 50D CITE dataset.}
\label{tab:esfm-cite-detailed}
\resizebox{\textwidth}{!}{
\begin{tabular}{lcccccc}
\toprule
\textbf{Method} & \multicolumn{2}{c}{$t=1$} & \multicolumn{2}{c}{$t=2$} & \multicolumn{2}{c}{$t=3$} \\
\cmidrule(lr){2-3}\cmidrule(lr){4-5}\cmidrule(lr){6-7}
& $\mathcal W_1$ & RME & $\mathcal W_1$ & RME & $\mathcal W_1$ & RME \\
\midrule
MMFM & 33.971 & --- & 36.854 & --- & 43.721 & --- \\
Metric FM & 28.314 & --- & 28.617 & --- & 33.212 & --- \\
SF2M & 29.543 & --- & 32.655 & --- & 36.265 & --- \\
MIOFlow & 28.290 & --- & 28.524 & --- & \second{32.230} & --- \\
TIGON & 28.196 & 0.186 & 27.921 & 0.545 & 32.846 & 0.653 \\
DeepRUOT & 28.245 & 0.168 & 27.908 & 0.525 & 32.950 & 0.634 \\
Var-RUOT & 30.219 & 0.331 & 32.702 & 0.325 & 40.613 & 0.486 \\
UOT-FM & 33.531 & \second{0.009} & 32.795 & 0.046 & 49.751 & 0.097 \\
VGFM & 29.449 & 0.020 & 29.722 & 0.057 & 33.752 & \best{0.001} \\
WFR-FM & \second{27.831} & 0.043 & \second{27.478} & \second{0.045} & 34.784 & 0.022 \\
USB & \best{27.021}{\tiny$\pm 5\times 10^{-5}$} & \best{0.002}{\tiny$\pm 1\times 10^{-6}$} & 28.081{\tiny$\pm 1\times 10^{-4}$} & \best{0.015}{\tiny$\pm 2\times 10^{-6}$} & 38.447{\tiny$\pm 0.003$} & \second{0.010}{\tiny$\pm 5\times 10^{-5}$} \\
\textbf{ESFM (ours)} & 27.860 & 0.055 & \best{26.609} & 0.095 & \best{28.813} & 0.094 \\
\bottomrule
\end{tabular}
}
\end{table}

{\color{black}%

\noindent\textbf{Held-out visualizations for CITE-seq.}
\Cref{fig:app-cite-heldout-growth} shows the learned ESFM growth fields at
the two held-out times $t=1,2$.
Using the corresponding hold-out models, we predict trajectories and cell
positions at these time points. We compare the predicted positions with the
observed snapshots and display the results together with the model-generated
trajectories in \cref{fig:app-cite-heldout-trajectories}.

\begin{figure}[H]
    \centering
    \begin{subfigure}[b]{0.32\textwidth}
        \centering
        \includegraphics[width=\linewidth]{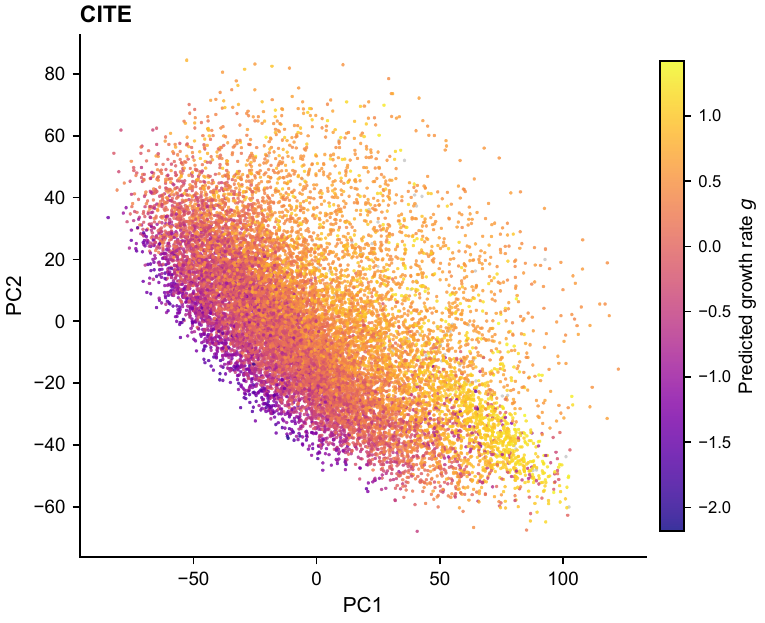}
        \caption{\textcolor{black}{ESFM growth predictions.}}
        \label{fig:app-cite-heldout-growth}
    \end{subfigure}\hfill
    \begin{subfigure}[b]{0.66\textwidth}
        \centering
        \includegraphics[width=\linewidth]{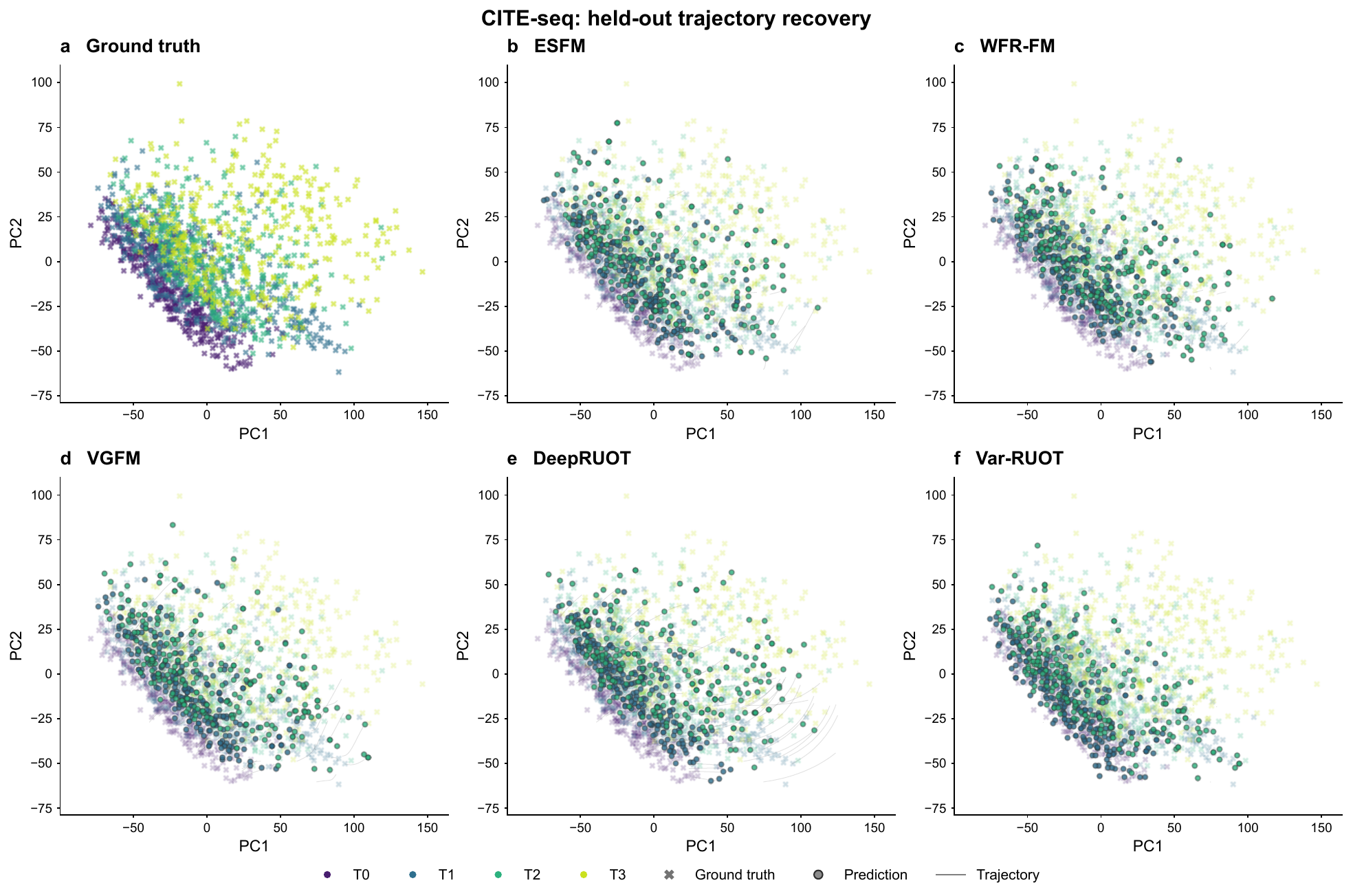}
        \caption{\textcolor{black}{Trajectory and marginal comparison.}}
        \label{fig:app-cite-heldout-trajectories}
    \end{subfigure}
    \caption{\textcolor{black}{Growth-field and trajectory predictions on the held-out 50D
    CITE-seq snapshots at $t=1,2$.}}
    \label{fig:app-cite-heldout-comparison}
\end{figure}
}%

\paragraph{Mouse hematopoiesis.}
The mouse hematopoiesis dataset contains 49,302 lineage-traced cells collected at three time points \citep{weinreb2020lineage}.
Its large cell population and substantial changes in
 cell abundance make it useful for evaluating both prediction and the computational cost of coupling construction. 
  We use PCA to reduce the gene expression data to 50 dimensions, and set the hyperparameters to
  $\delta=15$, $\lambda=225$ and $\kappa=1.5$, using mini-batch coupling with a mini-batch size of 2000. The prediction results in \cref{tab:esfm-mouse-detailed} show that ESFM achieves
  the lowest $\mathcal W_1$ and RME at $t=2$, while WFR-FM and USB attain lower distribution errors at $t=1$.

\begin{table}[H]
\centering
\caption{Comparison of method performance over time on the 50D Mouse dataset.}
\label{tab:esfm-mouse-detailed}
\resizebox{0.78\textwidth}{!}{
\begin{tabular}{lcccc}
\toprule
\textbf{Method} & \multicolumn{2}{c}{$t=1$} & \multicolumn{2}{c}{$t=2$} \\
\cmidrule(lr){2-3}\cmidrule(lr){4-5}
& $\mathcal W_1$ & RME & $\mathcal W_1$ & RME \\
\midrule
MMFM & 7.647 & --- & 10.156 & --- \\
Metric FM & 7.788 & --- & 11.449 & --- \\
SF2M & 8.217 & --- & 11.086 & --- \\
MIOFlow & 6.313 & --- & 6.746 & --- \\
TIGON & 6.140 & 0.382 & 6.973 & 0.326 \\
DeepRUOT & 6.052 & 0.062 & 6.757 & 0.041 \\
Var-RUOT & 7.951 & 0.131 & 10.862 & 0.154 \\
UOT-FM & 8.114 & 0.035 & 9.170 & 0.011 \\
VGFM & 6.274 & 0.076 & 6.796 & 0.070 \\
WFR-FM & \best{5.486} & \best{0.012} & \second{6.211} & 0.011 \\
USB & \second{5.589}{\tiny$\pm 1\times 10^{-6}$} & \second{0.013}{\tiny$\pm 6\times 10^{-8}$} & 6.548{\tiny$\pm 4\times 10^{-6}$} & \second{0.008}{\tiny$\pm 4\times 10^{-7}$} \\
\textbf{ESFM (ours)} & 5.604 & 0.073 & \best{5.539} & \best{0.004} \\
\bottomrule
\end{tabular}
}
\end{table}

{\color{black}%
\noindent\textbf{Held-out visualizations for Mouse.}
\Cref{fig:app-mouse-heldout-growth} shows ESFM's physical growth predictions
for the held-out intermediate snapshot at $t=1$.
Using the corresponding hold-out models, we predict trajectories and cell
positions at this time point. We compare the predicted positions with the
observed snapshot and display the results together with the model-generated
trajectories in \cref{fig:app-mouse-heldout-trajectories}.

\begin{figure}[H]
    \centering
    \begin{subfigure}[b]{0.32\textwidth}
        \centering
        \includegraphics[width=\linewidth]{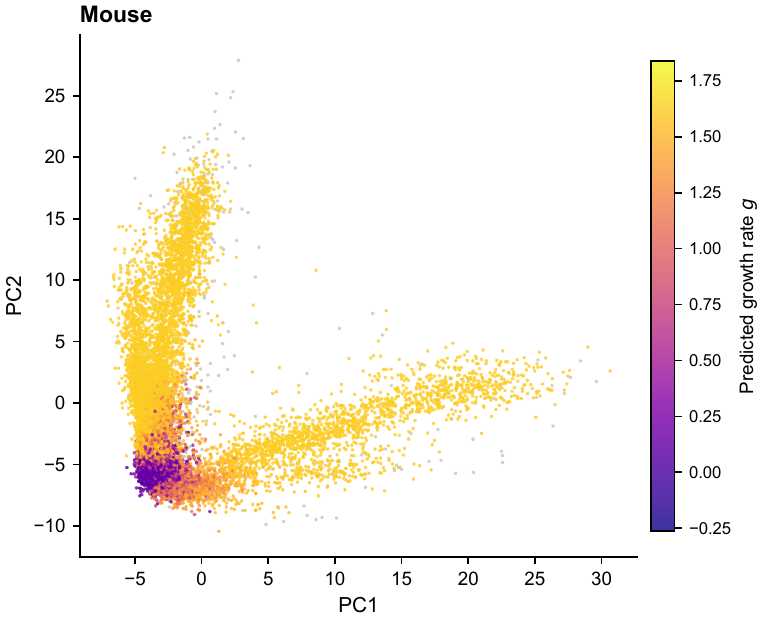}
        \caption{\textcolor{black}{ESFM growth predictions.}}
        \label{fig:app-mouse-heldout-growth}
    \end{subfigure}\hfill
    \begin{subfigure}[b]{0.66\textwidth}
        \centering
        \includegraphics[width=\linewidth]{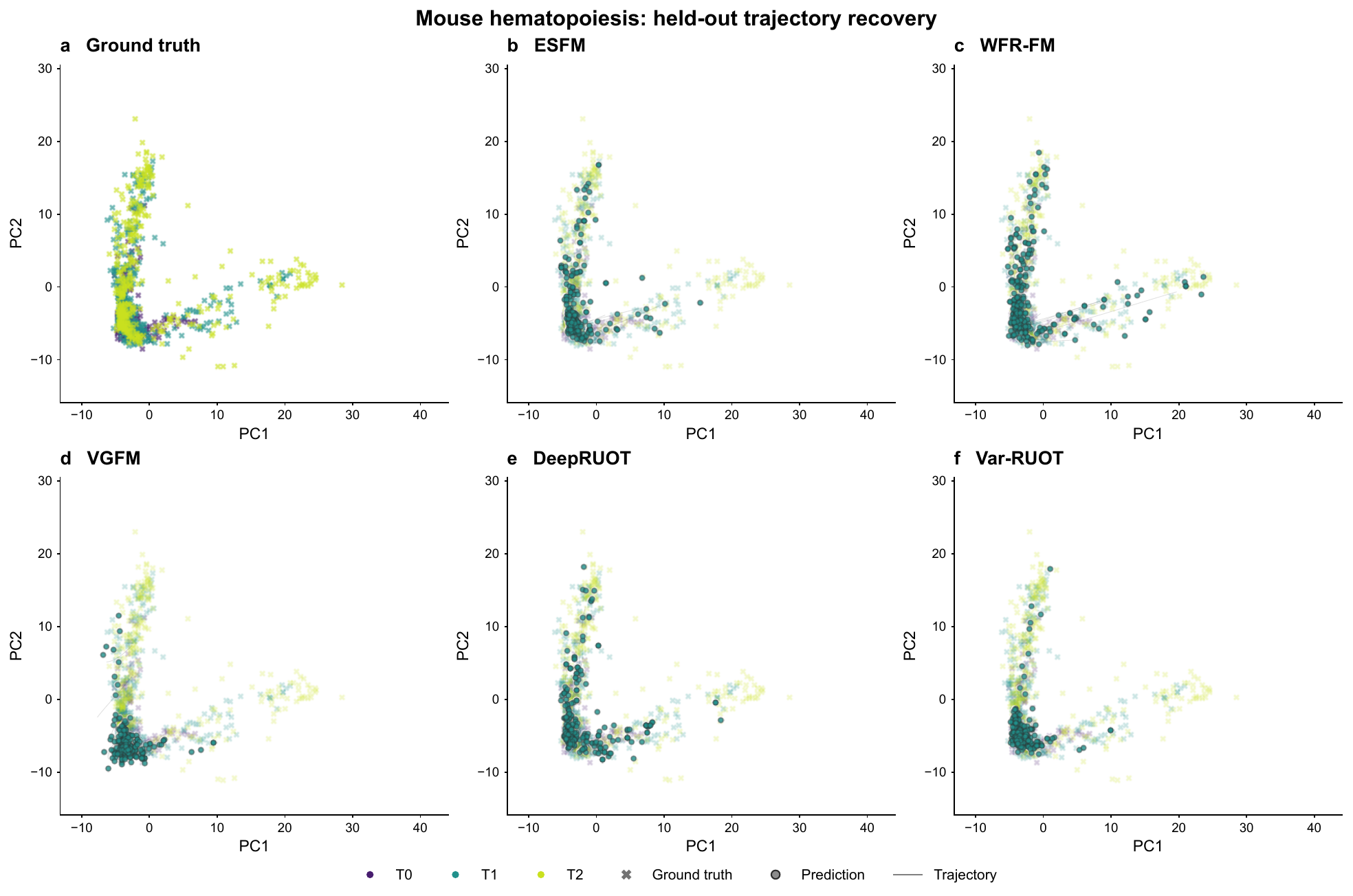}
        \caption{\textcolor{black}{Trajectory and marginal comparison.}}
        \label{fig:app-mouse-heldout-trajectories}
    \end{subfigure}
    \caption{\textcolor{black}{Growth-field and trajectory predictions on the held-out 50D
    Mouse snapshot at $t=1$.}}
    \label{fig:app-mouse-heldout-comparison}
\end{figure}

}%

\paragraph{Sensitivity to mini-batch coupling size.}
We examine the effect of coupling mini-batch size on the Mouse dataset in \cref{tab:esfm-mouse-minibatch-sensitivity}. 
Mini-batch coupling greatly reduces runtime compared with full coupling while maintaining comparable prediction performance.
The mini-batch sizes reported in the table yield similar runtime and comparable accuracy in both distribution and mass prediction.
  Full coupling yields slightly lower distribution error but requires much more computation.
  These results support mini-batch coupling as an efficient approximation, with the selected batch size providing a balance 
  between prediction quality and computational cost.

\begin{table}[H]
\centering
\caption{Sensitivity analysis for the batch size of mini-batch coupling on the 50D Mouse dataset.}
\label{tab:esfm-mouse-minibatch-sensitivity}
\begin{tabular}{lccccc}
\toprule
\textbf{Batch Size} & \multicolumn{2}{c}{\textbf{$t=1$}} & \multicolumn{2}{c}{\textbf{$t=2$}} & \textbf{Time (s)} \\
\cmidrule(lr){2-3}\cmidrule(lr){4-5}
& $\mathcal W_1$ & RME & $\mathcal W_1$ & RME & \\
\midrule
1000 & 5.712 & 0.113 & 5.583 & 0.0003 & 444 \\
1500 & 5.618 & 0.076 & 5.538 & 0.0007 & 407 \\
2000 & 5.604 & 0.073 & 5.539 & 0.004 & 389 \\
2500 & 5.597 & 0.084 & 5.522 & 0.003 & 386 \\
3000 & 5.611 & 0.116 & 5.520 & 0.003 & 394 \\
4000 & 5.560 & 0.097 & 5.518 & 0.002 & 407 \\
\midrule
w/o mini-batch & 5.483 & 0.101 & 5.459 & 0.001 & 5570 \\
\bottomrule
\end{tabular}
\end{table}

\subsection{Efficiency on the 100D EB dataset}
\label{app:eb100-efficiency}

\begin{figure}[H]
    \centering
    \includegraphics[width=0.6\textwidth]
    {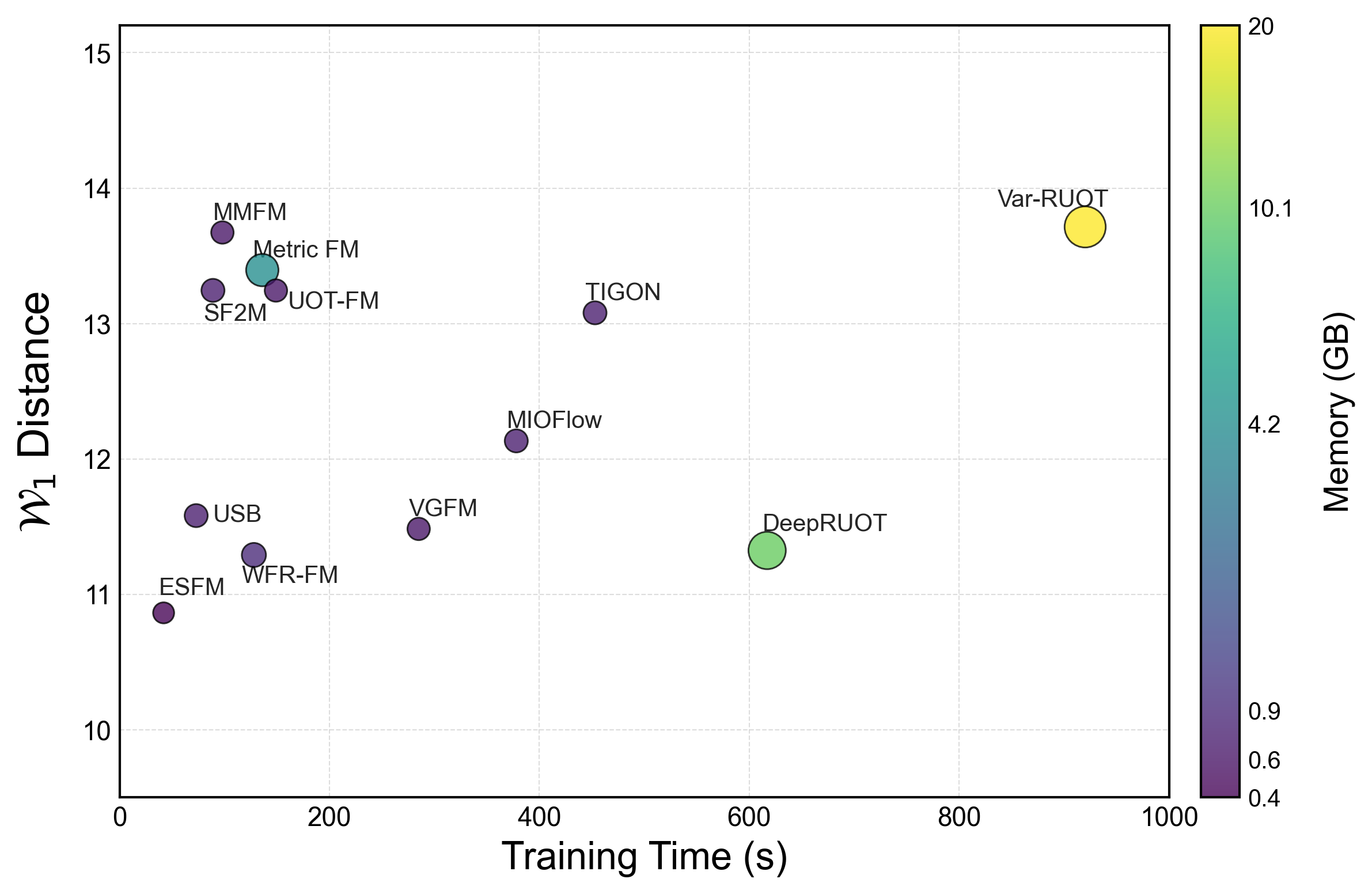}
    \caption{
    Efficiency comparison on the 100D EB dataset. 
    }
    \label{fig:esfm-efficiency}
\end{figure}

We evaluate the computational efficiency of ESFM on the 100-dimensional embryoid
body dataset (EB100) by comparing prediction accuracy, runtime and memory.
As shown in \cref{fig:esfm-efficiency}, ESFM achieves the lowest mean
$\mathcal W_1$ among the methods in the comparison, reaching $\mathcal W_1=10.863$
in only 42 seconds with 0.4 GB of memory. This runtime includes coupling
computation, endpoint pair presampling and network training. ESFM also has the
shortest reported runtime and lowest reported memory usage in the figure,
demonstrating an efficient combination of accurate prediction and low computational
cost on this large scale dataset.
Detailed prediction results, including RME, are provided in \cref{tab:esfm-eb-100d}.

\subsection{Evaluation of Static and Learned Dynamic Actions}
\label{app:path-coupling-action}
\paragraph{Static action.}
For an adjacent interval $[t_k,t_{k+1}]$ of duration $T_k=t_{k+1}-t_k$
with $n_k$ source particles, let
$\Gamma_{ij}^{(k)}$ and $R_{ij}^{(k)}$ denote the source coupling weight and endpoint mass
ratio returned by the entropy-regularized static solver. We compute the
geometric cost without the entropy term:
\begin{equation*}
 A_{\mathrm{stat}}^{(k)}
 =\frac{1}{n_kT_k}\sum_{i,j}\Gamma_{ij}^{(k)}
 \left[\|\bm x_j^{k+1}-\bm x_i^k\|^2+
 \delta^2\Phi_M(R_{ij}^{(k)})\right].
\end{equation*}

\paragraph{Network dynamic action.}
We initialize every observed source cell at
$\bm w_i(t_k)=(\bm x_i^k,0)$ and integrate
$\dot{\bm w}_i(t)=(\bm v_\theta(\bm w_i(t),t),g_\phi(\bm w_i(t),t))$.
The action is evaluated along these network-generated trajectories:
\begin{equation*}
\begin{aligned}
 A_{\mathrm{dyn}}^{(k)}
 =\frac{1}{n_k}\sum_{i=1}^{n_k}
 \int_{t_k}^{t_{k+1}}
 \Big[
 \|\bm v_\theta(\bm w_i(t),t)\|^2
 +\delta^2|g_\phi(\bm w_i(t),t)|^2
 \Big]\dd t.
\end{aligned}
\end{equation*}
We use forward Euler with 400 steps per interval.

\paragraph{Intervals and numerical checks.}
For each dataset, we compute the static and dynamic actions separately
within each interval between observed time points. The reported
actions are obtained by summing over all these intervals:
\[
 A_{\mathrm{stat}}=\sum_k A_{\mathrm{stat}}^{(k)},
 \qquad A_{\mathrm{dyn}}=\sum_k A_{\mathrm{dyn}}^{(k)},
 \qquad \mathrm{Gap}=\frac{|A_{\mathrm{dyn}}-A_{\mathrm{stat}}|}{A_{\mathrm{stat}}}\times100\%.
\]

\section{Relation to Other Unbalanced Flow Matching Methods}
\label{app:method-relations}

\subsection{Relation to VGFM}
\label{app:relation-vgfm}

Both VGFM \citep{wang2025joint} and ESFM jointly learn velocity and growth. They also use the same form of conditional interpolation once the endpoint states and masses are fixed. These paths and their targets are
\[
\begin{aligned}
    \bm{x}_t&=(1-t)\bm{x}_0+t\bm{x}_1,\\
    m_t&=m_0^{1-t}m_1^t,
    \qquad y_t=\ln m_t,\\
    \hat{\bm u}_t(\bm w_t\mid\bm z)&=\bm{x}_1-\bm{x}_0,\\
    \hat g_t(\bm w_t\mid\bm z)&=y_1-y_0=\ln(m_1/m_0).
\end{aligned}
\]
Let $\bm z=(\bm w_0,\bm w_1)$, where $\bm w_k=(\bm{x}_k,y_k)$ and $y_k=\ln m_k$ for $k=0,1$, and let $t\sim\mathcal U[0,1]$. The sampling laws $q_{\mathrm V}$ and $q_{\mathrm E}$ are induced by the OT solutions $P^\star$ and $(\Gamma^\star,R^\star)$, respectively, as specified below. The conditional matching losses are
\[
\begin{aligned}
    \mathcal L_{\mathrm{VGFM}}^{\mathrm{cond}}(\theta,\omega)
    &=\mathbb E_{\substack{t\sim\mathcal U[0,1]\\\bm z\sim q_{\mathrm V}}}
    \Big[
    \norm{\bm v_\theta(\bm{x}_t,t)-\hat{\bm u}_t(\bm w_t\mid\bm z)}^2\\
    &\qquad\qquad
    +\abs{g_\omega(\bm{x}_t,t)-\hat g_t(\bm w_t\mid\bm z)}^2
    \Big],\\[0.8em]
    \mathcal L_{\mathrm{ESFM}}^{\mathrm{cond}}(\theta,\phi)
    &=\mathbb E_{\substack{t\sim\mathcal U[0,1]\\\bm z\sim q_{\mathrm E}}}
    \Big[
    \norm{\bm v_\theta(\bm w_t,t)-\hat{\bm u}_t(\bm w_t\mid\bm z)}^2\\
    &\qquad\qquad
    +\lambda\abs{g_\phi(\bm w_t,t)-\hat g_t(\bm w_t\mid\bm z)}^2
    \Big].
\end{aligned}
\]
The main difference lies in the couplings underlying $q_{\mathrm V}$ and $q_{\mathrm E}$, which are used to sample endpoint pairs and growth labels. This changes the distribution of conditional regression targets and can therefore lead to different learned velocity fields.

For ESFM, the prescribed linear path in $(\bm{x},y)$, $y=\ln m$, minimizes the following action for fixed endpoints:
\[
\Acal_\delta(\bm w)
=\int_0^1\left(\norm{\dot{\bm{x}}_t}^2+\delta^2\abs{\dot y_t}^2\right)\dd t.
\]
Its minimum is the endpoint cost $c(\bm w_0,\bm w_1)=\norm{\bm{x}_1-\bm{x}_0}^2+\delta^2\abs{y_1-y_0}^2$, which induces the static OT formulation on the extended space $\mathcal W=\R^d\times\R$ (\cref{eq:main-extended-static}):
\[
\mathcal K_\delta(\rho_0,\rho_1)
=\inf_{\substack{\hat\rho_1\in\mathcal P(\mathcal W)\\
\hat\pi\in\Pi(\hat\rho_0,\hat\rho_1)}}
\int_{\mathcal W^2}c(\bm w_0,\bm w_1) \dd\hat\pi(\bm w_0,\bm w_1).
\]
For empirical weights $a_i,b_j$, let $\bm a=(a_i)_{i=1}^n$, write $C_{ij}=\norm{\bm{x}_i^0-\bm{x}_j^1}^2$ and let $R_{ij}$ be the local growth ratio. The discrete reduced problem is (\cref{eq:appendix-discrete}):
\[
\begin{aligned}
(\Gamma^\star,R^\star)\in\operatorname*{arg\,min}_{\Gamma\ge0, 0<R\le \e^M} 
&\sum_{i=1}^n\sum_{j=1}^m\left[C_{ij}+\delta^2\Phi_M(R_{ij})\right]\Gamma_{ij},\\
\text{s.t.}\quad 
&\sum_{j=1}^m\Gamma_{ij}=a_i,\qquad i=1,\ldots,n,\\
&\sum_{i=1}^n R_{ij}\Gamma_{ij}=b_j,\qquad j=1,\ldots,m.
\end{aligned}
\]

In contrast, VGFM obtains its coupling from a direct extension of quadratic Wasserstein OT (VGFM, Eq.~(11)):
\[
\begin{aligned}
P^\star\in\operatorname*{arg\,min}_{P\ge0} 
&\sum_{i=1}^n\sum_{j=1}^m C_{ij}P_{ij}+\tau\operatorname{KL}\!\left(P\bm 1\,\middle\|\,\bm a\right),\\
\text{s.t.}\quad &\sum_{i=1}^n P_{ij}=b_j,\qquad j=1,\ldots,m.
\end{aligned}
\]
VGFM computes its coupling using an extension of quadratic Wasserstein optimal transport, whereas ESFM uses the pairwise growth cost induced by its extended-space action. Although the conditional interpolations have the same form, ESFM derives its coupling and conditional paths from a common action, whereas VGFM combines a semi-relaxed OT coupling with a separately chosen constant relative-growth interpolation. VGFM does not establish that its coupling minimizes the action associated with the conditional paths used for training.

Coupling--path consistency matters because endpoint pairing and the path connecting each pair should reflect the same dynamical assumptions. The coupling determines which endpoints are paired, while the conditional path specifies how state and mass evolve between them. In ESFM, the cost of each extended endpoint pair is exactly the minimum action of the prescribed path connecting it. The coupling therefore minimizes the total action of the conditional paths used for training. If the coupling is instead chosen using a different cost, its preferred pairings and growth allocations need not minimize this action, even when the same interpolation formula is subsequently applied. Consistency therefore allows the assumed dynamics to inform the entire construction, from endpoint selection to conditional evolution. Under the assumptions of our static--dynamic equivalence theorem, the coupling and path construction together solve a fully specified OT problem.

An additional difference is the input state: ESFM parameterizes $\bm v_\theta(\bm w,t)$ and $g_\phi(\bm w,t)$, with $\bm w=(\bm{x},y)$ and $y=\ln m$, whereas VGFM uses $\bm v_\theta(\bm{x},t)$ and $g_\omega(\bm{x},t)$. Including log mass as an input expands the class of dynamics ESFM can represent.

\subsection{Relation to WFR-FM}

WFR-FM \citep{peng2026wfrfm} is designed as a simulation-free flow matching solver for the dynamic Wasserstein--Fisher--Rao optimal transport problem:
\begin{equation}
\begin{aligned}
    \mathrm{WFR}_{\delta}^{2}(\rho_0,\rho_1)
    :=
    \inf_{\rho_t,\bm u_t,g_t}\quad&
    \int_0^1\int_{\R^d}
    \left(
        \norm{\bm u_t(\bm{x})}^2
        +
        \delta^2\abs{g_t(\bm{x})}^2
    \right)
    \rho_t(\bm{x})\dd\bm{x}\dd t,
    \\
    \text{s.t.}\quad&
    \partial_t\rho_t
    +
    \nabla_{\bm{x}}\cdot(\rho_t\bm u_t)
    =
    \rho_t g_t,
    \qquad
    \rho_{t=0}=\rho_0,
    \quad
    \rho_{t=1}=\rho_1.
\end{aligned}
\label{eq:appendix-dynamic-wfr}
\end{equation}

WFR-FM builds its tractable solver from the traveling-Dirac geodesic between two Dirac measures. This geodesic provides the 
conditional state and mass paths together with their velocity and growth targets. The equivalence between the dynamic and static 
WFR formulations provides an OET coupling that can be computed with a Sinkhorn solver. Converting the OET coupling into a WFR 
semi-coupling specifies the paired endpoints and their masses. WFR-FM samples from this semi-coupling and trains its velocity 
and growth networks with a conditional regression loss weighted by mass.
 
ESFM reverses the usual design: it specifies a single-particle path first and then derives its action and coupling. 
The ESFM path enforces monotonic mass interpolation between its endpoints. This can better align with applications in which monotonic population change over an interval is an appropriate prior, whereas WFR geodesics may exhibit nonmonotonic mass evolution. 
More generally, choosing the WFR Dirac path within the extended space framework recovers the standard dynamic and static WFR 
formulations; WFR-FM is therefore a special case of the generalized ESFM framework. A detailed derivation is provided in \cref{app:wfr-as-extended-geodesic}.

\subsection{Relation to USB}

Like ESFM, USB \citep{ying2026beyond} learns a growth rate and avoids ODE or SDE trajectory simulation during training. It models the dynamics as a stochastic process:
\[
    \dd\bm{x}_t
    =
    \bm u_t(\bm{x}_t)\dd t
    +
    \nu\,\dd\bm W_t,
    \qquad
    \dd\ln m_t
    =
    g_t(\bm{x}_t)\dd t.
\]

This model connects USB to the Branching Schr\"odinger Bridge (BSB) problem. Following \citet{branRUOT}, USB relates BSB to a dynamic 
RUOT relaxation. Because the corresponding RUOT semi-coupling is 
not available in a tractable static form, its practical solver approximates the RUOT problem by WFR and computes a WFR/OET 
semi-coupling \citep{peng2026wfrfm}.

For a sampled endpoint pair $z=(\bm{x}_0,\bm{x}_1)\sim\gamma_0$, USB sets
$m_0(z)=1$ and $m_1(z)=\gamma_1(z)/\gamma_0(z)$. It then prescribes the
Poisson--Brownian conditional measure path
\[
    \rho_t(\bm{x}\mid z)
    =
    m_0(z)^{1-t}m_1(z)^t
    \mathcal N\!\left(
        \bm{x}\,\middle|\,
        (1-t)\bm{x}_0+t\bm{x}_1,
        \nu^2t(1-t)I
    \right),
\]
The state component follows a Brownian bridge with diffusion amplitude $\nu$, and the mass varies linearly on the log scale. USB uses neural networks to learn the probability flow velocity $v_\theta(\bm{x},t)$, growth rate $g_\theta(\bm{x},t)$, and score $s_\theta(\bm{x},t)$ through an objective for conditional unbalanced score matching.
 
The approximation of RUOT by WFR means that USB does not solve the original BSB problem exactly.
However, USB uses a WFR semi-coupling together with a Poisson--Brownian conditional path, so coupling--path consistency is 
 not established. By contrast, ESFM does not begin with a preselected OT functional. It first specifies the desired path in 
 the extended state space, constructs an action for which this path is a geodesic, and then derives the corresponding dynamic
  problem, static OT formulation, and endpoint coupling. Consequently, the ESFM conditional path and coupling are consistent 
  by construction.

\section{General Lagrangians and Alternative Geodesics}
\label{app:general-extensions}

The extended-space formulation also supports other choices of action and
conditional path. We present an action-first construction: starting from a
Lagrangian, we obtain minimizing paths, the induced endpoint cost and coupling,
and the corresponding FM targets. This complements the path-first design of
the specific ESFM action in the main text.
\Cref{app:general-geodesic-pipeline} gives the general construction and explains
how to obtain minimizing paths.
\Cref{app:m-linear-esfm} considers an action whose geodesics are linear in
physical mass, while \cref{app:wfr-as-extended-geodesic} recovers WFR formulation.

\subsection{From a Lagrangian to minimizing paths}
\label{app:general-geodesic-pipeline}

\noindent\textbf{Single-particle action and paths.}
Let $\bm w=(\bm{x},y)\in\mathcal W=\R^d\times\R$, with $y=\ln m$.
We start from a nonnegative Lagrangian
$L(\bm w,\bm v)$, which assigns a cost to the extended velocity $\bm v$
at state $\bm w$. For an absolutely continuous path, define
\[
    \Acal_L(\bm w)
    =
    \int_0^1 L(\bm w_t,\dot{\bm w}_t)\dd t,
    \qquad
    c_L(\bm w_0,\bm w_1)
    =
    \inf_{\bm w:\,\bm w_0\to\bm w_1}\Acal_L(\bm w).
\]
The conditional path is obtained by minimizing this action with fixed endpoints:
\[
    \Psi(\bm w_0,\bm w_1)
    \in
    \operatorname*{arg\,min}_{\bm w:\,\bm w_0\to\bm w_1}\Acal_L(\bm w),
    \qquad
    c_L(\bm w_0,\bm w_1)
    =
    \Acal_L\!\left(\Psi(\bm w_0,\bm w_1)\right).
\]
We refer to these minimizing paths as the geodesics associated with $L$.
When $L$ is the squared velocity norm of a Riemannian metric, they are
constant-speed minimizing geodesics of that metric.

\noindent\textbf{Computing the geodesics.}
For smooth $L$ and a regular minimizing path, the first variation gives
the Euler--Lagrange boundary-value problem
\[
    \frac{\dd}{\dd t}
    \partial_{\bm v}L(\bm w,\bm v)
    =
    \partial_{\bm w}L(\bm w,\bm v),
    \qquad
    \Psi_0=\bm w_0,\quad \Psi_1=\bm w_1.
\]
Here $\bm w=\Psi_t$ and $\bm v=\partial_t\Psi_t$.
This equation identifies candidate paths; minimality must still be verified.
For the main ESFM choice
$L((\bm{x},y),(\bm u,g))=\norm{\bm u}^2+\delta^2g^2$,
it yields $\partial_t^2\Psi_t=0$, hence
$\Psi_t=(1-t)\bm w_0+t\bm w_1$.
Cauchy--Schwarz verifies minimality and gives
$c_L=\norm{\bm{x}_1-\bm{x}_0}^2+\delta^2(y_1-y_0)^2$.

\noindent\textbf{Dynamic and static transport.}
Integrating the single-particle action over the extended distribution gives
\begin{equation}
\begin{aligned}
    \inf_{\hat\rho,\hat{\bm u},\hat g}\quad&
    \int_0^1\!\!\int_{\R^d\times\R}
    L\!\left(
        (\bm{x},y),(\hat{\bm u}_t,\hat g_t)
    \right)
    \hat\rho_t(\bm{x},y)
    \dd\bm{x}\dd y\dd t,
    \\
    \text{s.t.}\quad&
    \partial_t\hat\rho_t
    +\nabla_{\bm x}\cdot(\hat\rho_t\hat{\bm u}_t)
    +\partial_y(\hat\rho_t\hat g_t)=0,
    \\
    &
    \hat\rho_0(\bm{x},y)=\rho_0(\bm{x})\Dirac_0(y),
    \qquad
    \int_{\R}\e^y\hat\rho_1(\bm{x},y)\dd y=\rho_1(\bm{x}).
\end{aligned}
\label{eq:appendix-general-dynamic}
\end{equation}
The endpoint cost $c_L$ defines the corresponding extended static problem,
\begin{equation}
\begin{aligned}
    \inf_{\hat\pi,\hat\rho_1\ge0}\quad&
    \iiiint
    c_L\!\left(
        (\bm{x}_0,y_0),(\bm{x}_1,y_1)
    \right)
    \hat\pi(\bm{x}_0,y_0,\bm{x}_1,y_1)
    \dd\bm{x}_0\dd y_0\dd\bm{x}_1\dd y_1,
    \\
    \text{s.t.}\quad&
    \iint
    \hat\pi(\bm{x}_0,y_0,\bm{x}_1,y_1)
    \dd\bm{x}_1\dd y_1
    =
    \hat\rho_0(\bm{x}_0,y_0)
    =
    \rho_0(\bm{x}_0)\Dirac_0(y_0),
    \\
    &
    \iint
    \hat\pi(\bm{x}_0,y_0,\bm{x}_1,y_1)
    \dd\bm{x}_0\dd y_0
    =
    \hat\rho_1(\bm{x}_1,y_1),
    \\
    &
    \int_{\R}
    \e^{y_1}\hat\rho_1(\bm{x}_1,y_1)\dd y_1
    =
    \rho_1(\bm{x}_1).
\end{aligned}
\label{eq:appendix-general-static}
\end{equation}
If $L$ is convex in velocity, admissible dynamics admit trajectory
decompositions, and minimizing paths can be selected measurably,
\cref{eq:appendix-general-dynamic,eq:appendix-general-static} have the same
optimal value. The proof follows the particle-trajectory argument in
\cref{app:dynamic-static-equivalence}, using the convexity of $L$ in place
of the quadratic-action inequality.

\noindent\textbf{Conditional FM targets.}
Given a coupling from the static problem, sample endpoint pairs from its
normalized law and set $\bm w_t=\Psi_t(\bm w_0,\bm w_1)$. The regression targets are
\[
    (\bm u_t^\star,g_t^\star)
    =
    \partial_t\Psi_t(\bm w_0,\bm w_1),
    \qquad
    g_t^\star=\frac{\dd}{\dd t}\ln m_t.
\]
The networks $(\bm v_\theta,g_\phi)$ are trained with
\begin{equation}
    \mathbb E\left[
        \norm{\bm v_\theta(\bm{x}_t,y_t,t)-\bm u_t^\star}^2
        +
        \lambda
        \abs{g_\phi(\bm{x}_t,y_t,t)-g_t^\star}^2
    \right],
    \qquad \lambda>0,
    \label{eq:appendix-general-fm-loss}
\end{equation}
where $t\sim\mathcal U[0,1]$. 

The construction therefore proceeds as
\[
\begin{gathered}
    \text{Lagrangian and action}
    \;\Longrightarrow\;
    \text{minimizing paths and endpoint cost}
    \\
    \;\Longrightarrow\;
    \text{coupling}
    \;\Longrightarrow\;
    \text{FM targets}.
\end{gathered}
\]
An explicit endpoint cost or an efficient solver still depends on the chosen
$L$. 

\subsection{Example: ESFM with linear physical mass}
\label{app:m-linear-esfm}

We now replace the log-mass-linear ESFM geodesic by one that is linear in physical mass. For positive endpoint masses $m_0,m_1>0$, define
\begin{equation*}
    \bm{x}_t=(1-t)\bm{x}_0+t\bm{x}_1,
    \qquad
    m_t=(1-t)m_0+tm_1,
    \qquad
    y_t=\ln m_t.
\end{equation*}
This is the constant-speed straight path in the coordinate $(\bm{x},m)$. It minimizes
\begin{equation*}
\begin{aligned}
    \Acal_{\mathrm{lin}}(\bm{x},m)
    &=
    \int_0^1
    \left(
        \norm{\dot{\bm{x}}_t}^2
        +
        \delta^2\abs{\dot m_t}^2
    \right)\dd t
    \\
    &=
    \int_0^1
    \left(
        \norm{\dot{\bm{x}}_t}^2
        +
        \delta^2\e^{2y_t}\abs{\dot y_t}^2
    \right)\dd t,
\end{aligned}
\end{equation*}
where the second line rewrites exactly the same action in the ESFM coordinate $y=\ln m$. By Cauchy--Schwarz, its endpoint cost is
\begin{equation*}
    c_{\mathrm{lin}}(\bm w_0,\bm w_1)
    =
    \norm{\bm{x}_1-\bm{x}_0}^2
    +
    \delta^2\left(\e^{y_1}-\e^{y_0}\right)^2.
\end{equation*}
Hence this is still ESFM on $(\bm{x},y)$, but with the nonconstant mass-coordinate metric factor $\e^{2y}$ instead of the constant metric used in the main text.

As in the main ESFM formulation, normalize the source particle mass to $m_0=1$, so $y_0=0$. The extended static problem becomes
\begin{equation*}
\begin{aligned}
    \inf_{\pi\ge0}\quad&
    \iiint
    \left[
        \norm{\bm{x}_1-\bm{x}_0}^2
        +
        \delta^2(\e^{y_1}-1)^2
    \right]
    \pi(\bm{x}_0,\bm{x}_1,y_1)
    \dd\bm{x}_0\dd\bm{x}_1\dd y_1,
    \\
    \text{s.t.}\quad&
    \iint\pi(\bm{x}_0,\bm{x}_1,y_1)
    \dd\bm{x}_1\dd y_1
    =\rho_0(\bm{x}_0),
    \\
    &
    \iint\e^{y_1}\pi(\bm{x}_0,\bm{x}_1,y_1)
    \dd\bm{x}_0\dd y_1
    =\rho_1(\bm{x}_1).
\end{aligned}
\end{equation*}
For a fixed spatial pair with $A>0$, let $f$ be the nonnegative measure of terminal log-mass and set
\[
    A=\int_{\R}\dd f(y_1),
    \qquad
    B=\int_{\R}\e^{y_1}\dd f(y_1),
    \qquad
    R=\frac{B}{A}.
\]
The local mass problem is
\[
    \inf_{f\ge0}
    \int_{\R}(\e^{y_1}-1)^2\dd f(y_1),
    \qquad
    \int_{\R}\dd f=A,
    \quad
    \int_{\R}\e^{y_1}\dd f=B.
\]
Normalize $f$ to the probability measure $\mu$ by $\dd\mu(y_1)=\dd f(y_1)/A$. The two constraints then give
\[
    \int_{\R}\dd\mu=1,
    \qquad
    \int_{\R}\e^{y_1}\dd\mu(y_1)=\frac{B}{A}=R.
\]
Therefore, applying Jensen's inequality to the strictly convex function $\phi(r)=(r-1)^2$ gives
\begin{equation*}
\begin{aligned}
    \int_{\R}(\e^{y_1}-1)^2\dd f(y_1)
    &=
    A\int_{\R}\phi(\e^{y_1})\dd\mu(y_1)
    \\
    &\ge
    A\phi\!\left(\int_{\R}\e^{y_1}\dd\mu(y_1)\right)
    =A(R-1)^2.
\end{aligned}
\end{equation*}
Since $\phi$ is strictly convex, equality holds if and only if $\e^{y_1}=R$, or equivalently $y_1=\ln R$. Hence
\[
    f^\star=A\Dirac_{\ln R}.
\]
Unlike the capped log-mass quadratic cost in the main ESFM model, this local problem does not require a two-point optimizer: all local mass is assigned the deterministic terminal value $y_1=\ln R$.

Defining the semi-couplings $\gamma_0$ and $\gamma_1$ as in the main text, with $R=\dd\gamma_1/\dd\gamma_0>0$ , eliminates $y_1$ and gives
\begin{equation}
\begin{aligned}
    \inf_{\gamma_0,\gamma_1\ge0}\quad&
    \iint
    \left[
        \norm{\bm{x}_1-\bm{x}_0}^2
        +
        \delta^2
        \left(
            \frac{\dd\gamma_1}{\dd\gamma_0}-1
        \right)^2
    \right]
    \dd\gamma_0,
    \\
    \text{s.t.}\quad&
    \int\gamma_0(\bm{x}_0,\bm{x}_1)\dd\bm{x}_1
    =\rho_0(\bm{x}_0),
    \\
    &
    \int\gamma_1(\bm{x}_0,\bm{x}_1)\dd\bm{x}_0
    =\rho_1(\bm{x}_1).
\end{aligned}
\label{eq:appendix-m-linear-reduced}
\end{equation}
The only change from the main ESFM reduced problem is therefore the local growth penalty: $\Phi_M(R)$ is replaced by $(R-1)^2$.

\subsection{Entropic coupling and conditional targets for the example}
\label{app:m-linear-solver}

For discrete measures with source weights $a_i$, target weights $b_j$, and $C_{ij}=\norm{\bm{x}_i^0-\bm{x}_j^1}^2$, an entropic version of \cref{eq:appendix-m-linear-reduced} is
\begin{equation*}
\begin{aligned}
    \min_{\Gamma_{ij}\ge0,\,R_{ij}>0}\quad&
    \sum_{i,j}
    \left[
        C_{ij}\Gamma_{ij}
        +
        \delta^2\Gamma_{ij}(R_{ij}-1)^2
        +
        \varepsilon\Gamma_{ij}(\ln\Gamma_{ij}-1)
    \right],
    \\
    \text{s.t.}\quad&
    \sum_j\Gamma_{ij}=a_i,
    \qquad
    \sum_iR_{ij}\Gamma_{ij}=b_j.
\end{aligned}
\end{equation*}
Introducing row and column dual variables $u_i$ and $v_j$, elimination of $R$ uses the one-dimensional conjugate
\[
    H_j(v_j)
    =
    \sup_{R>0}
    \left\{v_jR-\delta^2(R-1)^2\right\}.
\]
On the interior branch, its maximizer and value are
\begin{equation}
    R_j=1+\frac{v_j}{2\delta^2},
    \qquad
    H_j=\delta^2(R_j^2-1).
    \label{eq:appendix-m-linear-conjugate}
\end{equation}
The row update has the familiar ESFM form
\begin{equation*}
    u_i
    =
    \varepsilon\ln a_i
    -
    \varepsilon\ln
    \sum_j
    \exp\!\left(\frac{H_j-C_{ij}}{\varepsilon}\right).
\end{equation*}
For fixed $u$, define
\[
    K_j
    =
    \frac{b_j}{\sum_i\exp((u_i-C_{ij})/\varepsilon)}.
\]
The column constraint and \cref{eq:appendix-m-linear-conjugate} reduce the column update to the scalar equation
\begin{equation}
    \ln R_j
    +
    \frac{\delta^2}{\varepsilon}(R_j^2-1)
    =
    \ln K_j.
    \label{eq:appendix-m-linear-column-update}
\end{equation}
Its left-hand side is strictly increasing on $R_j>0$, so \cref{eq:appendix-m-linear-column-update} has a unique positive solution.
 The semi-couplings are recovered as
\begin{equation*}
    \Gamma_{ij}^{(0)}
    =
    \exp\!\left(\frac{u_i+H_j-C_{ij}}{\varepsilon}\right),
    \qquad
    \Gamma_{ij}^{(1)}
    =
    R_j\Gamma_{ij}^{(0)}.
\end{equation*}
Thus changing the geodesic changes the scalar growth update, but preserves the same coupling architecture used by ESFM.

Finally, sample $(i,j)$ from the normalized base coupling $\Gamma^{(0)}$, set $m_0=1$ and $m_1=R_j$, and use
\begin{equation*}
\begin{aligned}
    \bm{x}_t
    &=(1-t)\bm{x}_i^0+t\bm{x}_j^1,
    \\
    m_t
    &=1+t(R_j-1),
    \qquad
    y_t=\ln\!\left(1+t(R_j-1)\right),
    \\
    \bm u_t^\star
    &=\bm{x}_j^1-\bm{x}_i^0,
    \qquad
    g_t^\star
    =\dot y_t
    =
    \frac{R_j-1}{1+t(R_j-1)}
    =(R_j-1)\e^{-y_t}.
\end{aligned}
\end{equation*}
Substitution of these targets into \cref{eq:appendix-general-fm-loss} gives a simulation-free ESFM training objective for the linear-mass geodesic.

\subsection{Example: recovering WFR from the WFR geodesic}
\label{app:wfr-as-extended-geodesic}

The general construction in \cref{app:general-geodesic-pipeline} also contains WFR as a particular choice of extended space 
geometry. Keep the coordinate $y=\ln m$, but replace the main ESFM action by the microscopic action
\begin{equation}
    \Acal_{\mathrm{WFR}}(\bm w)
    =
    \int_0^1
    \e^{y_t}
    \left(
        \norm{\dot{\bm{x}}_t}^2
        +
        \delta^2\abs{\dot y_t}^2
    \right)\dd t.
    \label{eq:appendix-wfr-action}
\end{equation}
Because $m_t=\e^{y_t}$ and $\dot y_t=\dot m_t/m_t$, this is exactly the microscopic WFR action. Its minimizing Dirac-to-Dirac paths are the WFR geodesics, including the traveling Dirac in the moving regime. Define
\[
    \cos_+(s)
    =
    \cos\!\left(
        \min\!\left\{
            s,
            \frac{\pi}{2}
        \right\}
    \right),\qquad s\ge0.
\]
The endpoint cost induced by \cref{eq:appendix-wfr-action} is
\begin{equation*}
\begin{aligned}
    c_{\mathrm{WFR}}(\bm w_0,\bm w_1)
    =4\delta^2\Big[
        \e^{y_0}+\e^{y_1}
        -2\cos_+\!\left(\frac{\norm{\bm{x}_1-\bm{x}_0}}{2\delta}\right)
        \e^{(y_0+y_1)/2}
    \Big].
\end{aligned}
\end{equation*}

Apply the extended static construction in \cref{eq:appendix-general-static} with $y_0=0$. At a fixed spatial pair, let $f$ denote the terminal log-mass measure and set
\[
    A=\int\dd f(y_1),
    \qquad
    B=\int\e^{y_1}\dd f(y_1).
\]
The Cauchy--Schwarz inequality gives
\[
    \left(\int\e^{y_1/2}\dd f(y_1)\right)^2
    \le AB,
\]
and the bound is attained by $f^*=A\Dirac_{\ln(B/A)}$. Consequently, the optimized local cost is
\begin{equation*}
    4\delta^2
    \left[
        A+B
        -2\cos_+\!\left(\frac{\norm{\bm{x}_1-\bm{x}_0}}{2\delta}\right)\sqrt{AB}
    \right].
\end{equation*}

Defining the semi-couplings $\gamma_0=\int\pi\dd y_1$ and $\gamma_1=\int\e^{y_1}\pi\dd y_1$ and integrating the local value over spatial pairs therefore yields
\begin{equation*}
\begin{aligned}
    &\inf_{(\gamma_0,\gamma_1)\in\Gamma(\rho_0,\rho_1)}
    4\delta^2\iint
    \Big[
        \gamma_0(\bm{x}_0,\bm{x}_1)
        +\gamma_1(\bm{x}_0,\bm{x}_1)
        \\
        &\qquad
        -2\cos_+\!\left(\frac{\norm{\bm{x}_1-\bm{x}_0}}{2\delta}\right)
        \sqrt{\gamma_0(\bm{x}_0,\bm{x}_1)
               \gamma_1(\bm{x}_0,\bm{x}_1)}
    \Big]
    \dd\bm{x}_0\dd\bm{x}_1,
\end{aligned}
\end{equation*}
where $\Gamma(\rho_0,\rho_1)$ imposes the source marginal on $\gamma_0$ and the target marginal on $\gamma_1$. This is the standard static WFR semi-coupling functional corresponding to \cref{eq:appendix-dynamic-wfr}, equivalent to the formulation used in WFR-FM \citep[Equation~(3.7)]{peng2026wfrfm}. Our action omits their overall factor of $1/2$, which leaves the optimal couplings unchanged. Using the same geodesic to construct conditional targets and retaining $\e^{y_t}=m_t$ as the regression weight gives the mass-weighted conditional loss used by WFR-FM. Hence WFR-FM is a special case of the generalized extended space framework when the WFR geodesic and action are selected.

\section{Intuition and Limitations of Minibatch OT}
\label{app:limitations}

For large single-cell datasets, ESFM may use mini-batch or chunked coupling computations for memory reasons. This is a practical 
approximation to the full reduced coupling problem in \cref{eq:main-reduced-static}. As in WFR-FM, the theoretical convergence 
of mini-batch unbalanced or extended space couplings to the full coupling is not established here. Empirically, chunking is 
used as an engineering device.

We also conducted an ablation study on the choice of OT batch size on the Mouse dataset
(see \cref{tab:esfm-mouse-minibatch-sensitivity}). Prediction performance remains broadly
stable across the tested batch sizes, suggesting that ESFM is robust to the choice of
mini-batch coupling size in this setting.

\section{Pseudocode for ESFM Inference}
\label{app:inference-details}
Given the trained field $(\bm v_\theta,g_\phi)$, inference starts from $(\bm{x}_0,0)$ and solves
\begin{equation*}
    \frac{\dd}{\dd t}
    \begin{pmatrix}
    \bm{x}_t\\
    y_t
    \end{pmatrix}
    =
    \begin{pmatrix}
    \bm v_\theta(\bm{x}_t,y_t,t)\\
    g_\phi(\bm{x}_t,y_t,t)
    \end{pmatrix}.
\end{equation*}
The predicted physical density is recovered by
\[
    \rho_1^{\mathrm{pred}}(\bm{x})
    =
    \int \e^{y}\hat\rho_1^{\mathrm{pred}}(\bm{x},y)\dd y.
\]

\Cref{alg:esfm-inference} summarizes the inference procedure using forward Euler steps.

\begin{algorithm}[H]
\caption{ESFM inference workflow}
\label{alg:esfm-inference}
\begin{algorithmic}[1]
\STATE \textbf{Input:} initial samples $\{\bm{x}_0^{(i)}\}_{i=1}^N$, trained field $\bm s_{\theta,\phi}=(\bm v_\theta,g_\phi)$, final time $T$, step size $\Delta t$.
\STATE Initialize $\bm{x}^{(i)}\gets \bm{x}_0^{(i)}$, $y^{(i)}\gets0$, and $t\gets0$ for all $i$.
\WHILE{$t<T$}
    \STATE Set $\Delta t_{\mathrm{step}}\gets\min(\Delta t,T-t)$.
    \FOR{$i=1,\dots,N$}
        \STATE Evaluate $(\bm v,g)=\bm s_{\theta,\phi}(\bm{x}^{(i)},y^{(i)},t)$.
        \STATE Update $\bm{x}^{(i)}\gets\bm{x}^{(i)}+\bm v\,\Delta t_{\mathrm{step}}$.
        \STATE Update $y^{(i)}\gets y^{(i)}+g\,\Delta t_{\mathrm{step}}$.
    \ENDFOR
    \STATE Set $t\gets t+\Delta t_{\mathrm{step}}$.
\ENDWHILE
\STATE \textbf{Output:} evolved extended particles $\{(\bm{x}^{(i)},y^{(i)})\}_{i=1}^N$ and physical weights $\{\e^{y^{(i)}}\}_{i=1}^N$.
\end{algorithmic}
\end{algorithm}

\end{document}